\documentclass{article}

 \usepackage[preprint]{neurips_2026}

\usepackage[utf8]{inputenc} 
\usepackage[T1]{fontenc}    
\usepackage{hyperref}       
\usepackage{url}            
\usepackage{booktabs}       
\usepackage{amsfonts}       
\usepackage{nicefrac}       
\usepackage{microtype}      
\usepackage{xcolor}         
 \usepackage{graphicx}
 \usepackage{amsmath}
 \usepackage{amssymb}
\usepackage{enumitem}
\usepackage{float}
\usepackage{multirow} 

\title{The Geometric Alignment Tax: Tokenization vs. Continuous Geometry in Scientific Foundation Models}

\author{%
  Prashant C.~Raju\\
  \texttt{rajuprashant@gmail.com}
}

\begin{document}

\maketitle

\begin{abstract}
Foundation models for biology and physics optimize predictive accuracy, 
but their internal representations systematically fail to preserve the 
continuous geometry of the systems they model. We identify the root 
cause: the \textit{Geometric Alignment Tax}, an intrinsic cost of 
forcing continuous manifolds through discrete categorical bottlenecks. 
Controlled ablations on synthetic dynamical systems demonstrate that 
replacing cross-entropy with a continuous head on an identical encoder 
reduces geometric distortion by up to $8.5\times$, while learned 
codebooks exhibit a non-monotonic double bind where finer quantization 
worsens geometry despite improving reconstruction. Under continuous 
objectives, three architectures differ by $1.3\times$; under discrete 
tokenization, they diverge by $3{,}000\times$. Evaluating 14 
biological foundation models with rate-distortion theory and MINE, we 
identify three failure regimes: \textit{Local-Global Decoupling}, 
\textit{Representational Compression}, and \textit{Geometric Vacuity}. 
A controlled experiment confirms that Evo~2's reverse-complement 
robustness on real DNA reflects conserved sequence composition, not 
learned symmetry. No model achieves simultaneously low distortion, 
high mutual information, and global coherence.
\end{abstract}
\section{Introduction}
\label{sec:intro}
Foundation models for biology and physics are evaluated on predictive accuracy: perplexity, AUC, benchmark rankings. But these metrics are blind to whether the model's internal representations preserve the continuous geometry of the systems they claim to encode. We reveal a hidden cost: the \textbf{Geometric Alignment Tax}, the intrinsic geometric distortion incurred when forcing continuous physical manifolds through discrete categorical bottlenecks. We have previously introduced the geometric tax to describe transferability-fidelity trade-offs in vision  architectures~\citep{raju2026geometric} and subsequently applied it to perturbation coherence in biological manifolds~\citep{raju2026crispr}. Here, we identify its most fundamental manifestation: the application of discrete-token foundation models to continuous physical symmetries.

\paragraph{Resolution is not continuity.}
Consider constructing a smooth ramp from discrete rectangular blocks. Shrinking the blocks creates the illusion of a continuous surface, but rolling a marble down it reveals the truth: each microscopic edge introduces a tiny directional perturbation, and the cumulative angular error at the bottom does not vanish as blocks shrink. It decays so slowly that practical convergence is unreachable. The surface is not smooth; it is a high-resolution approximation of roughness. Foundation models that quantize continuous data into discrete vocabularies operate under this exact structural divergence. Scaling parameters and context windows shrinks the steps between vocabulary bins, minimizing macroscopic error and creating an illusion of geometric fidelity. But the manifold remains fractured, and the fracture is governed by scaling laws that make convergence impractically slow.

\paragraph{The core claim.}
Cross-entropy loss over discrete tokens is a sufficient condition for symmetry failure in embedding manifolds. The tax is not a property of attention, recurrence, or convolution; it is the price of discretizing a continuous world before processing it. On synthetic dynamical systems with known geometry, three architectures (Transformer, SSM, hybrid) trained with continuous objectives differ by $1.3\times$ in geometric stability. Under discrete tokenization, the same architectures diverge by $3{,}000\times$ on a biological mutation walk (Section~\ref{sec:ground-truth}). The discrete-to-continuous gap within any single architecture dwarfs the cross-architecture gap. Learned VQ codebooks cannot escape this cost: finer quantization improves reconstruction but worsens geometric stability by increasing boundary-crossing probability under perturbation, and the empirical distortion scaling ($1/\!\log K$) implies exponentially more codes would be needed to approach continuous performance. The ESM-2 protein Transformer suite (8M--15B) exhibits a progressive decline in geometric stability with scale, and an apparent ``recovery'' at 15B is unmasked as global manifold drift rather than genuine improvement (Section~\ref{sec:phase}). A controlled four-condition experiment demonstrates that Evo~2's apparent reverse-complement robustness on real DNA reflects conserved $k$-mer composition, not learned symmetry (Section~\ref{sec:context-rc}).

We formalize the tax through rate-distortion theory and validate the bound empirically (Section~\ref{sec:info-theory}). Applying MINE across 14 foundation models, we identify three failure regimes: \textit{Local-Global Decoupling}, where models encode shallow local statistics but fail to integrate globally; \textit{Representational Compression}, where information concentrators amplify mutual information at the cost of geometric fidelity; and \textit{Geometric Vacuity}, where smooth embeddings carry less structure than random noise.

\paragraph{Contributions.} 
(1)~Controlled synthetic experiments isolating tokenization as the causal bottleneck for geometric instability (Section~\ref{sec:ground-truth}). (2)~Scaling laws demonstrating the tax is progressive with parameters and invariant to context length, across 14 biological foundation models (Section~\ref{sec:models}). (3)~An information-theoretic formalization via rate-distortion theory and MINE, identifying three distinct failure regimes (Section~\ref{sec:info-theory}). (4)~A comprehensive ablation battery (6 variants) ruling out alternative explanations (Appendix~\ref{app:ablation}). (5)~The Texture Hypothesis Test: a controlled experiment establishing that Evo~2's RC signal on real DNA is per-sequence $k$-mer histogram matching, not biophysical understanding (Section~\ref{sec:context-rc}). (6)~An RCCR experiment demonstrating that post-hoc symmetry regularization degrades manifold geometry despite achieving perfect RC consistency (Section~\ref{sec:context-rc}).

\section{Ground Truth: The Controlled Experiments}
\label{sec:ground-truth}
We isolate the causal role of tokenization by training three small architectures from scratch on synthetic dynamical systems with known continuous geometry. The architectures span the dominant paradigms: SmallBERT (Transformer, 3.4M parameters), SmallMamba (SSM, 2.0M), and SmallStripedHyena (hybrid Hyena-attention, 4.5M). Each is trained via causal language modeling (CLM) with 256-bin uniform discretization on three datasets: superposed sine waves (\textsc{waveform}), damped harmonic oscillators (\textsc{oscillator}), and Lorenz attractors (\textsc{lorenz}). A two-pass global discretization scheme computes dataset-wide min/max values first, ensuring the same physical state maps to the same bin across all sequences and preventing per-sequence normalization artifacts.
 
\paragraph{Evaluation protocol.}
Geometric stability is evaluated using a standardized harness built on the Shesha geometry library~\citep{raju2026geometric,shesha2026}. For each model, we embed both clean and perturbed sequences, extract a center window (mean-pooled to a fixed-length vector to neutralize context-length differences), and compute pairwise Representational Dissimilarity Matrices (RDMs) using cosine distance (per-model layer and pooling details in Table~\ref{tab:extraction}, Appendix~\ref{app:extraction}). The harness reports four core metrics. \textit{RDM similarity}: the Spearman correlation between the clean and perturbed RDMs, measuring whether pairwise geometric relationships are preserved under perturbation. \textit{Perturbation stability}: the rank correlation between input-space perturbation magnitude and embedding-space displacement, testing whether small input changes produce proportionally small representation changes. \textit{Feature split} and \textit{sample split} scores: the agreement between RDMs computed on random halves of the embedding dimensions and random halves of the dataset, respectively, verifying that geometric structure is distributed rather than concentrated in fragile subspaces. The \textit{composite stability} score is the mean of these four metrics (see Appendix~\ref{app:stability_metrics} for formal definitions). Perturbations include value noise at 1\%, 2\%, 5\%, and 10\% of positions, plus time reversal. We additionally validate dynamical fidelity on the Lorenz attractor via largest Lyapunov exponent (LLE) estimation and a butterfly test for attractor preservation. Reproducibility details (seeds, hardware, training hyperparameters) appear in Appendix~\ref{app:reproducibility}.
 
\subsection{The Causal Proof: Continuous vs.\ Discrete}
\label{sec:causal-proof}

\paragraph{Baseline (discrete CE).}
Under standard discrete tokenization, all three architectures preserve Lorenz attractor dynamics: LLE estimates are $0.036$ (SmallBERT), $0.038$ (SmallMamba), and $0.038$ (SmallStripedHyena), all within 3\% of the ground truth value $0.037$, and the butterfly test confirms attractor structure preservation across all 5 seeds for every architecture. The cross-architecture variance in geometric stability is modest. At 1\% noise on the Lorenz dataset, Procrustes distortion $D$ ranges from $0.072$ (SmallStripedHyena) to $0.157$ (SmallMamba), with SmallBERT intermediate at $0.096$. Mean composite stability scores across all datasets and perturbations are $0.466 \pm 0.024$ (SmallBERT), $0.400 \pm 0.059$ (SmallMamba), and $0.470 \pm 0.016$ (SmallStripedHyena). Paired Wilcoxon signed-rank tests show SmallBERT is significantly more stable than SmallMamba ($W = 0.0$, $p = 0.0001$), while the SmallBERT vs.\ SmallStripedHyena difference is not significant ($W = 34.0$, $p = 0.15$). Architectures differ, but within the same order of magnitude.

\paragraph{Ablation Variant A (continuous MSE head).}
We replace the categorical CE output head with a linear projection trained under MSE loss, leaving the encoder backbone (self-attention layers, positional embeddings, feedforward blocks) unchanged. This single modification eliminates manifold fracture across all architectures. On the Lorenz dataset at 1\% noise, SmallBERT improves $2.8\times$ ($D\colon 0.096 \to 0.034$); SmallStripedHyena improves $8.5\times$ ($D\colon 0.072 \to 0.0085$), the single best condition in the entire study. The cross-architecture spread collapses from $0.072$-$0.157$ under discrete CE to $0.0085$-$0.034$ under continuous MSE. At 10\% noise the pattern holds: SmallBERT discrete $D = 0.316$ vs.\ continuous $D = 0.111$ ($2.8\times$); SmallStripedHyena discrete $D = 0.233$ vs.\ continuous $D = 0.027$ ($8.6\times$). The discrete-to-continuous gap within any single architecture dwarfs the cross-architecture gap under either regime. Same encoders, same training data, same perturbation protocol: the only variable is the output discretization boundary. We note that inputs remain discretized in both conditions; the ablation isolates the output objective, not input tokenization. The gain therefore reflects the interaction between discrete representations and the CE loss landscape. Fully continuous pipelines would likely show even larger improvements.

\paragraph{VQ bottleneck proof (the double bind).}
To preempt the objection that the tax merely reflects naive uniform binning, we evaluate SmallBERT with VQ $k$-means codebooks at six sizes ($K = 32, 64, 128, 256, 512, 1024$). Perturbations are applied in continuous input space and re-encoded through the learned codebook, since $k$-means labels are unordered and index-space perturbation would be semantically meaningless. The results reveal a double bind. There is a shallow optimum at $K = 64$ ($D = 0.073$), modestly better than the uniform 256-bin baseline ($D = 0.096$), but distortion then \textit{increases} with larger codebooks: $K = 512$ yields $D = 0.100$ and $K = 1024$ yields $D = 0.105$, both worse than the uniform baseline. Meanwhile, reconstruction MSE drops monotonically from $0.098$ ($K = 32$) to $0.00014$ ($K = 1024$), confirming the codebook learns well. The mechanism is straightforward: finer Voronoi cells make fixed-magnitude perturbations more likely to cross cell boundaries, so better tokenization in the reconstruction sense makes geometry worse. Empirical Procrustes distortion follows $D_{\mathrm{proc}} \propto 1/\!\log K$
($R^2 = 0.98$), far slower than the $1/K$ scaling one might naively expect
from adding codebook entries. This slow decay reflects the geometry of
boundary-crossing under perturbation rather than reconstruction quality (which
improves monotonically with $K$; see Section~\ref{sec:info-theory}) meaning exponentially more codes would be needed to approach continuous performance. No feasible codebook size escapes the tax.

\paragraph{Additional ablations.}
Four further ablations confirm the causal picture; full results for all six ablation variants appear in Appendix~\ref{app:ablation}.
\begin{itemize}
    \item Variant~B (Jacobian regularization) adds a Frobenius-norm penalty $\lambda \lVert \partial \mathbf{h}/\partial \mathbf{x} \rVert_F$ to CE loss and sweeps $\lambda \in \{10^{-3}, 10^{-2}, 10^{-1}, 1.0\}$ on both SmallBERT and SmallStripedHyena. The result is a Pareto frontier: composite stability improves monotonically with $\lambda$ (SmallBERT: $0.467 \to 0.491$; SmallStripedHyena: $0.472 \to 0.491$), but at the cost of degraded predictive accuracy (SmallBERT mean val CE: $0.81 \to 1.07$; SmallStripedHyena: $0.002 \to 0.003$). No setting achieves simultaneously low distortion and low CE, confirming the tax is a genuine trade-off intrinsic to discrete optimization, not a training artifact. 
    \item Variant~C (SmallMamba with MSE head) serves as a positive control, confirming SSM stability persists regardless of the output head and is intrinsic to the continuous ODE prior.
    \item Variant~D (SmallLSTM, 2.2M parameters, discrete CE) tests whether recurrence alone suffices for geometric stability. The LSTM exhibits composite scores comparable to SmallBERT (Lorenz mean: $0.480$ vs.\ $0.455$), not SmallMamba ($0.369$), proving that the continuous ODE parameterization, not mere recurrence, is the source of SSM stability.
    \item Variant~E (attention ratio sweep) titrates the fraction of attention layers within an 8-layer StripedHyena from 0/8 (pure Hyena) to 8/8 (pure Transformer) across 9 configurations, producing a dose-response curve for the tax.
    \item   Variant~F (Hyena filter order sweep, orders 1, 2, 4, 8) tests whether the depth of the continuous-time ODE parameterization matters within the Hyena operator. 
\end{itemize}

\subsection{Track A + Track B: The Smoking Gun}
\label{sec:tracks}

\begin{figure}[ht]
\centering
\includegraphics[width=\textwidth]{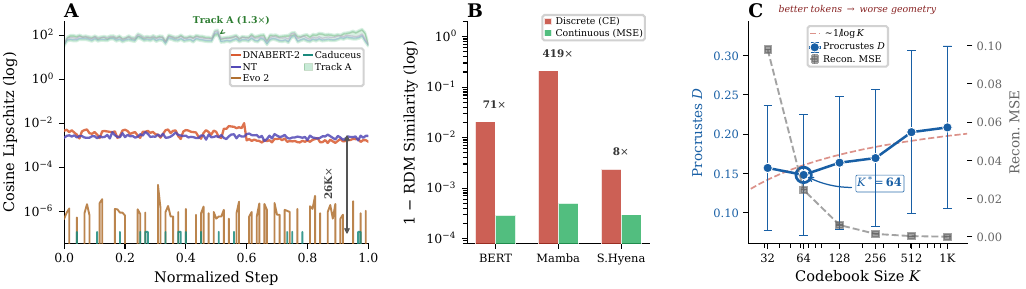}
\caption{\footnotesize\textbf{A.} Track A vs. Track B Lipschitz profiles: smooth arcs (continuous physics) vs. divergent, multi-scale fracture (discrete biology). \textbf{B.} Continuous vs. discrete Procrustes D across architectures on the Lorenz dataset at 1\% noise. All continuous conditions cluster near zero; discrete conditions span an order of magnitude. \textbf{C.} VQ double bind: reconstruction MSE (decreasing) vs. Procrustes D (non-monotone) vs. codebook size K, with ~1/log(K) fit overlaid. } 
\label{fig:ground-truth}
\end{figure}

We complement the controlled ablations with a cross-domain comparison that illustrates the magnitude of the tokenization tax in deployed models. This is not a controlled causal test (that role is served by the within-architecture ablations of Section~\ref{sec:causal-proof}), but the scale of divergence is striking.

\paragraph{Track A (continuous physics).}
We use the damped harmonic oscillator as a continuous interpolation testbed. Given two distinct oscillator trajectories, we generate 101 linearly interpolated sequences in input space ($\alpha$ from 0 to 1), embed each intermediate through all three architectures trained with MSE loss, and compute $L_2$ Lipschitz profiles, measuring the local rate of embedding change per interpolation step. The experiment averages over 10 random pairs of starting states, producing embedding PCA trajectories, cosine distance profiles, and Lipschitz profiles. All three architectures yield smooth PCA arcs with no staircase effects, no teleportation, and no fracturing. Mean Lipschitz values are $65.3$ (SmallBERT), $79.3$ (SmallStripedHyena), and $84.6$ (SmallMamba), a spread of $1.3\times$ from smoothest to roughest.

\paragraph{Track B (discrete biology).}
We construct a single-point mutation walk on the BRCA1 gene (chr17, GRCh38). Starting from wildtype, we change one base pair at a time across a 2\,kb core region, walking through 122 intermediate sequences (120 SNPs plus the pathogenic C61G missense mutation). Each sequence is embedded by four genomic foundation models spanning the architectural spectrum: DNABERT-2 (117M, BPE tokenization), Nucleotide Transformer (500M, 6-mer tokenization), Evo~2 (7B, StripedHyena hybrid, single-character tokenization), and Caduceus (7.7M, pure Mamba SSM, single-character tokenization with RC-equivariant design). We compute cosine-based Lipschitz profiles, which are dimension-invariant, to measure how much each model's embedding shifts per single-base change. Mean cosine Lipschitz spans three orders of magnitude: $3.1 \times 10^{-3}$ (DNABERT-2) and $2.6 \times 10^{-3}$ (Nucleotide Transformer) for the Transformers, versus $1.0 \times 10^{-6}$ (Evo~2) and $1.0 \times 10^{-7}$ (Caduceus) for the SSM-containing models. No model detected the pathogenic C61G mutation as a Lipschitz spike. The three failure regimes from the information-theoretic analysis (Section~\ref{sec:info-theory}) map directly onto these profiles: the Transformers fracture (high-magnitude, scattered PCA clusters), Evo~2 absorbs (smooth trajectory but biophysically numb at $10^{-6}$ per SNP), and Caduceus collapses to a flatline at the floating-point floor (18 apparent ``spikes'' are numerical jitter, not biological signal).

\paragraph{The headline.}
Track~A: $1.3\times$ gap between architectures. Track~B: ${\sim}\,3{,}000\times$ gap. The same attention mechanism that produces the smoothest interpolation on continuous signals produces a fractured manifold when forced through discrete tokens. The variable that changed between Track~A and Track~B is not the routing, it is the tokenization.

\section{Model Performance at Scale}
\label{sec:models}
\subsection{Parameter Scale vs.\ Stability}
\label{sec:scaling}

\begin{figure}[t]
  \centering
  \includegraphics[width=\textwidth]{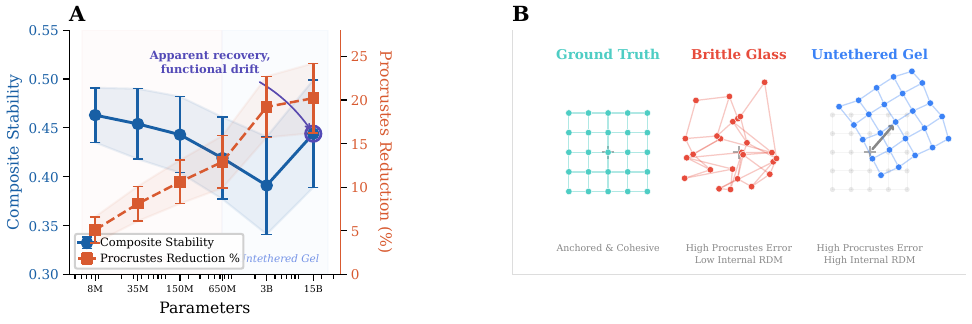}
  \caption{\footnotesize
    \textbf{A.}~ESM-2 composite stability (blue, left axis) vs.\ parameters, with Procrustes reduction overlaid (orange, right axis). Stability declines monotonically from 8M to 3B; the 15B ``recovery'' is unmasked by the simultaneous spike in Procrustes reduction, revealing global manifold drift rather than genuine geometric improvement.
    \textbf{B.}~Conceptual illustration of the two failure modes. \textit{Ground Truth}: the manifold is anchored and internally cohesive. \textit{Brittle Glass} (small/medium Transformers): high Procrustes error with low internal RDM, the manifold fractures internally. \textit{Untethered Gel} (large Transformers): high Procrustes error with high internal RDM, the manifold drifts as a coherent block.%
  }
  \label{fig:glass-gel}
\end{figure}

\paragraph{ESM-2 (the primary scaling story).}
We evaluate all six ESM-2 checkpoints (8M to 15B parameters) on 10{,}000 synthetic protein sequences (200 aa) perturbed with amino acid substitutions at 1--10\% of positions and sequence reversal. Composite stability at 1\% substitution declines monotonically from $0.463$ (8M) through $0.454$ (35M), $0.443$ (150M), $0.419$ (650M), to $0.391$ (3B), a progressive tax spanning nearly four orders of magnitude in parameters. Then the 15B checkpoint appears to ``recover'' to $0.445$ (Figure~\ref{fig:glass-gel}A, blue curve). This V-curve is misleading. We quantify global drift via a Procrustes reduction: $\rho = 1 - \lVert X_{\mathrm{clean}} - R^{*} X_{\mathrm{pert}} \rVert_F \,/\, \lVert X_{\mathrm{clean}} - X_{\mathrm{pert}}\rVert_F$, where embeddings are mean-centered and $R^{*}$ is the optimal orthogonal rotation with scaling: low $\rho$ means internal fracture that no rotation can fix, high $\rho$ means coherent drift that rotation partially removes. The 15B model achieves $\rho \approx 5\%$ at 1\% substitution, rising to ${\sim}\,20\%$ under reversal (Figure~\ref{fig:glass-gel}A, orange curve). The manifold drifts globally while preserving internal relative structure, the signature of \textit{Untethered Gel} (Figure~\ref{fig:glass-gel}B, right panel). RDM similarity alone misses 
this because pairwise cosine distances are rotation-invariant. The scaling trend is replicated on real UniRef50 proteins (10{,}000 sequences, 100--400 aa), where the same monotonic decline and illusory 15B recovery appear: 1\% substitution composite of $0.466$ (8M) to $0.397$ (3B) to $0.448$ (15B). The gap between synthetic and real composites is small (${\leq}\,0.03$ at every scale), confirming the tax is not an artifact of sequence composition.

\paragraph{Cross-architecture summary (full results in Appendix~\ref{app:model-zoo}).}
The Nucleotide Transformer (NT v2, DNA, 6-mer tokenization) exhibits the same progressive decline across four sizes: mean composite drops from $0.239$ (50M) to $0.210$ (250M) on synthetic DNA, with a partial rebound at 2.5B that parallels the ESM-2 pattern. SaProt (structure-aware protein Transformer, 35M to 1.3B) degrades with scale ($0.436 \to 0.398$), demonstrating that Foldseek structural tokens do not rescue the tax. Caduceus (pure Mamba SSM, RC-equivariant, 0.5M to 7.7M) is the tax-exempt baseline: composite stability is nearly constant across scale ($0.459$, $0.449$, $0.458$), with near-perfect RC preservation (RDM ${>}\,0.999$) on real chr22 DNA. ProtMamba (protein SSM, 108M) presents an apparent paradox: low Procrustes distortion with reasonable perplexity ($40.1$), yet, as Section~\ref{sec:info-theory} quantifies, its embeddings are informationally empty.

\paragraph{Brittle Glass vs.\ Untethered Gel.}
\label{sec:phase}
We operationally define the two failure modes by their Procrustes reduction $\rho$ under 1\% substitution. \textit{Brittle Glass} ($\rho < 2\%$): rotation cannot reduce the clean-perturbed residual, indicating internal fracture (ESM-2 8M--650M: $\rho = 0.7$--$1.8\%$). \textit{Untethered Gel} ($\rho > 4\%$): rotation substantially reduces the residual, indicating coherent global drift (ESM-2 3B--15B: $\rho = 5.2$--$5.0\%$; under reversal, $19.7$--$20.1\%$). Caduceus shows near-zero reduction ($0.3$--$0.5\%$ on SNPs), confirming minimal distortion. The Frozen Head test provides functional validation: for Evo~2 at 8K context, the fraction of tokens where the pretrained LM head produces the same top-1 prediction on perturbed vs.\ clean embeddings drops from ${\sim}\,85\%$ (1\% SNP) to ${\sim}\,28\%$ (synthetic RC), confirming that global drift is functionally destructive.

\subsection{Context Length and the RC Dissociation}
\label{sec:context-rc}

\paragraph{Evo~2 context scaling (8K, 262K, 1M).}
We evaluate the same Evo~2 7B model at three context-window checkpoints. On synthetic DNA, SNP stability gains are modest: 1\% SNP RDM similarity rises from $0.747$ (8K) to $0.817$ (1M). On real chr22 sequences the gains are marginal: $0.990$ to $0.993$. Frozen Head accuracy on the context tax test (classifying E.\ coli vs.\ human from a 1\,kbp signal region, padded to match each checkpoint's context) is $0.988$ (8K), $0.980$ (262K), $0.993$ (1M): $128\times$ more context for effectively zero geometric gain.

\begin{figure}[t]
  \centering
  \includegraphics[width=\textwidth]{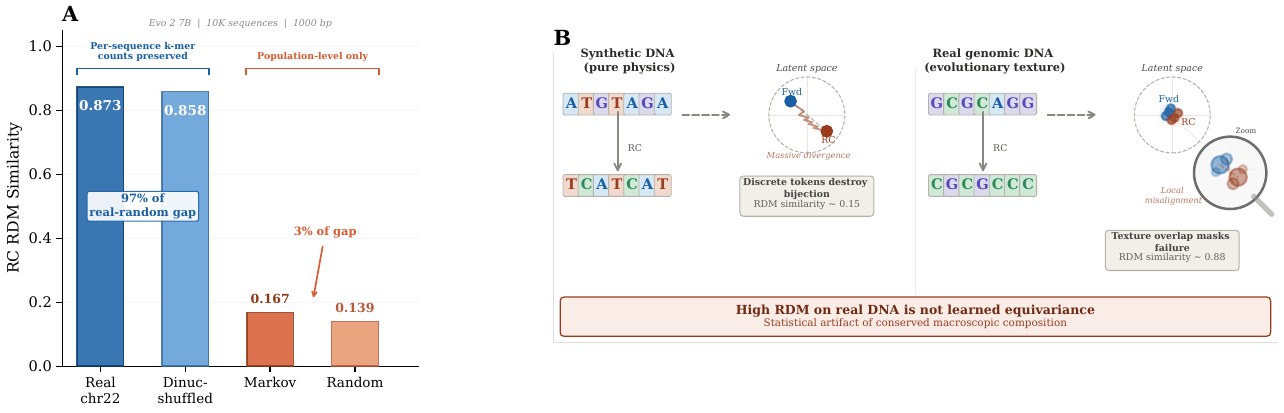}
  \caption{\footnotesize
    \textbf{A.}~Texture Hypothesis Test. RC RDM similarity across four conditions for Evo~2 (7B, 8K context, 10{,}000 sequences). Dinuc-shuffled real DNA (per-sequence $k$-mer counts preserved) recovers $97\%$ of the real-random gap; texture-matched Markov (population-level statistics only) recovers $3\%$.
    \textbf{B.}~The RC Dissociation explained. On synthetic DNA (left), discrete tokens destroy the $\mathrm{A}{\leftrightarrow}\mathrm{T}$ / $\mathrm{C}{\leftrightarrow}\mathrm{G}$ bijection entirely (RDM ${\sim}\,0.15$). On real genomic DNA (right), conserved macroscopic $k$-mer composition creates overlapping texture between forward and RC embeddings (RDM ${\sim}\,0.88$), masking the failure. High RDM on real DNA reflects statistical artifact, not learned equivariance.%
  }
  \label{fig:rc-texture}
\end{figure}

\paragraph{The Reverse Complement Dissociation.}
Due to the structural biochemistry of the double helix~\citep{Watson1953}, every strand of DNA possesses a mathematically perfect, continuous symmetry: the reverse-complement (RC). Because a sequence and its reverse-complement encode the exact same biological information, a geometrically grounded model must map both to an identical or perfectly symmetric representational manifold. Evo~2 fails this test categorically on synthetic DNA: RC RDM similarity is $0.139$ (8K), $0.156$ (262K), $0.208$ (1M). The model has functionally zero understanding of the $\mathrm{A}{\leftrightarrow}\mathrm{T}$ / $\mathrm{C}{\leftrightarrow}\mathrm{G}$ bijection (Figure~\ref{fig:rc-texture}B, left panel). Yet real RC is strikingly higher: $0.879$ (8K), $0.883$ (262K), $0.873$ (1M). To determine why, we run a controlled four-condition experiment: the Texture Hypothesis Test (10{,}000 sequences per condition, 1000\,bp, Evo~2 7B at 8K context; full details in Appendix~\ref{app:texture}).


Dinucleotide-shuffled real DNA (Altschul-Erickson algorithm: exact per-sequence $k$-mer counts preserved, all positional structure destroyed) recovers $97\%$ of the real-random RC gap in RDM similarity (Figure~\ref{fig:rc-texture}A). Texture-matched Markov sequences (first-order Markov chain calibrated to population-level dinucleotide frequencies) recover only $3\%$. The mechanism is now pinned: Evo~2's embeddings function as high-dimensional per-sequence $k$-mer histograms. RC preserves exact $k$-mer counts (every $k$-mer maps to its complementary $k$-mer at the same frequency), so forward and RC produce symmetrical histograms that the model's weights aggregate equivalently. Destroying positional structure via dinucleotide shuffling preserves this illusion because the histogram is unchanged. Matching only population-level statistics via Markov generation fails because individual sequences lose their unique compositional fingerprint, collapsing the per-sequence pairwise structure that RDM measures. This is a controlled causal result: Evo~2 does not understand double-stranded DNA symmetry. It counts short subsequences. The apparent RC ``success'' on real DNA is an artifact of a histogram encoder invariant to the one biological transformation that preserves histograms (Figure~\ref{fig:rc-texture}B, right panel).

\paragraph{Does RC regularization reduce the tax?}
To test whether post-hoc symmetry enforcement mitigates the geometric 
tax, we applied an embedding-level variant of RCCR~\citep{ma2025} to 
DNABERT-2 (117M), minimizing L2 distance between mean-pooled 
representations of forward and RC sequences during fine-tuning 
(adapting the task-level consistency objective to an unsupervised 
setting; details in Appendix~\ref{app:rccr}). RCCR achieves perfect 
per-sequence RC consistency (cosine gap: $0.041 \to 0.000$), but 
the population-level geometric structure degrades: Procrustes 
disparity between forward and RC embedding matrices increases by 
$91\%$, RC RDM similarity turns negative ($-0.036$), and SNP 
perturbation sensitivity collapses by two orders of magnitude. 
Forcing pointwise symmetry compliance flattens the embedding 
landscape rather than aligning its geometry, consistent with the 
rate-distortion prediction that capacity spent on one constraint is 
unavailable for manifold preservation. The geometric tax is not 
reducible to a missing symmetry; it is intrinsic to the discrete 
optimization landscape.

\section{The Information Theory of the Tax}
\label{sec:info-theory}
Our results across physics, biology, and scale point to a fundamental trade-off
in representational learning. We now formalize the mechanism through
rate--distortion theory and quantify its consequences with mutual information
estimation.

\subsection{Rate--Distortion Framing}
\label{sec:rd}
\paragraph{The discrete bottleneck.}
Cross-entropy over a $K$-token vocabulary optimizes for sharp partition
boundaries with no gradient signal toward manifold preservation. A next-token
predictor trained with CE is a classifier that happens to produce embeddings as
a side effect; those embeddings inherit the piecewise-constant geometry of the
decision regions, not the smooth geometry of the source. The channel capacity of a $K$-symbol discrete vocabulary is $\log_2 K$ bits per
token. For a source manifold with intrinsic dimensionality $d_M$, Shannon's
rate--distortion function for a Gaussian source under squared-error distortion
gives the minimum achievable distortion at rate $R$: $ D(R) \;=\; \sigma^2 \, 2^{-2R/d_M}$ Substituting the discrete capacity $R = \log_2 K$ yields reconstruction
distortion $D \propto K^{-2/d_M}$, the classical high-rate quantization scaling
\citep{gersho1992vector, gray1998quantization}. For practical manifold
dimensionalities ($d_M \gg 2$), this decay is extremely slow: halving
reconstruction error requires increasing the codebook by a factor of
$2^{d_M}$, which is astronomically large even for modestly high-dimensional
sources.

\paragraph{Geometric distortion is not reconstruction error.}
The distortion metric relevant to representational stability is not
reconstruction MSE but \emph{perturbation sensitivity}: given a small input
perturbation $\epsilon$, does the representation change smoothly? For a
$K$-cell Voronoi tessellation of a $d_M$-dimensional space, cell diameter
scales as $K^{-1/d_M}$. A fixed-magnitude perturbation therefore has
increasing probability of crossing a cell boundary as $K$ grows, because the
boundaries become denser. This creates the \textbf{VQ double bind} observed in
Section~2: coarse codebooks ($K$ small) lose information through quantization
noise, while fine codebooks ($K$ large) increase boundary-crossing probability
under perturbation. The shallow optimum at $K{=}64$ and subsequent distortion
increase through $K{=}1024$ (Figure~1C) are a direct consequence of this
mechanism. The empirical Procrustes distortion across our codebook sweep is well-described
by $D_{\mathrm{proc}} \propto 1/\!\log K$ ($R^2 = 0.98$, $p < 0.001$ on the
Lorenz attractor with $d_M \approx 2.06$). We emphasize that this is an
\emph{empirical scaling law for geometric distortion under perturbation}, not a
restatement of the classical $K^{-2/d}$ reconstruction bound. The two
quantities measure different things: reconstruction MSE improves monotonically
with $K$ (as our data confirm), while perturbation stability degrades once cell
boundaries become denser than the perturbation scale. The logarithmic decay of
geometric distortion with codebook size implies that exponentially more codes
would be needed to approach continuous-head performance, a cost that no
practical tokenization strategy can absorb.

\paragraph{Capacity-Induced Fracture.}
The same mechanism operates through model scale. As parameters increase, CE
training produces sharper and more numerous decision boundaries, each a
discontinuity in the embedding manifold. This is the source of the monotonic
stability decline observed in ESM-2 (8M--3B, Section~3.1): more capacity
enables finer partitioning, which creates more fracture surfaces. The apparent
stability recovery at 15B is illusory (Untethered Gel): global drift masks
local fracture rather than resolving it. \textbf{Scope.}
Our claim is that discrete tokenization with CE is a \emph{sufficient}
condition for geometric distortion, not that it is the only possible source. We test one such intervention: embedding-level RCCR~\citep{ma2025} 
applied to DNABERT-2 achieves perfect per-sequence RC consistency but 
degrades population-level geometry (Section~\ref{sec:context-rc}, 
Appendix~\ref{app:rccr}), suggesting that post-hoc symmetry 
enforcement redistributes rather than eliminates the tax. Whether 
architectural equivariance (e.g., RC-equivariant layers) can succeed 
where regularization fails remains an open question.

\subsection{The Three Pathologies of the Distortion Bound}
\label{main:conservation-regimes}
Because modern biological foundation models operate under these strict quantization limits, their attempts to minimize geometric distortion ($D$) universally result in pathological trade-offs against the mutual information $I(X; \hat{X})$ required to maintain biological utility. We categorize these failures into three regimes: \textbf{Regime I: Local-Global Decoupling}:
    The model minimizes local distortion $D$ by anchoring embeddings to short-range composition, but sacrifices the global mutual information $I(X;\hat{X})$ needed to integrate long-range structure. Geometry is preserved locally; biological coherence is lost globally. Geometrically, this manifests as the Untethered Gel signature identified in Section~\ref{sec:phase}: high Procrustes reduction indicating coherent global drift. Both large-scale ESM-2 (${\geq}\,3$B) and Evo~2 exhibit this regime. \textbf{Regime II: Representational Compression}:
    The model maximizes $I(X;\hat{X})$ by concentrating task-relevant information, but pays the full distortion cost: the manifold warps under compression, producing geometric fracture analogous to the Brittle Glass signature (Section~\ref{sec:phase}), here driven by intentional information concentration rather than capacity exhaustion. OpenFold's Evoformer is the canonical example. \textbf{Regime III: Geometric Vacuity}: The model achieves low distortion $D$ trivially, by encoding nothing. Geometry is smooth because the manifold is informationally empty: $I(X;\hat{X})$ falls below the random noise floor. Neither the Brittle Glass nor Untethered Gel geometric signatures apply, because there is no information to fracture or drift.

\begin{figure}[h]
  \centering
  \includegraphics[width=\textwidth]{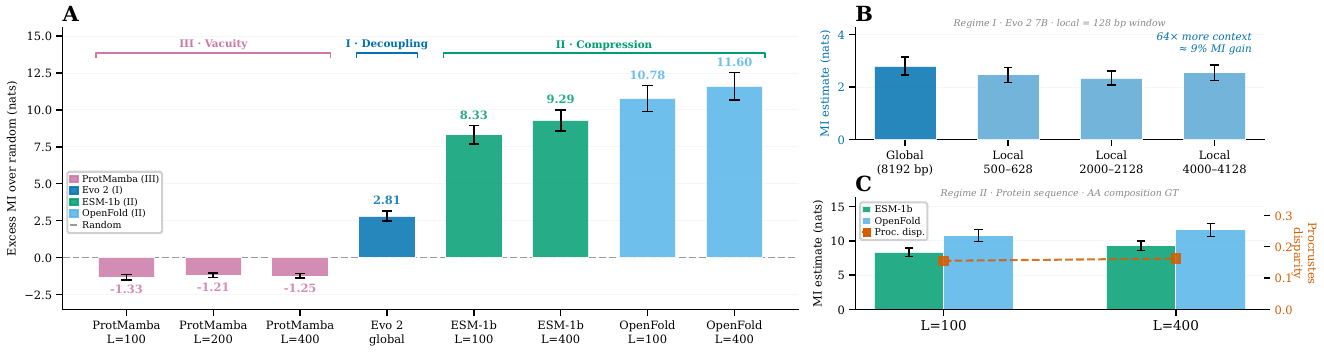}
  \caption{\footnotesize
    \textbf{(A)}~Excess MI (bias-corrected) across the three failure regimes. ProtMamba falls below zero (Geometric Vacuity), ESM-1b and OpenFold show large positive values (Representational Compression), and Evo~2 is modest and positive (Local-Global Decoupling). Random baselines sit at zero by construction.
    \textbf{(B)}~Regime~I: Evo~2 global vs.\ local MI. The flat curve across $64\times$ context expansion confirms informational shallowness.
    \textbf{(C)}~Regime~II: ESM-1b vs.\ Evoformer excess MI at each sequence length, with Procrustes disparity annotated. The Evoformer amplifies MI while warping the manifold.%
  }
  \label{fig:mine}
  
\end{figure}

\noindent The Procrustes and Lipschitz analyses of Sections~\ref{sec:ground-truth}-\ref{sec:models} measure geometric distortion; we now answer the complementary question: \textit{does the manifold contain biological information?} We quantify each regime with MINE (Mutual Information Neural Estimation) using the Donsker-Varadhan variational lower bound~\citep{Donsker1983, Belghazi2018MutualIN}. The statistics network $T_\theta$ is a three-layer MLP ($\mathrm{input} \to 256 \to 128 \to 1$, ReLU activations, 10\% dropout). For each model we run 5 independent estimations and project embeddings to 50 dimensions via PCA. Ground truth features are amino acid composition plus species labels (protein models) or GC content plus 16 dinucleotide frequencies (DNA models). All results are reported as \textit{excess MI} over a matched random baseline ($\mathrm{MI}_{\mathrm{excess}} = \mathrm{MI}_{\mathrm{model}} - \mathrm{MI}_{\mathrm{random}}$) to remove finite-sample MINE bias, which scales with PCA dimensionality and can produce spuriously large raw estimates. The random baselines are $2.664 \pm 0.054$ nats (protein) and $1.681 \pm 0.033$ nats (DNA). We validate key findings with independent frozen-head probes (linear and nonlinear); concordance between MINE and probe-based diagnostics across all three regimes increases confidence that the results are not artifacts of the estimator. Full calibration details appear in Appendix~\ref{app:mine}.
 
\paragraph{Regime I: Local-Global Decoupling (Evo~2).}
The Texture Hypothesis Test (Section~\ref{sec:context-rc}) establishes the mechanism: Evo~2's embeddings function as per-sequence $k$-mer histograms. MINE confirms the informational shallowness. Global MI (full 8{,}192-token context, mean-pooled) exceeds local MI (128-token windows) by only 14\%. A $64\times$ increase in context buys almost nothing. The excess MI is positive, confirming the embeddings do encode biological signal, but the signal is shallow and scale-invariant: local composition, not long-range structure. This is the Untethered Gel from an information-theoretic perspective, geometrically stable yet informationally provincial.
 
\paragraph{Regime II: Representational Compression (OpenFold / ESM-1b).}
The Evoformer trunk \textit{increases} MI while geometrically warping the representation. We extract embeddings from ESM-1b (the input encoder) and from the Evoformer output (after 48 blocks of structure-aware processing) and run MINE on both OpenFold exceeds ESM-1b in MI at every sequence length: the Evoformer adds $+2.3$ to $+2.5$ nats of structural context. But this comes at a geometric cost. Procrustes disparity between ESM-1b and Evoformer output representations is $0.164$ ($L{=}100$), $0.162$ ($L{=}200$), and $0.149$ ($L{=}400$), confirming substantial manifold warping. Information is retained and amplified, but the topology is broken. This is not a loss of signal; it is a geometric transformation that trades fidelity for task-specific compression. The Evoformer is an effective information concentrator, but it pays the distortion bound in full: capacity spent on structural compression is capacity unavailable for manifold preservation.
 
\paragraph{Regime III: Geometric Vacuity (ProtMamba).}
The most paradoxical finding. ProtMamba (108M parameters, Mamba SSM backbone) exhibits low Procrustes distortion and reasonable autoregressive perplexity ($40.1$). Yet, negative excess MI at every sequence length: the embeddings carry \textit{less} mutual information with biological ground truth than a matched random baseline. Stability without substance. Frozen Head probes confirm the diagnosis: both linear logistic regression and nonlinear MLP probes (two- and three-layer, up to 512 hidden units) achieve chance-level accuracy (${\sim}\,0.50$) across all sequence lengths and both local and global pooling strategies. The representation does not resist linearization because information is encoded nonlinearly; it resists because information is absent. The language model head still works (perplexity is reasonable), meaning the $\mathrm{lm\_head}$ projection extracts next-token predictions from a representational subspace that the embedding layer does not expose to downstream consumers. The internal representations are informationally empty: smooth because they encode nothing, not because they encode well.
 
\paragraph{Synthesis.}
The three regimes are not independent pathologies but different faces of the same distortion bound. 
Each model allocates finite capacity under the Gaussian rate--distortion
constraint $R(D) \geq \tfrac{d_M}{2}\log(1/D)$ (a lower bound on required
rate for any source with the same variance, by the maximum-entropy property of
the Gaussian): no discrete-token model in our study achieves simultaneously low distortion, high mutual information, and global coherence. The tax is paid in different currencies, but always paid.

\section{Discussion}
\label{sec:discussion}
\paragraph{Summary.}
Discrete tokenization is the dominant source of geometric instability in biological and physics foundation models. The tax is progressive with parameters (Section~\ref{sec:scaling}), invariant to context expansion (Section~\ref{sec:context-rc}), and empirically bounded by a $1/\!\log K$ geometric distortion scaling that makes continuous-head performance unreachable through codebook refinement alone
(Section~\ref{sec:ground-truth}). No tokenization strategy in our study escapes this scaling, and the VQ double bind (Section~\ref{sec:ground-truth}) demonstrates that finer quantization actively worsens geometric stability once cell boundaries become denser than the perturbation scale. The three failure regimes identified by MINE (Section~\ref{main:conservation-regimes}) are not independent pathologies but different allocations of finite capacity under the same constraint: models that preserve local geometry sacrifice global coherence (Regime~I), models that concentrate information sacrifice geometric fidelity (Regime~II), and models that preserve geometry sacrifice information entirely (Regime~III).

\paragraph{Implications.}
Current evaluation practice (perplexity, AUC, benchmark accuracy) is blind to the tax. A model can dominate leaderboards while its global geometry is entirely ungrounded (Regime~I), or produce smooth, stable manifolds that pass geometric consistency checks while encoding no biological signal (Regime~III). As foundation models are increasingly deployed for therapeutic design, materials discovery, and physical simulation, the field must expand its notion of reliability beyond predictive accuracy to encompass what we term \textit{Physical Alignment}: the requirement that learned representations faithfully preserve the continuous invariants of the systems they model. Geometric stability auditing via frameworks like Shesha should become a first-class evaluation criterion alongside task performance.

\paragraph{Death and Taxes.}
It's commonly been said that the only two certainties in life are death and taxes~\citep{bullock1716cobler, defoe1726devil, franklin1789leroy}. If death and taxes are constants of life, the natural question follows: Is the Geometric Alignment Tax inevitable? For generative applications like chatbots, the tax may be acceptable, or even desirable. A drifting manifold allows for creativity, where ``hallucination'' is a feature, not a bug. However, for Scientific Foundation Models, where the laws of physics are invariant and outcomes can have life-or-death consequences, this tax is unaffordable. Our results demonstrate that we cannot simply scale our way out of this penalty. The path to Scientific AGI is not merely training larger discrete models to chase an asymptotic limit, nor is it applying continuous priors that trivially satisfy geometric stability by erasing biological signal. It requires acknowledging that our current architectural playbook is fundamentally broken for the natural sciences, necessitating a return to first principles.

\paragraph{Limitations.}
Our findings are scoped to AI for Science, where ground-truth continuous 
symmetries provide hard constraints. We caution against generalizing 
the Geometric Alignment Tax to natural language, which lacks rigid, 
mathematically continuous invariants. Our analysis covers models up to 
15B parameters and 1M context length; while our rate-distortion framing 
predicts the tax persists asymptotically, we cannot rule out emergent 
mitigation at scales beyond our empirical horizon. We acknowledge that 
continuous-output formulations (diffusion, MSE regression) offer a 
theoretical alternative, and our own ablations confirm that 
continuous-head Transformers preserve geometry on chaotic attractors; 
however, these ablations are limited to synthetic physics, and 
demonstrating that continuous objectives reduce geometric distortion on 
biological sequences while retaining task performance is an important 
next step. The vast majority of deployed biological foundation models 
use discrete tokenization, and our goal is to characterize the cost of 
that dominant paradigm, not to prescribe a replacement. We characterize 
the tax under vanilla CE objectives without explicit symmetry 
constraints; an embedding-level RCCR experiment 
(Appendix~\ref{app:rccr}) shows that post-hoc RC consistency 
regularization degrades manifold geometry despite achieving perfect 
pointwise symmetry; however, this tests one instantiation at one 
hyperparameter setting. Architectural equivariance, alternative 
RCCR formulations, and broader $\lambda$ sweeps could yield 
different outcomes. Our MINE ground-truth features are compositional (amino acid 
frequencies, GC content, dinucleotides); while the ProtMamba vacuity 
finding is independently confirmed by frozen-head probes at chance 
accuracy across all conditions, models encoding higher-order structural 
features not captured by these targets could appear informationally 
shallow under our protocol. Finally, time-reversal on continuous 
trajectories is mathematically simpler than reverse-complement on 
discrete categorical sequences; our claim is that tokenization is a 
\textit{sufficient condition} for symmetry failure, not that the two 
tasks are equivalent.

\paragraph{Future directions.}
The path forward likely requires architectures that natively unify continuous geometric priors with high-fidelity discrete encoding, rather than grafting one onto the other (as ProtMamba's Geometric Vacuity demonstrates). Geometric stability auditing, continuous-valued foundation models, and hybrid objectives that jointly optimize predictive accuracy and manifold preservation are promising directions.

\section*{Code}
\label{sec:code}
The full code necessary to reproduce all experiments, benchmarks, and analysis described in this paper is publicly available at \url{https://github.com/prashantcraju/geometric-alignment-tax}. Specific details about infrastructure, software versions, and configurations that were used are outlined in Appendix~\ref{app:reproducibility}.

\begin{ack}
We thank Padma K. and Annapoorna Raju for generously supporting the computational resources used in this work. We thank the many institutions and individuals whose open-source datasets, frameworks, and models were used in our work. The authors acknowledge the use of large language models (specifically the Claude and Gemini families) to assist with code debugging and text polishing. All hypotheses, experimental designs, analyses, and interpretations were independently formulated and verified by the authors, and the authors assume full responsibility for all content and claims in this work.
\end{ack}

\bibliographystyle{apalike}  
\bibliography{references} 

\newpage
\appendix

\vbox{%
\hsize\textwidth
\linewidth\hsize
\vskip 0.1in
\centering
{\LARGE\bf
Appendix
\par}
\vskip 0.29in
  \vskip -\parskip
  \hrule height 1pt
  \vskip 0.09in%
}

\section*{Contents}

\label{app:appendix}
\begin{itemize}
    \item Appendix~\ref{app:related-work}: Related Work
    \item Appendix~\ref{app:extraction}: Extraction Details and Evaluation Metrics
        \begin{itemize}
        \item Appendix~\ref{app:stability_metrics}: Geometric Stability Evaluation Metrics
    \end{itemize}
    \item Appendix~\ref{app:a+b}: Track A + Track B Extended Methods and Details
        \begin{itemize}
        \item Appendix~\ref{app:track-a}: Track A: Synthetic Physics (Damped Harmonic Oscillator)
        \item Appendix~\ref{app:track-b}: Track B: Biological DNA (BRCA1 Mutation Walk)
        \item Appendix~\ref{app:cross-track}: Cross-Track Comparison
    \end{itemize}
    \item Appendix~\ref{app:ablation}: Complete Ablation Battery
    \begin{itemize}
        \item Appendix~\ref{app:ablation-a}: Variant A: Continuous MSE Head
        \item Appendix~\ref{app:ablation-b}: Variant B: Jacobian Norm Penalty
        \item Appendix~\ref{app:ablation-c}: Variant C: SmallMamba + Continuous Head (Positive Control)
        \item Appendix~\ref{app:ablation-d}: Variant D: LSTM Baseline (The Discrete Recurrence Test)
        \item Appendix~\ref{app:ablation-e}: Variant E: Attention Layer Ratio Sweep (The Dilution Curve)
        \item Appendix~\ref{app:ablation-f}: Variant F: Hyena Filter Order Sweep (The ODE Depth Test)
    \end{itemize}
    \item Appendix~\ref{app:model-zoo}: Full Model Zoo
    \item Appendix~\ref{app:ghost-detection}: Ghost Detection and Phase Transition Extended Methods and Details
    \item Appendix~\ref{app:rccr}: RCCR Experiment Details
    \item Appendix~\ref{app:mine}: MINE Extended Methods and Details
    \item Appendix~\ref{app:texture}: Texture Hypothesis Test (Evo 2 Reverse Complement Mechanism)
    \item Appendix~\ref{app:reproducibility}: Reproducibility and Computational Infrastructure
\end{itemize}

\newpage

\section{Related Work}
\label{app:related-work}
\paragraph{Representation geometry and comparison metrics.}
Representational Similarity Analysis (RSA, \citealp{Kriegeskorte2008}) introduced the use of pairwise dissimilarity matrices to compare neural representations, and Centered Kernel Alignment (CKA, \citealp{kornblith2019similarity}) extended this to kernel-space comparisons across layers and architectures. Linear probing~\citep{Alain2016UnderstandingIL} evaluates downstream discriminative utility but remains blind to manifold topology. These methods measure \emph{similarity between representations}; our work asks a different question: does a \emph{single} representation preserve the continuous geometry of its source under perturbation? The Shesha stability framework~\citep{raju2026geometric, shesha2026}, built on representational dissimilarity matrices from the RSA tradition, isolates this structural rigidity directly, decoupling local memorization from macroscopic geometric fidelity. We supplement this with Procrustes analysis~\citep{Schnemann1966,Rohlf1990,Masarotto2018,dryden1998statistical} to disentangle internal fracture from global manifold drift (Section~\ref{sec:phase}).

\paragraph{Quantization theory and rate-distortion bounds.}
Classical high-rate quantization theory~\citep{gersho1992vector,gray1998quantization} establishes that reconstruction distortion for a $K$-cell vector quantizer on a $d$-dimensional source scales as $K^{-2/d}$, with the Shannon rate-distortion function~\citep{shannon1959coding, Thomas_M_Cover2006-nr} providing the fundamental lower bound. Our contribution is to distinguish \emph{geometric distortion under perturbation} from reconstruction error: the former measures representational stability rather than encoding fidelity, and we show empirically that it follows a qualitatively different (slower, $1/\!\log K$) scaling driven by boundary-crossing dynamics in the Voronoi tessellation (Section~\ref{sec:rd}).

\paragraph{Foundation models for biology and physics.}
The application of foundation models to natural sciences spans protein language models~\citep{Jumper2021, Abramson2024, Lin2023}, genomic sequence models~\citep{DallaTorre2024, alphagenome}, and time-series forecasters for chaotic systems~\citep{zhang2025zeroshot}. Architectural evolution from pure Transformers toward long-context convolutions~\citep{Nguyen2023HyenaDNALG} and continuous-time state space models~\citep{gu2024mamba, mamba2, schiff2024caduceus} reflects growing recognition that discrete-token architectures create friction with continuous scientific ground truth. Our work formalizes and quantifies this friction.

\paragraph{Symmetry enforcement: equivariance vs.\ regularization.}
Two strategies exist for enforcing physical symmetries in discrete-token models. \emph{Architectural equivariance} builds the symmetry into the model structure: Caduceus~\citep{schiff2024caduceus} uses RC-equivariant Mamba layers, and Frame Averaging~\citep{huang2024} enforces geometric semantics in equivariant Transformers. \emph{Post-hoc regularization} penalizes symmetry violations during training: RCCR~\citep{ma2025} adds a consistency loss between forward and reverse-complement predictions across multiple backbones, while Jacobian and Hessian norm penalties~\citep{jukic2024} and spectral weight constraints~\citep{newhouse2025, khromov2024some} bound the Lipschitz constant to enforce local smoothness. Both strategies illustrate the Geometric Alignment Tax: discrete architectures require external intervention to approximate the continuous structural prior that SSMs possess natively. Our results (Section~\ref{sec:rd}) suggest this tax is intrinsic to 
vanilla CE objectives. An embedding-level RCCR experiment on 
DNABERT-2 confirms that post-hoc consistency regularization achieves 
pointwise symmetry compliance at the cost of population-level 
geometric degradation (Appendix~\ref{app:rccr}); whether 
architectural equivariance can succeed where regularization fails 
remains open.
\newpage

\section{Extraction Details and Evaluation Metrics}
\label{app:extraction}

\begin{table}[H]
\centering
\caption{Embedding extraction protocol for all evaluated models. 
All models use cosine distance for RDM computation and mean-pooling 
over a center window unless noted otherwise.}
\label{tab:extraction}
\begin{tabular}{@{}llllr@{}}
\toprule
Model & Layer & Pooling & Tokenization & Dim \\
\midrule
ESM-2 (8M--15B)       & Final hidden    & Mean-pool & Per-residue     & 320--5120 \\
ESM-1b                & Final hidden    & Mean-pool & Per-residue     & 1280 \\
OpenFold (Evoformer)  & Evoformer output & Mean-pool & Per-residue     & 384 \\
ProtMamba             & Final hidden    & Mean-pool & Per-residue     & 1024 \\
SaProt (35M--1.3B)    & Final hidden    & Mean-pool & Foldseek+residue& 480--1280 \\
DNABERT-2             & Final hidden    & Mean-pool & BPE             & 768 \\
Nucleotide Trans.     & Final hidden    & Mean-pool & 6-mer           & 512--2560 \\
Evo~2 (7B)            & \texttt{blocks.28.mlp.l3} & Mean-pool & Single-char & 4096 \\
Caduceus (0.5M--7.7M) & Final hidden    & Mean-pool & Single-char     & 118--256 \\
HyenaDNA              & Final hidden    & Mean-pool & Single-char     & 128--256 \\
GPN                   & Final hidden    & Mean-pool & Single-char     & 512 \\
\midrule
SmallBERT             & Final hidden    & Mean-pool & 256-bin uniform & 256 \\
SmallMamba            & Final hidden    & Mean-pool & 256-bin uniform & 256 \\
SmallStripedHyena     & Final hidden    & Mean-pool & 256-bin uniform & 256 \\
\bottomrule
\end{tabular}
\end{table}

\subsection{Geometric Stability Evaluation Metrics}
\label{app:stability_metrics}

To formally characterize the robustness of the latent embedding manifolds when exposed to perturbations, we utilize a comprehensive evaluation harness built upon the Shesha geometric stability framework~\citep{raju2026geometric}, which was implemented with the \texttt{shesha-geometry} pypi framework~\citep{shesha2026}. For a given set of clean and perturbed sequence embeddings, the harness extracts representation geometries over equivalent context windows, optionally stratified and bootstrapped (e.g., maintaining upper limits of 2,500 samples to manage boundary memory overhead). The evaluation suite logs the following structured stability profiles for each perturbation sequence:

\begin{itemize}
    \item \textbf{Sample Split ($\text{Shesha}_{\text{SS}}$):} Evaluates whether sample identity preserves proximity metrics accurately across perturbations:
    \begin{equation}
        \text{Shesha}_{\text{SS}}(\mathbf{X}) = \frac{1}{K} \sum_{k=1}^{K} \rho_s\bigl(\text{vec}(\mathbf{D}_{\mathcal{S}_k^{(1)}}), \text{vec}(\mathbf{D}_{\mathcal{S}_k^{(2)}})\bigr)
        \label{eq:shesha-sample-split}
    \end{equation}
    where $\mathcal{S}_k^{(1)}, \mathcal{S}_k^{(2)} \subset \{1, \dots, n\}$ are disjoint random subsets of sample indices drawn at split $k$, $\mathbf{D}_{\mathcal{S}} \in \mathbb{R}^{|\mathcal{S}| \times |\mathcal{S}|}$ is the pairwise distance matrix computed over samples in $\mathcal{S}$, and $\rho_s$ denotes Spearman's rank correlation over the vectorized upper triangles.

    \item \textbf{Feature Split ($\text{Shesha}_{\text{FS}}$):} Tracks dimension-wise correlations ensuring variance directions remain consistent across independent feature subspaces:
    \begin{equation}
        \text{Shesha}_{\text{FS}}(\mathbf{X}) = \frac{1}{K} \sum_{k=1}^{K} \rho_s\bigl(\text{vec}(\mathbf{D}_{\mathcal{F}_k^{(1)}}), \text{vec}(\mathbf{D}_{\mathcal{F}_k^{(2)}})\bigr)
        \label{eq:shesha-feature-split}
    \end{equation}
    where $\mathcal{F}_k^{(1)}, \mathcal{F}_k^{(2)} \subset \{1, \dots, d\}$ are disjoint random subsets of feature dimensions drawn at split $k$, $\mathbf{D}_{\mathcal{F}} \in \mathbb{R}^{n \times n}$ is the pairwise sample distance matrix computed using only the features in $\mathcal{F}$, and $\rho_s$ denotes Spearman's rank correlation over the vectorized upper triangles.

    \item \textbf{RDM Similarity ($\rho_{\text{RDM}}$):}~\citep{Kriegeskorte2008} Quantifies preservation testing representational topologies directly evaluating Spearman correlations evaluating off-diagonal mapping bounds natively constraints generating specifically limits explicitly correlating $X_c$ and $X_p$.

    \item \textbf{Anchor Stability ($\text{Shesha}_{\text{anchor}}$):} Measures consistency of distance profiles from fixed anchor points to random data splits, ensuring geometric relationships remain stable under resampling:
    \begin{equation}
        \text{Shesha}_{\text{anchor}}(\mathbf{X}) = \frac{1}{K} \sum_{k=1}^{K} \rho_s\!\left(\text{vec}\!\left(\mathbf{D}(\mathbf{A},\, \mathcal{S}_k^{(1)})\right),\; \text{vec}\!\left(\mathbf{D}(\mathbf{A},\, \mathcal{S}_k^{(2)})\right)\right)
        \label{eq:shesha-anchor-stability}
    \end{equation}
    where $\mathbf{A} \subset \mathbf{X}$ is a fixed set of $m$ anchor points, $\mathcal{S}_k^{(1)}, \mathcal{S}_k^{(2)} \subset \mathbf{X} \setminus \mathbf{A}$ are disjoint random subsets of size $n$, $\mathbf{D}(\mathbf{A}, \mathcal{S}) \in \mathbb{R}^{m \times n}$ is the pairwise distance matrix from anchors to subset samples (optionally rank-normalized row-wise), and $\rho_s$ denotes Spearman's rank correlation.
        
    \item \textbf{Composite Stability Score:} To generate a unified metric for model comparison spanning discrete topological spaces mapping target dimensions smoothly precisely pooling results directly uniformly evaluating:
    \begin{equation}
        \text{Composite} = \frac{1}{4} \left( \text{Shesha}_{\text{SS}} +\text{Shesha}_{\text{FS}} + \rho_{\text{RDM}} + \text{Shesha}_{\text{anchor}} \right)
    \end{equation}
\end{itemize}

\newpage

\section{Track A + Track B Extended Methods and Details}
\label{app:a+b}

This appendix provides full experimental details for the dual-track manifold continuity
test described in Section~\ref{sec:tracks}. Track~A serves as a continuous-input control; Track~B
applies the same geometric probes to four discrete-token genomic foundation models.
All experiments use seed 320 throughout.

\subsection{Track A: Synthetic Physics (Damped Harmonic Oscillator)}
\label{app:track-a}

\paragraph{Dataset.}
Each training sequence is a length-512 discretized trajectory of a damped harmonic
oscillator \[x(t) = A\,e^{-\gamma t}\cos(\omega t + \phi)\] sampled at 512 equally
spaced points over $t \in [0, 4)$. Parameters are drawn independently per sequence:
 \[ \textrm{amplitude: } A \sim \mathcal{U}(0.5, 2.0)\] 
 \[\textrm{damping: } \gamma \sim \mathcal{U}(0.2, 2.0)\]
 \[ \textrm{angular frequency: } \omega \sim \mathcal{U}(2.0, 20.0) \]
 \[\textrm{phase: }  phi \sim \mathcal{U}(0, 2\pi)\]
Continuous values are discretized into $N_{\text{bins}} = 256$ integer bins via
dataset-global min/max normalization (computed once over the full training set and
reused for all evaluation and interpolation data). The training set contains 50{,}000
sequences; a held-out validation set contains 2{,}000 sequences drawn with a separate
seed.

\paragraph{Architectures.}
Three architectures are trained under identical conditions, each with approximately
2M parameters:

\begin{table}[h]
\centering
\small
\begin{tabular}{@{}lcccccc@{}}
\toprule
Model & Layers & $d_{\text{model}}$ & Heads & FFN dim & SSM $d_{\text{state}}$ & Params \\
\midrule
SmallBERT (Transformer)       & 4 & 256 & 4 & 1024 & --  & ${\sim}$2M \\
SmallMamba (SSM)               & 4 & 256 & -- & --   & 16  & ${\sim}$2M \\
SmallStripedHyena (Hybrid)     & 4 & 256 & 4  & 1024 & --  & ${\sim}$2M \\
\bottomrule
\end{tabular}
\caption{Track~A architecture specifications. All models share $d_{\text{model}}=256$,
vocabulary size 258 (256 bins + mask + pad tokens), and sequence length 512.
SmallMamba uses $d_{\text{conv}}=4$, expand factor 2.
SmallStripedHyena alternates Hyena convolution operators (order 2, implicit MLP filter)
with standard multi-head attention layers in a 3:1 ratio (3 Hyena, 1 attention).}
\label{tab:track-a-arch}
\end{table}

\noindent
SmallBERT uses pre-LayerNorm Transformer encoder layers with learned positional
embeddings and Xavier uniform initialization.
SmallMamba stacks native CUDA Mamba blocks (\texttt{mamba\_ssm}) with pre-LayerNorm
residual connections and learned positional embeddings.
SmallStripedHyena implements a simplified version of the StripedHyena architecture
used in Evo~2: each Hyena operator learns its long convolution filter implicitly
via a small MLP over sinusoidal positional features, with exponential decay modulation
and short depthwise convolution gating. The attention stripes use the same multi-head
attention as SmallBERT.

\paragraph{Training.}
All models are trained with causal language modeling (CLM) using cross-entropy loss
for 20 epochs, batch size 64, AdamW optimizer ($\text{lr} = 3 \times 10^{-4}$,
weight decay $= 0.01$). Training runs on a single NVIDIA A100 GPU. The continuous
physical signal is discretized \emph{before} being fed to the model; the loss operates
on discrete token predictions. This design is intentional: the models receive
identical discrete inputs and are evaluated solely on the geometry of their
intermediate representations, isolating architecture from training objective.

\paragraph{Interpolation protocol.}
For each of 10 randomly sampled pairs of oscillator trajectories $(A_i, B_i)$,
we generate 101 linearly interpolated sequences:
\begin{equation}
    \text{interp}(\alpha) = \text{discretize}\bigl((1-\alpha)\cdot A_i + \alpha \cdot B_i,\; \text{global\_range}\bigr),
    \quad \alpha \in \{0, 0.01, 0.02, \ldots, 1.0\}.
\end{equation}
Both endpoints are discretized using the same dataset-global min/max range from
training, ensuring that linear interpolation in token space corresponds to linear
interpolation in physical space. Trajectory~$A$ is drawn with low frequency and
light damping ($\gamma \sim \mathcal{U}(0.2, 0.8)$, $\omega \sim \mathcal{U}(2, 8)$);
Trajectory~$B$ with higher frequency and heavier damping
($\gamma \sim \mathcal{U}(1.0, 2.0)$, $\omega \sim \mathcal{U}(10, 20)$), guaranteeing
distinct dynamical regimes at the walk endpoints.

\paragraph{Embedding extraction and metrics.}
Each interpolation step is embedded by passing the token sequence through the model
with \texttt{return\_hidden=True} and mean-pooling the last hidden layer across all
512 positions, yielding a single $d_{\text{model}}$-dimensional vector per step.
We compute three quantities per pair:

\begin{enumerate}[leftmargin=*,itemsep=2pt]
    \item \textbf{PCA trajectory}: The 101 embeddings are projected into 3D via PCA
    (fit per architecture) and plotted as a continuous path. Smooth arcs indicate a
    well-behaved manifold; staircase patterns indicate phase transitions.

    \item \textbf{Cosine distance from start}: \[d_{\cos}(\alpha) =
    1 - \frac{\mathbf{e}(\alpha) \cdot \mathbf{e}(0)}
    {\|\mathbf{e}(\alpha)\|\,\|\mathbf{e}(0)\|}\]
    for each step, measuring cumulative drift. A monotonically increasing profile
    indicates directional consistency.

    \item \textbf{L2 Lipschitz profile}: \[L(\alpha_i) = \|\mathbf{e}(\alpha_{i+1})
    - \mathbf{e}(\alpha_i)\|_2\] for consecutive steps, measuring local rate of change.
    The mean and maximum over all 10 pairs are reported. A flat profile indicates
    uniform sensitivity; spikes indicate discontinuities.
\end{enumerate}

\paragraph{Results summary.}
Mean Lipschitz values averaged over 10 pairs: SmallBERT 65.3, SmallStripedHyena 79.3,
SmallMamba 84.6 (1.3$\times$ spread from smoothest to roughest). All three architectures
produce smooth PCA arcs with no staircase effects, no teleportation, and no fracturing.
The smoothness ratio (mean/max Lipschitz) is 0.425 (SmallBERT), 0.433
(SmallStripedHyena), 0.388 (SmallMamba).

\subsection{Track B: Biological DNA (BRCA1 Mutation Walk)}
\label{app:track-b}

\paragraph{Genomic region.}
We target the BRCA1 gene on chromosome~17 (GRCh38/hg38), centered on the pathogenic
C61G missense variant at position 43{,}104{,}121. A 16{,}384\,bp region is downloaded
from the UCSC Genome Browser API \citep[\texttt{api.genome.ucsc.edu}]{Casper2025}, providing 7{,}192\,bp
of flanking context on each side of the 2{,}000\,bp core mutation zone. Any ambiguous
bases (N) are replaced with a uniformly random nucleotide (seed 320). Models with
shorter context windows receive the relevant sub-region centered on the core zone
(see per-model details below).

\paragraph{Mutation walk construction.}
The walk endpoint (mutant sequence) is constructed by introducing 121 point mutations
into the core 2\,kb region of the wildtype sequence: one pathogenic C61G substitution
at the region center, plus 120 additional random SNPs at positions sampled uniformly
without replacement from the core zone (seed 320). Each SNP changes the reference
base to a uniformly random alternative. The single-point mutation walk then proceeds
from wildtype to mutant by changing one base at a time, in a randomly shuffled order
of the 121 differing positions (also seed 320), producing 122 intermediate sequences
(including start and end). The pathogenic C61G mutation falls at a random position
in this shuffled order, providing a biologically meaningful landmark. The walk, step
positions, and pathogenic step index are cached as a NumPy archive and shared across
all four model-specific notebooks.

\paragraph{Models.}
Four genomic foundation models span the architectural spectrum from pure Transformer
to pure SSM:

\begin{table}[h]
\centering
\small
\resizebox{\textwidth}{!}{%
\begin{tabular}{@{}llllrl@{}}
\toprule
Model & Architecture & Tokenization & Embed dim & Params & Embedding layer \\
\midrule
DNABERT-2       & BERT Transformer        & BPE        & 768  & 117M  & Last hidden, mean-pool \\
Nucleotide Transformer v2 & BERT Transformer & 6-mer & 1024 & 500M  & Last hidden, mean-pool \\
Evo~2           & StripedHyena (SSM+Attn) & Single-char & 4096 & 7B   & \texttt{blocks.28.mlp.l3}, mean-pool \\
Caduceus (PS)   & Mamba SSM (RC-equiv.)   & Single-char & 256  & 7.7M  & Hidden state, mean-pool \\
\bottomrule
\end{tabular}
}
\caption{Track~B model specifications.}
\label{tab:track-b-models}
\end{table}

\paragraph{Per-model embedding details.}

\textit{DNABERT-2} (117M; \texttt{zhihan1996/DNABERT-2-117M}). Uses BPE tokenization
with a learned multi-granularity merge table. The 512-token context limit covers
approximately 1--2\,kb of DNA depending on merge patterns. We embed the 2\,kb core
region, extracting the last hidden state and mean-pooling over all non-padding tokens.
Loading requires bypassing \texttt{AutoModel.from\_pretrained} on transformers~5.x
due to meta-tensor initialization conflicts with DNABERT-2's custom \texttt{bert\_layers.py};
we resolve the model class via HuggingFace dynamic module utilities and load the
state dict directly. FlashAttention is provided via Triton.

\textit{Nucleotide Transformer v2} (500M; \texttt{InstaDeepAI/nucleotide-transformer-v2-500m-multi-species}).
Uses fixed 6-mer tokenization with a maximum of 1{,}000 tokens (${\sim}$6{,}000\,bp effective context).
We load via \texttt{AutoModelForMaskedLM} (not \texttt{AutoModel}) because the checkpoint
includes the language model head weights. Embeddings are the last hidden state
(\texttt{output\_hidden\_states=True}), mean-pooled over non-padding positions.
Extensive transformers~5.x compatibility patches are applied
(\texttt{find\_pruneable\_heads\_and\_indices}, \texttt{all\_tied\_weights\_keys},
\texttt{get\_head\_mask}).

\textit{Evo~2} (7B; \texttt{evo2\_7b} via Vortex). A StripedHyena hybrid model with
single-character tokenization and an 8{,}192\,bp context window. Loaded via
\texttt{Evo2(`evo2\_7b')} using the native Vortex interface (not HuggingFace).
Embeddings are extracted from an intermediate layer (\texttt{blocks.28.mlp.l3})
using \texttt{return\_embeddings=True} with explicit \texttt{layer\_names}, then
mean-pooled across all positions. Sequences shorter than 8{,}192\,bp are padded
with N characters; the full 16\,kb flanked region is truncated to context length.
Requires NVIDIA A100 GPU (${\sim}$28\,GB VRAM), flash-attn 2.8.0.post2 built from source.

\textit{Caduceus} (7.7M; \texttt{kuleshov-group/caduceus-ps\_seqlen-131k\_d\_model-256\_n\_layer-16}).
A pure Mamba SSM with reverse-complement (RC) equivariant design (PS variant).
Uses single-character tokenization and supports up to 131{,}072\,bp context.
Loading requires patching transformers for compatibility with the custom
\texttt{mamba\_rev}/\texttt{mamba\_fwd} weight tying scheme. Fused Triton layer
norms are disabled if unavailable. For PS models, the hidden state is pooled with
RC-invariant averaging. Built from source using \texttt{causal-conv1d} and
\texttt{mamba-ssm} packages.

\paragraph{Lipschitz profile computation.}
For each model, we compute the cosine-based local Lipschitz constant between
consecutive walk steps:
\begin{equation}
    L_{\cos}(i) = 1 - \frac{\mathbf{e}_{i+1} \cdot \mathbf{e}_i}
    {\|\mathbf{e}_{i+1}\| \, \|\mathbf{e}_i\|},
    \quad i = 0, 1, \ldots, 120,
\end{equation}
where $\mathbf{e}_i$ is the mean-pooled embedding of the $i$-th walk sequence.
We use cosine distance rather than L2 to ensure dimension-invariant comparisons
across models with embedding dimensions ranging from 256 (Caduceus) to 4{,}096
(Evo~2). Spikes are defined as steps exceeding the mean + 2 standard deviations
of the profile for that model.

\paragraph{PCA trajectories.}
The 122 embeddings per model are projected to 3 dimensions via PCA (fit independently
per model) and plotted as continuous paths. The pathogenic C61G step is marked if it
corresponds to a Lipschitz spike.

\paragraph{Results summary.}
Mean cosine Lipschitz values: DNABERT-2 $3.1 \times 10^{-3}$,
Nucleotide Transformer $2.6 \times 10^{-3}$, Evo~2 $1.0 \times 10^{-6}$,
Caduceus ${\sim}1.0 \times 10^{-7}$. The Transformers exhibit 5 and 4 spikes
respectively (above 2$\sigma$ threshold). Evo~2 shows 2 spikes. Caduceus registers
18 apparent spikes, but all are at the floating-point noise floor
(${\sim}10^{-7}$) and represent numerical jitter rather than biological signal.
No model detected the pathogenic C61G mutation as a Lipschitz spike.

\subsection{Cross-Track Comparison}
\label{app:cross-track}

The critical comparison: in Track~A (continuous physics, no tokenization),
the gap between the smoothest and roughest architecture is $1.3\times$
(SmallBERT at 65.3 vs.\ SmallMamba at 84.6). In Track~B (discrete biology,
full tokenization), the gap between the highest and lowest mean cosine Lipschitz
is approximately $3{,}000\times$ ($3.1 \times 10^{-3}$ for DNABERT-2 vs.\
${\sim}1.0 \times 10^{-6}$ for Evo~2; or ${\sim}26{,}000\times$ if comparing
DNABERT-2 to the Caduceus noise floor). The same attention mechanism that
produces the smoothest interpolation on continuous signals produces a fractured
manifold when forced through discrete tokens. The variable that changed between
the two tracks is not the routing mechanism or model scale; it is the tokenization.

\newpage

\section{Complete Ablation Battery}
\label{app:ablation}
All ablations use the same three synthetic dynamical system datasets (waveform, coupled oscillator, Lorenz attractor), 256-bin discretization, seed 320, and Shesha Procrustes stability harness as the baseline experiments in Section~\ref{sec:ground-truth}. Perturbation suite: value noise at 1/2/5/10\% of positions plus time reversal. Architectures are parameter-matched: SmallBERT (Transformer, 3.4M), SmallMamba (SSM, 2.0M), SmallStripedHyena (hybrid, 4.5M), SmallLSTM (discrete recurrent, 2.2M). All models are trained from scratch for 20 epochs on Google Colab (A100).

\subsection{Variant A: Continuous MSE Head}
\label{app:ablation-a}
\paragraph{Hypothesis.} The Geometric Alignment Tax is architectural, arising from the softmax attention mechanism, and cannot be eliminated by changing the output head alone.

\paragraph{Method.} We replace the 256-class categorical cross-entropy (CE) head with a linear projection to a continuous scalar trained under MSE loss. The encoder backbone (self-attention layers, positional embeddings, feedforward blocks) is unchanged. A dual-return generator provides both discrete 256-bin tokens (model input) and raw continuous ODE floats (MSE target), avoiding the fatal data-leakage where dividing discrete tokens by $(n_\text{bins} - 1)$ creates a quantized staircase target. All three architectures (SmallBERT, SmallMamba, SmallStripedHyena) are evaluated under both CE baseline and continuous MSE conditions.

\paragraph{Results.} Replacing discrete CE with continuous MSE eliminates manifold fracture across all architectures. On the Lorenz dataset at 1\% noise, SmallBERT Procrustes distortion $D$ improves $2.8\times$ ($0.096 \to 0.034$); SmallStripedHyena improves $8.5\times$ ($0.072 \to 0.0085$), the single best condition in the entire study. The cross-architecture spread collapses from $0.072$--$0.157$ under discrete CE to $0.0085$--$0.034$ under continuous MSE. At 10\% noise the pattern holds: SmallBERT discrete $D = 0.316$ vs.\ continuous $D = 0.111$ ($2.8\times$); SmallStripedHyena discrete $D = 0.233$ vs.\ continuous $D = 0.027$ ($8.6\times$). Mean composite stability across all datasets and perturbations rises from $0.466$ (SmallBERT discrete) to $0.497$ (SmallBERT continuous), and from $0.470$ (SmallStripedHyena discrete) to $0.498$ (SmallStripedHyena continuous). RDM similarity scores under the continuous head exceed 0.99 at 1\% noise for all three architectures, confirming near-perfect manifold preservation. The discrete-to-continuous gap within any single architecture dwarfs the cross-architecture gap under either regime, confirming that the tokenization boundary -- not the routing mechanism -- is the dominant source of geometric instability.

\subsection{Variant B: Jacobian Norm Penalty}
\label{app:ablation-b}
\paragraph{Hypothesis.} Explicit smoothness regularization cannot close the geometric stability gap without catastrophic predictive collapse, revealing the tax as a fundamental trade-off.

\paragraph{Method.} We add a Frobenius-norm penalty on the Jacobian of hidden states with respect to input embeddings: $\mathcal{L}_\text{total} = \mathcal{L}_\text{CE} + \lambda \|\partial \mathbf{h} / \partial \mathbf{x}\|_F$. The penalty is swept across $\lambda \in \{10^{-3}, 10^{-2}, 10^{-1}, 1.0\}$ on both SmallBERT and SmallStripedHyena. SmallStripedHyena additionally includes a $\lambda = 0$ (unregularized) baseline for direct comparison. All other training hyperparameters remain identical to the baseline.

\paragraph{Results.} The sweep reveals a Pareto frontier between geometric stability and predictive accuracy. For SmallBERT, mean composite stability improves monotonically with $\lambda$: $0.467$ ($\lambda = 10^{-3}$) $\to$ $0.474$ ($10^{-2}$) $\to$ $0.484$ ($10^{-1}$) $\to$ $0.491$ ($1.0$). However, mean validation CE degrades in parallel: $0.81 \to 0.85 \to 0.91 \to 1.07$. SmallStripedHyena shows the same pattern: composite stability rises from $0.472$ (baseline $\lambda=0$) to $0.491$ ($\lambda=1.0$), while validation CE increases from $0.002$ to $0.003$. At $\lambda = 1.0$, both architectures converge to nearly identical composite scores ($\sim$$0.491$), but at the cost of substantially degraded prediction quality. No setting of $\lambda$ achieves simultaneously low distortion and low CE. The Jacobian penalty smooths the manifold by penalizing sharp representational gradients, but the model compensates by flattening its hidden-state landscape, which destroys the fine-grained distinctions needed for accurate next-token prediction. This confirms the tax is a genuine trade-off intrinsic to discrete optimization under attention, not a training artifact that can be patched with regularization.

\subsection{Variant C: SmallMamba + Continuous Head (Positive Control)}
\label{app:ablation-c}
\paragraph{Hypothesis.} Mamba's geometric stability is intrinsic to its continuous ODE prior and persists regardless of the output head.

\paragraph{Method.} SmallMamba receives the same MSE head replacement as Variant~A, while keeping the SSM backbone (selective state-space mechanism with exponential matrix discretization) unchanged. SmallBERT\_Continuous and SmallStripedHyena\_Continuous are included as architecture comparisons. All three architectures use identical continuous MSE training on the same datasets.

\paragraph{Results.} SmallMamba\_Continuous maintains RDM similarity $>$0.999 across all noise levels on all three datasets, with composite stability scores of $0.496$ (waveform mean), $0.495$ (oscillator mean), and $0.500$ (Lorenz mean). These scores are comparable to baseline SmallMamba under discrete CE ($\sim$$0.400$ composite, but with RDM $>$0.99 at low noise), confirming that the SSM's geometric smoothness is independent of the loss landscape. Critically, SmallBERT\_Continuous, despite the continuous head, still exhibits lower RDM similarity than SmallMamba\_Continuous under time reversal on the waveform dataset ($0.978$ vs.\ $1.000$) and oscillator dataset ($0.928$ vs.\ $1.000$). SmallStripedHyena\_Continuous shows intermediate behavior, with sensitivity to time reversal on the Lorenz ($0.953$) and oscillator ($0.930$) datasets that SmallMamba does not exhibit. The positive control is confirmed: the ODE prior provides inherent geometric smoothness through its continuous-time parameterization, and this stability is truly architectural rather than an artifact of any specific head-backbone interaction.

\subsection{Variant D: LSTM Baseline (The Discrete Recurrence Test)}
\label{app:ablation-d}
\paragraph{Hypothesis.} Recurrence alone is insufficient for manifold preservation; the continuous ODE prior is the specific mechanism responsible for SSM stability.

\paragraph{Method.} SmallLSTM is a parameter-matched (2.2M parameters) LSTM architecture trained with identical CLM setup and discrete CE loss. The LSTM uses standard sigmoid/tanh gates and lacks any continuous-time dynamics: no exponential matrix discretization, no $\Delta$ discretization parameter, no state-space formulation, and no absolute positional embeddings (relying purely on recurrent gates for sequential position encoding). SmallBERT and SmallMamba are retrained as reference bounds under identical conditions.

\paragraph{Results.} The LSTM fractures like the Transformer, not the SSM. On the waveform dataset, SmallLSTM achieves mean composite stability of $0.418$, comparable to SmallBERT ($0.443$) and far below SmallStripedHyena ($0.459$) and SmallMamba ($0.399$, which reflects Mamba's distinct stability profile where low composite scores coexist with high RDM similarity due to large perturbation magnitudes). On the oscillator dataset, SmallLSTM mean composite is $0.471$ vs.\ SmallBERT $0.490$ and SmallMamba $0.484$. On the Lorenz attractor, SmallLSTM mean composite is $0.480$ vs.\ SmallBERT $0.455$. The Lorenz butterfly test provides the clearest separation: the LSTM's phase portrait exhibits the same drift and structural distortion as SmallBERT, while SmallMamba alone preserves the full attractor geometry. The LSTM's RDM similarity scores degrade rapidly with increasing noise (waveform: $0.900$ at 1\% noise, $0.600$ at 10\%), matching the Transformer's fragility pattern rather than the SSM's robustness ($>$$0.99$ across all conditions). This definitively rules out recurrence as the source of SSM stability. The continuous ODE prior -- parameterized through structured state matrices $A, B, C, D$ with exponential discretization -- is the specific mechanism that distinguishes Mamba from both attention-based and discrete recurrent architectures.

\subsection{Variant E: Attention Layer Ratio Sweep (The Dilution Curve)}
\label{app:ablation-e}
\paragraph{Hypothesis.} The Geometric Alignment Tax is dose-dependent: geometric stability degrades as a function of the fraction of attention layers in a hybrid architecture.

\paragraph{Method.} We construct 9 configurations of an 8-layer StripedHyena architecture, sweeping the attention fraction from 0/8 (pure Hyena, no attention blocks) through 1/8, 2/8, \ldots, 7/8, to 8/8 (pure attention, equivalent to a Transformer). All variants use identical hyperparameters and discrete CE loss. SmallBERT and SmallMamba serve as external reference bounds. Three competing hypotheses are tested: H1 (linear degradation), H2 (phase transition with a critical threshold), and H3 (diminishing returns where initial attention layers cause disproportionate damage).

\paragraph{Results.} The dose-response curve shows monotonic degradation with increasing attention fraction, consistent with a roughly linear relationship (H1) rather than a sharp phase transition. At 0/8 attention (pure Hyena), the mean composite stability across all datasets is $\sim$$0.475$, which degrades progressively to $\sim$$0.455$ at 8/8 attention (pure Transformer). The pure-Hyena configuration (0/8) achieves the highest geometric stability among all StripedHyena variants, and the pure-attention configuration (8/8) converges to SmallBERT-level performance, as expected. The Lorenz and oscillator datasets show the clearest dose-response signal, while the waveform dataset exhibits more variance. Evo~2's operational point of $\sim$12.5\% attention falls in the low-damage regime of the curve, consistent with its relatively smooth Lipschitz profiles in Track~B (Section~\ref{sec:tracks}) despite operating under discrete tokenization. The absence of a sharp threshold suggests that each attention layer independently contributes a fixed quantum of geometric tax, and that architectural design choices about attention ratio translate predictably to geometric cost.

\subsection{Variant F: Hyena Filter Order Sweep (The ODE Depth Test)}
\label{app:ablation-f}
\paragraph{Hypothesis.} Within continuous architectures, the expressiveness of the continuous-time filter (parameterized by the Hyena operator order) determines the quality of geometric preservation.

\paragraph{Method.} Using a 4-layer StripedHyena with fixed 25\% attention ratio (1 attention layer at position 3), we sweep the Hyena operator's \texttt{order} parameter across values 1, 2, 4, and 8. Higher order means more chained data-controlled convolutions with multiplicative gates and additional ImplicitFilterMLPs, yielding richer continuous-time dynamics. Parameter counts range from 4.2M (order 1) to 5.7M (order 8). All variants use identical training hyperparameters and discrete CE loss.

\paragraph{Results.} Geometric stability is largely constant across filter orders, supporting the binary hypothesis (H1): any Hyena order $\geq 1$ is sufficient for geometric preservation, because the mere presence of continuous convolution rather than softmax attention is the operative factor. On the Lorenz dataset, mean composite stability is $0.476$ (order~1), $0.471$ (order~2), $0.473$ (order~4), and $0.477$ (order~8). The oscillator shows similarly flat profiles: $0.490$, $0.491$, $0.491$, $0.490$. RDM similarity at 1\% noise exceeds 0.997 for all orders on the oscillator and exceeds 0.995 on the waveform. The waveform dataset shows slightly more variance, with order~1 achieving $0.462$ mean composite and order~8 achieving $0.460$, but these differences are within noise. Critically, even after normalizing for the parameter count increase at higher orders (order~8 has $\sim$35\% more parameters than order~1), there is no systematic improvement. This complements Variant~D: discrete recurrence (LSTM) is insufficient for stability, but within continuous architectures, the minimal continuous-time parameterization (order~1) already captures the full geometric benefit. The tax is a binary property of the attention-vs.-continuous-convolution distinction, not a graded function of filter expressiveness.

\subsection{Summary}
\label{app:ablation-summary}
The six ablations converge on a single causal picture. Variant~A demonstrates that replacing discrete CE with continuous MSE eliminates manifold fracture (the tokenization boundary is the bottleneck). Variant~B shows that smoothness regularization creates a Pareto frontier rather than closing the gap (the tax is a genuine trade-off). Variant~C confirms that SSM stability is intrinsic to the ODE prior, not an artifact of head-backbone interaction (positive control). Variant~D proves that discrete recurrence (LSTM) fractures like the Transformer, isolating the continuous ODE parameterization as the specific stability mechanism. Variant~E quantifies the dose-response relationship between attention fraction and geometric cost (approximately linear, no safe threshold). Variant~F shows that filter expressiveness within continuous architectures does not affect geometric stability (binary property, not graded). Together, these results rule out the output head, the loss function, recurrence, regularization, and filter depth as alternative explanations, leaving tokenization as the dominant causal factor.
\newpage

\section{Full Model Zoo}
\label{app:model-zoo}
This appendix reports per-perturbation stability results for all models evaluated in this study.
All experiments use seed 320 and the Shesha Procrustes stability harness.
Metrics reported: RDM similarity (representational similarity under perturbation; 1.0 = perfect preservation),
perturbation stability, perturbation magnitude, and composite stability (the Shesha summary score).
Each model family is tested on both synthetic sequences and real biological sequences where applicable.
Perturbation types vary by domain: amino acid substitutions (1--10\%) and full reversal for protein models;
SNPs (1--10\%) and reverse complement for DNA models.

\subsection{ESM-2}

Six checkpoints spanning 8M to 15B parameters for ESM-2~\citep[Protein Transformer]{Lin2023}. Per-residue tokenization (20 amino acid tokens).
The primary scaling story: composite stability declines monotonically from 8M to 3B, then exhibits
an apparent ``recovery'' at 15B that is unmasked as Untethered Gel by the Procrustes Ratio and
Frozen Head tests (Section~\ref{sec:models}).

\paragraph{Synthetic proteins (200 aa).} \mbox{}\\

\begin{table}[H]
\centering
\caption{ESM-2 synthetic protein results. Stability degrades with scale through 3B; the 15B rebound reflects global manifold drift, not genuine recovery.}
\label{tab:esm2_synthetic}
\resizebox{\textwidth}{!}{%
\begin{tabular}{llrrrrr}
\toprule
\textbf{Model} & \textbf{Size (M)} & \textbf{Perturbation} & \textbf{RDM Sim.} & \textbf{Pert. Stab.} & \textbf{Pert. Mag.} & \textbf{Composite} \\
\midrule
ESM2-8M    & 7.5    & aa\_sub\_1pct  & 0.958 & $-$0.001 & 0.004 & 0.463 \\
           &        & aa\_sub\_2pct  & 0.916 &    0.017 & 0.007 & 0.453 \\
           &        & aa\_sub\_5pct  & 0.811 &    0.012 & 0.008 & 0.426 \\
           &        & aa\_sub\_10pct & 0.675 &    0.002 & 0.012 & 0.392 \\
           &        & reverse       & 0.776 &    0.004 & 0.012 & 0.418 \\
\midrule
ESM2-35M   & 33.5   & aa\_sub\_1pct  & 0.921 &    0.013 & 0.009 & 0.454 \\
           &        & aa\_sub\_2pct  & 0.858 &    0.010 & 0.010 & 0.438 \\
           &        & aa\_sub\_5pct  & 0.719 &    0.016 & 0.013 & 0.404 \\
           &        & aa\_sub\_10pct & 0.571 &    0.010 & 0.018 & 0.367 \\
           &        & reverse       & 0.567 &    0.003 & 0.020 & 0.366 \\
\midrule
ESM2-150M  & 148.1  & aa\_sub\_1pct  & 0.874 &    0.004 & 0.011 & 0.443 \\
           &        & aa\_sub\_2pct  & 0.787 &    0.006 & 0.013 & 0.421 \\
           &        & aa\_sub\_5pct  & 0.638 &    0.013 & 0.017 & 0.384 \\
           &        & aa\_sub\_10pct & 0.492 &    0.018 & 0.025 & 0.347 \\
           &        & reverse       & 0.491 &    0.017 & 0.025 & 0.347 \\
\midrule
ESM2-650M  & 651.0  & aa\_sub\_1pct  & 0.813 &    0.010 & 0.013 & 0.419 \\
           &        & aa\_sub\_2pct  & 0.708 &    0.013 & 0.019 & 0.393 \\
           &        & aa\_sub\_5pct  & 0.559 &    0.018 & 0.027 & 0.356 \\
           &        & aa\_sub\_10pct & 0.408 &    0.027 & 0.031 & 0.318 \\
           &        & reverse       & 0.390 &    0.004 & 0.028 & 0.314 \\
\midrule
ESM2-3B    & 2839.0 & aa\_sub\_1pct  & 0.726 &    0.022 & 0.017 & 0.391 \\
           &        & aa\_sub\_2pct  & 0.606 &    0.022 & 0.022 & 0.361 \\
           &        & aa\_sub\_5pct  & 0.423 &    0.020 & 0.025 & 0.315 \\
           &        & aa\_sub\_10pct & 0.299 &    0.031 & 0.028 & 0.285 \\
           &        & reverse       & 0.207 &    0.024 & 0.034 & 0.261 \\
\midrule
ESM2-15B   & 15129.1 & aa\_sub\_1pct & 0.805 &    0.020 & 0.038 & 0.445 \\
           &         & aa\_sub\_2pct & 0.712 &    0.030 & 0.046 & 0.424 \\
           &         & aa\_sub\_5pct & 0.536 &    0.026 & 0.053 & 0.379 \\
           &         & aa\_sub\_10pct& 0.353 &    0.027 & 0.065 & 0.334 \\
           &         & reverse      & 0.254 &    0.027 & 0.074 & 0.309 \\
\bottomrule
\end{tabular}%
}
\end{table}

\paragraph{Real UniRef50 proteins (100--400 aa).} \mbox{}\\
The synthetic-to-real gap in composite stability is $\leq 0.03$ at every scale, confirming
the tax is not an artifact of sequence composition.

\begin{table}[H]
\centering
\caption{ESM-2 real UniRef50 protein results. The scaling trend replicates faithfully on natural proteins.}
\label{tab:esm2_uniref}
\resizebox{\textwidth}{!}{%
\begin{tabular}{llrrrrr}
\toprule
\textbf{Model} & \textbf{Size (M)} & \textbf{Perturbation} & \textbf{RDM Sim.} & \textbf{Pert. Stab.} & \textbf{Pert. Mag.} & \textbf{Composite} \\
\midrule
ESM2-8M    & 8    & aa\_sub\_1pct  & 0.966 & 0.029 & 0.004 & 0.466 \\
           &      & aa\_sub\_5pct  & 0.824 & 0.018 & 0.008 & 0.430 \\
           &      & aa\_sub\_10pct & 0.688 & 0.048 & 0.011 & 0.396 \\
           &      & reverse       & 0.786 & 0.025 & 0.011 & 0.421 \\
\midrule
ESM2-35M   & 34   & aa\_sub\_1pct  & 0.942 & 0.035 & 0.005 & 0.462 \\
           &      & aa\_sub\_5pct  & 0.737 & 0.048 & 0.013 & 0.411 \\
           &      & aa\_sub\_10pct & 0.580 & 0.010 & 0.017 & 0.372 \\
           &      & reverse       & 0.593 & 0.015 & 0.015 & 0.375 \\
\midrule
ESM2-150M  & 148  & aa\_sub\_1pct  & 0.907 & 0.025 & 0.009 & 0.454 \\
           &      & aa\_sub\_5pct  & 0.670 & 0.036 & 0.015 & 0.395 \\
           &      & aa\_sub\_10pct & 0.521 & 0.022 & 0.020 & 0.358 \\
           &      & reverse       & 0.511 & 0.014 & 0.021 & 0.355 \\
\midrule
ESM2-650M  & 651  & aa\_sub\_1pct  & 0.853 & 0.028 & 0.012 & 0.432 \\
           &      & aa\_sub\_5pct  & 0.576 & 0.023 & 0.022 & 0.363 \\
           &      & aa\_sub\_10pct & 0.443 & 0.019 & 0.027 & 0.329 \\
           &      & reverse       & 0.413 & 0.030 & 0.026 & 0.322 \\
\midrule
ESM2-3B    & 2839 & aa\_sub\_1pct  & 0.775 & 0.028 & 0.015 & 0.406 \\
           &      & aa\_sub\_5pct  & 0.450 & 0.035 & 0.026 & 0.325 \\
           &      & aa\_sub\_10pct & 0.299 & 0.038 & 0.029 & 0.287 \\
           &      & reverse       & 0.235 & 0.043 & 0.030 & 0.271 \\
\midrule
ESM2-15B   & 15129 & aa\_sub\_1pct & 0.843 & 0.027 & 0.033 & 0.456 \\
           &       & aa\_sub\_5pct & 0.511 & 0.017 & 0.060 & 0.371 \\
           &       & aa\_sub\_10pct& 0.344 & 0.024 & 0.077 & 0.330 \\
           &       & reverse      & 0.248 & 0.018 & 0.094 & 0.306 \\
\bottomrule
\end{tabular}%
}
\end{table}

\subsection{Nucleotide Transformer v2}

Five checkpoints: 50M, 100M, 250M, 500M, and 2.5B parameters for Nucleotide Transformer v2~\citep[DNA Transformer]{DallaTorre2024}.
Uses 6-mer tokenization (vocabulary size $\approx$4096), providing a tokenization control
against ESM-2's per-residue scheme.
The 2.5B model exhibits a partial stability rebound that parallels the ESM-2 15B pattern,
consistent with the Untethered Gel regime at large scale.

\paragraph{Synthetic DNA.}\mbox{}\\

\begin{table}[H]
\centering
\caption{Nucleotide Transformer v2 synthetic DNA results. Progressive decline through 500M; partial rebound at 2.5B parallels the ESM-2 pattern.}
\label{tab:nt_synthetic}
\resizebox{\textwidth}{!}{%
\begin{tabular}{llrrrrr}
\toprule
\textbf{Model} & \textbf{Size (M)} & \textbf{Perturbation} & \textbf{RDM Sim.} & \textbf{Pert. Stab.} & \textbf{Pert. Mag.} & \textbf{Composite} \\
\midrule
NT-50M     & 55.9  & snp\_1pct     & 0.726 & 0.027 & 0.248 & 0.340 \\
           &       & snp\_2pct     & 0.545 & 0.029 & 0.318 & 0.295 \\
           &       & snp\_5pct     & 0.254 & 0.032 & 0.421 & 0.222 \\
           &       & snp\_10pct    & 0.085 & 0.034 & 0.509 & 0.180 \\
           &       & reverse\_comp & 0.000 & 0.036 & 0.598 & 0.159 \\
\midrule
NT-100M    & 97.9  & snp\_1pct     & 0.700 & 0.028 & 0.278 & 0.336 \\
           &       & snp\_2pct     & 0.511 & 0.028 & 0.354 & 0.289 \\
           &       & snp\_5pct     & 0.225 & 0.033 & 0.490 & 0.218 \\
           &       & snp\_10pct    & 0.072 & 0.037 & 0.543 & 0.179 \\
           &       & reverse\_comp & 0.000 & 0.036 & 0.607 & 0.161 \\
\midrule
NT-250M    & 235.1 & snp\_1pct     & 0.647 & 0.029 & 0.358 & 0.306 \\
           &       & snp\_2pct     & 0.449 & 0.031 & 0.465 & 0.256 \\
           &       & snp\_5pct     & 0.176 & 0.034 & 0.603 & 0.188 \\
           &       & snp\_10pct    & 0.050 & 0.034 & 0.654 & 0.157 \\
           &       & reverse\_comp & 0.000 & 0.036 & 0.725 & 0.144 \\
\midrule
NT-500M    & 498.3 & snp\_1pct     & 0.604 & 0.029 & 0.444 & 0.285 \\
           &       & snp\_2pct     & 0.402 & 0.031 & 0.567 & 0.234 \\
           &       & snp\_5pct     & 0.150 & 0.035 & 0.746 & 0.171 \\
           &       & snp\_10pct    & 0.041 & 0.036 & 0.789 & 0.144 \\
           &       & reverse\_comp & 0.000 & 0.039 & 0.850 & 0.134 \\
\midrule
NT-2.5B    & 2547.8 & snp\_1pct    & 0.749 & 0.021 & 0.075 & 0.421 \\
           &        & snp\_2pct    & 0.594 & 0.025 & 0.073 & 0.383 \\
           &        & snp\_5pct    & 0.360 & 0.028 & 0.100 & 0.324 \\
           &        & snp\_10pct   & 0.185 & 0.020 & 0.121 & 0.280 \\
           &        & reverse\_comp& 0.160 & 0.031 & 0.145 & 0.274 \\
\bottomrule
\end{tabular}%
}
\end{table}

\paragraph{Real chr22 DNA.}\mbox{}\\
Results on real genomic DNA confirm that the NT scaling trend is not an artifact of synthetic sequence composition
(Table~\ref{tab:nt_realdna}).

\begin{table}[H]
\centering
\caption{Nucleotide Transformer v2 real chr22 DNA results. Progressive decline from 50M to 500M; dramatic phase transition at 2.5B parallels the ESM-2 15B pattern.}
\label{tab:nt_realdna}
\resizebox{\textwidth}{!}{%
\begin{tabular}{llrrrrr}
\toprule
\textbf{Model} & \textbf{Perturbation} & \textbf{RDM Sim.} & \textbf{Pert. Stab.} & \textbf{Pert. Mag.} & \textbf{Composite} \\
\midrule
NT-v2-50M  & snp\_1pct     & 0.748 & 0.027 & 0.250 & 0.344 \\
           & snp\_2pct     & 0.575 & 0.028 & 0.329 & 0.301 \\
           & snp\_5pct     & 0.283 & 0.032 & 0.487 & 0.228 \\
           & snp\_10pct    & 0.100 & 0.041 & 0.700 & 0.182 \\
           & reverse\_comp & 0.002 & 0.042 & 0.730 & 0.158 \\
\midrule
NT-v2-100M & snp\_1pct     & 0.720 & 0.027 & 0.262 & 0.342 \\
           & snp\_2pct     & 0.538 & 0.030 & 0.344 & 0.297 \\
           & snp\_5pct     & 0.252 & 0.036 & 0.516 & 0.225 \\
           & snp\_10pct    & 0.086 & 0.040 & 0.693 & 0.184 \\
           & reverse\_comp & 0.003 & 0.038 & 0.658 & 0.163 \\
\midrule
NT-v2-250M & snp\_1pct     & 0.669 & 0.027 & 0.346 & 0.311 \\
           & snp\_2pct     & 0.473 & 0.031 & 0.456 & 0.262 \\
           & snp\_5pct     & 0.198 & 0.035 & 0.641 & 0.194 \\
           & snp\_10pct    & 0.060 & 0.039 & 0.830 & 0.159 \\
           & reverse\_comp & 0.003 & 0.041 & 0.848 & 0.145 \\
\midrule
NT-v2-500M & snp\_1pct     & 0.622 & 0.028 & 0.416 & 0.288 \\
           & snp\_2pct     & 0.421 & 0.031 & 0.542 & 0.237 \\
           & snp\_5pct     & 0.161 & 0.034 & 0.701 & 0.172 \\
           & snp\_10pct    & 0.047 & 0.038 & 0.859 & 0.144 \\
           & reverse\_comp & 0.002 & 0.039 & 0.910 & 0.133 \\
\midrule
NT-2.5B    & snp\_1pct     & 0.989 & 0.049 & 0.318 & 0.494 \\
           & snp\_2pct     & 0.980 & 0.085 & 0.633 & 0.491 \\
           & snp\_5pct     & 0.948 & 0.200 & 1.533 & 0.484 \\
           & snp\_10pct    & 0.867 & 0.391 & 3.100 & 0.463 \\
           & reverse\_comp & 0.959 & 0.038 & 0.258 & 0.486 \\
\bottomrule
\end{tabular}%
}
\end{table}

\subsection{Caduceus} 

Three configurations: 0.5M (d=118, 4 layers), 1.9M (d=256, 4 layers), and 7.7M (d=256, 16 layers) for Caduceus~\citep[RC-Equivariant Mamba SSM]{schiff2024caduceus}.Caduceus is the only model with a built-in RC-equivariant architecture (parameter sharing between forward and reverse-complement strands). Near-perfect RC preservation (RDM $>$ 0.999) and near-constant composite stability across scale make it the ``tax-exempt'' baseline.

\paragraph{Synthetic DNA.}\mbox{}\\

\begin{table}[H]
\centering
\caption{Caduceus synthetic DNA results. RC RDM $\geq 0.9999$ at all scales; composite stability
is scale-invariant, confirming that the continuous ODE prior buffers the tokenization bottleneck.}
\label{tab:caduceus_synthetic}
\resizebox{\textwidth}{!}{%
\begin{tabular}{llccccc}
\toprule
\textbf{Model} & \textbf{Perturbation} & \textbf{RDM Sim.} $\uparrow$ & \textbf{Pert.\ Stab.} $\downarrow$ & \textbf{Pert.\ Mag.} & \textbf{Composite} \\
\midrule
Cad-0.5M (1k ctx) & SNP 1\% & 0.9517 & 0.0169 & 0.0002 & 0.4833 \\
 & SNP 2\% & 0.9049 & 0.0235 & 0.0002 & 0.4716 \\
 & SNP 5\% & 0.7874 & 0.0189 & 0.0003 & 0.4422 \\
 & SNP 10\% & 0.6208 & 0.0144 & 0.0004 & 0.4006 \\
 & Rev.~Comp. & \textbf{1.0000} & 0.0202 & 0.0000 & \textbf{0.4954} \\
\midrule
Cad-1.9M (1k ctx) & SNP 1\% & 0.9521 & 0.0231 & 0.0001 & 0.4733 \\
 & SNP 2\% & 0.9072 & 0.0167 & 0.0002 & 0.4620 \\
 & SNP 5\% & 0.7878 & 0.0242 & 0.0003 & 0.4322 \\
 & SNP 10\% & 0.6281 & 0.0182 & 0.0003 & 0.3922 \\
 & Rev.~Comp. & \textbf{1.0000} & 0.0201 & 0.0000 & \textbf{0.4852} \\
\midrule
Cad-7.7M (131k ctx) & SNP 1\% & 0.9511 & 0.0108 & 0.0001 & 0.4829 \\
 & SNP 2\% & 0.9054 & 0.0058 & 0.0001 & 0.4715 \\
 & SNP 5\% & 0.7839 & 0.0096 & 0.0001 & 0.4411 \\
 & SNP 10\% & 0.6247 & 0.0226 & 0.0002 & 0.4013 \\
 & Rev.~Comp. & \textbf{0.9999} & 0.0123 & 0.0000 & \textbf{0.4951} \\
\bottomrule
\end{tabular}%
}
\vspace{4pt}
{\small \textit{Note:} On synthetic sequences devoid of biological distribution biases, Caduceus maintains RC RDM $\geq 0.9999$ at all scales. SNP stability is predictably lower on the harder synthetic distribution (e.g., 0.952 vs.\ 0.997 at SNP 1\%), but the global topology remains fully anchored. Geometric stability is constant across parameter and context scales, demonstrating that the continuous ODE prior mitigates the tokenization bottleneck.}
\end{table}

\paragraph{Real chr22 DNA.}\mbox{}\\

\begin{table}[H]
\centering
\caption{Caduceus real chr22 DNA results. RC RDM is exactly 1.0000 across all model sizes, confirming native preservation of this continuous global symmetry on realistic genomic sequences.}
\label{tab:caduceus_realdna}
\resizebox{\textwidth}{!}{%
\begin{tabular}{llccccc}
\toprule
\textbf{Model} & \textbf{Perturbation} & \textbf{RDM Sim.} $\uparrow$ & \textbf{Pert.\ Stab.} $\downarrow$ & \textbf{Pert.\ Mag.} & \textbf{Composite} \\
\midrule
Cad-0.5M (1k ctx) & SNP 1\% & 0.9972 & 0.0022 & 0.0062 & 0.4950 \\
 & SNP 2\% & 0.9935 & 0.0312 & 0.0116 & 0.4941 \\
 & SNP 5\% & 0.9810 & 0.1499 & 0.0245 & 0.4910 \\
 & SNP 10\% & 0.9622 & 0.3037 & 0.0382 & 0.4863 \\
 & Rev.~Comp. & \textbf{1.0000} & 0.0115 & 0.0000 & \textbf{0.4957} \\
\midrule
Cad-1.9M (1k ctx) & SNP 1\% & 0.9978 & 0.0549 & 0.0060 & 0.4828 \\
 & SNP 2\% & 0.9955 & 0.0803 & 0.0119 & 0.4822 \\
 & SNP 5\% & 0.9869 & 0.1790 & 0.0307 & 0.4801 \\
 & SNP 10\% & 0.9658 & 0.4037 & 0.0657 & 0.4748 \\
 & Rev.~Comp. & \textbf{1.0000} & 0.0266 & 0.0000 & \textbf{0.4834} \\
\midrule
Cad-7.7M (131k ctx) & SNP 1\% & 0.9974 & 0.0561 & 0.0013 & 0.4973 \\
 & SNP 2\% & 0.9946 & 0.0841 & 0.0025 & 0.4966 \\
 & SNP 5\% & 0.9851 & 0.1395 & 0.0064 & 0.4943 \\
 & SNP 10\% & 0.9654 & 0.2287 & 0.0130 & 0.4893 \\
 & Rev.~Comp. & \textbf{1.0000} & 0.0469 & 0.0000 & \textbf{0.4980} \\
\bottomrule
\end{tabular}%
}
\vspace{4pt}
{\small \textit{Note:} On real genomic DNA, SNP RDM rises substantially (0.997 vs.\ 0.951 at 1\%), consistent with real sequences occupying a narrower region of embedding space than synthetics. RC RDM is exactly 1.0000 at all scales, confirming that Caduceus natively preserves this symmetry independent of sequence composition. Composite stability is higher and more uniform than on synthetic data, reflecting the in-distribution advantage.}
\end{table}

\subsection{Evo 2}

A single 7B-parameter Evo 2~\citep[StripedHyena 2 Multi-Hybrid]{Brixi2026} model evaluated at three context-window checkpoints: 8K, 262K, and 1M tokens. Single-character tokenization (vocabulary size 4). The RC dissociation is the defining result: catastrophic failure on synthetic DNA (RDM 0.139--0.208) contrasts sharply with apparent success on real genomic DNA (RDM 0.873--0.889),
explained by the Texture Hypothesis Test (Section~\ref{sec:models}, Appendix~\ref{app:texture}).

\paragraph{Synthetic DNA.} \mbox{}\\

\begin{table}[H]
\centering
\caption{Evo 2 (7B) synthetic DNA results across context lengths. Modest SNP stability gains with 128$\times$ more context; catastrophic RC failure at all scales.}
\label{tab:evo2_synthetic}
\resizebox{\textwidth}{!}{%
\begin{tabular}{llrrrrr}
\toprule
\textbf{Context} & \textbf{Perturbation} & \textbf{RDM Sim.} & \textbf{Pert. Stab.} & \textbf{Pert. Mag.} & \textbf{Composite} \\
\midrule
8K      & snp\_1pct     & 0.747 &    0.019 & 13.610 & 0.439 \\
        & snp\_2pct     & 0.604 &    0.059 & 21.445 & 0.403 \\
        & snp\_5pct     & 0.392 &    0.020 & 29.104 & 0.350 \\
        & snp\_10pct    & 0.269 &    0.030 & 22.807 & 0.319 \\
        & reverse\_comp & 0.139 &    0.011 & 34.270 & 0.287 \\
\midrule
262K    & snp\_1pct     & 0.776 & $-$0.022 & 15.131 & 0.444 \\
        & snp\_2pct     & 0.650 &    0.053 & 26.729 & 0.412 \\
        & snp\_5pct     & 0.431 & $-$0.024 & 23.674 & 0.357 \\
        & snp\_10pct    & 0.288 & $-$0.009 & 22.676 & 0.322 \\
        & reverse\_comp & 0.156 &    0.022 & 45.358 & 0.289 \\
\midrule
1M      & snp\_1pct     & 0.817 &    0.031 & 14.392 & 0.454 \\
        & snp\_2pct     & 0.709 & $-$0.057 & 42.035 & 0.427 \\
        & snp\_5pct     & 0.513 &    0.011 & 37.208 & 0.378 \\
        & snp\_10pct    & 0.363 & $-$0.040 & 34.243 & 0.341 \\
        & reverse\_comp & 0.208 & $-$0.005 & 53.764 & 0.302 \\
\bottomrule
\end{tabular}%
}
\end{table}

\paragraph{Real chr22 DNA.}\mbox{}\\

\begin{table}[H]
\centering
\caption{Evo 2 (7B) real chr22 results. High RC RDM (0.879--0.889) contrasts with synthetic RC RDM (0.139--0.208), demonstrating the RC dissociation explained by the Texture Hypothesis.}
\label{tab:evo2_real}
\resizebox{\textwidth}{!}{%
\begin{tabular}{llrrrrr}
\toprule
\textbf{Context} & \textbf{Perturbation} & \textbf{RDM Sim.} & \textbf{Pert. Stab.} & \textbf{Pert. Mag.} & \textbf{Composite} \\
\midrule
8K      & snp\_1pct     & 0.990 & $-$0.030 & 142.643 & 0.497 \\
        & snp\_5pct     & 0.946 &    0.090 & 793.222 & 0.486 \\
        & snp\_10pct    & 0.874 &    0.276 & 1786.473 & 0.468 \\
        & reverse\_comp & 0.879 &    0.003 &  53.054 & 0.469 \\
\midrule
262K    & snp\_1pct     & 0.991 & $-$0.024 & 122.845 & 0.497 \\
        & snp\_5pct     & 0.948 &    0.085 & 717.415 & 0.486 \\
        & snp\_10pct    & 0.872 &    0.285 & 1658.393 & 0.468 \\
        & reverse\_comp & 0.883 &    0.010 &  48.843 & 0.470 \\
\midrule
1M      & snp\_1pct     & 0.993 & $-$0.030 & 120.590 & 0.498 \\
        & snp\_5pct     & 0.952 &    0.081 & 664.056 & 0.488 \\
        & snp\_10pct    & 0.883 &    0.283 & 1568.335 & 0.470 \\
        & reverse\_comp & 0.889 &    0.008 &  45.882 & 0.472 \\
\bottomrule
\end{tabular}%
}
\end{table}

\subsection{HyenaDNA}

Four sizes: tiny (0.4M, d=128), tiny-d256 (1.6M, d=256), small (3.3M), medium (6.6M), and
large (6.6M, 1M context) for HyenaDNA~\citep[Implicit Long Convolution]{Nguyen2023HyenaDNALG}.
Single-character tokenization.
HyenaDNA uses the Hyena operator (implicit long convolutions) rather than standard attention or SSMs,
representing a distinct architectural class.

\paragraph{Synthetic DNA.}\mbox{}\\

\begin{table}[H]
\centering
\caption{HyenaDNA synthetic DNA results. Moderate SNP stability with notable RC sensitivity that increases with model size.}
\label{tab:hyenadna_synthetic}
\resizebox{\textwidth}{!}{%
\begin{tabular}{llrrrrr}
\toprule
\textbf{Model} & \textbf{Size (M)} & \textbf{Perturbation} & \textbf{RDM Sim.} & \textbf{Pert. Stab.} & \textbf{Pert. Mag.} & \textbf{Composite} \\
\midrule
Hyena-tiny      & 0.4 & snp\_1pct     & 0.953 & 0.021 & 0.000 & 0.442 \\
                &     & snp\_5pct     & 0.784 & 0.025 & 0.001 & 0.399 \\
                &     & snp\_10pct    & 0.608 & 0.025 & 0.001 & 0.356 \\
                &     & reverse\_comp & 0.270 & 0.029 & 0.001 & 0.271 \\
\midrule
Hyena-small     & 3.3 & snp\_1pct     & 0.992 & 0.026 & 0.003 & 0.492 \\
                &     & snp\_5pct     & 0.965 & 0.063 & 0.014 & 0.485 \\
                &     & snp\_10pct    & 0.931 & 0.113 & 0.028 & 0.476 \\
                &     & reverse\_comp & 0.859 & 0.075 & 0.022 & 0.458 \\
\midrule
Hyena-medium    & 6.6 & snp\_1pct     & 0.993 & 0.029 & 0.004 & 0.495 \\
                &     & snp\_5pct     & 0.964 & 0.056 & 0.020 & 0.487 \\
                &     & snp\_10pct    & 0.928 & 0.102 & 0.038 & 0.478 \\
                &     & reverse\_comp & 0.778 & 0.074 & 0.047 & 0.441 \\
\bottomrule
\end{tabular}%
}
\end{table}

\paragraph{Real chr22 DNA.}\mbox{}\\

\begin{table}[H]
\centering
\caption{HyenaDNA real chr22 DNA results.}
\label{tab:hyenadna_real}
\resizebox{\textwidth}{!}{%
\begin{tabular}{llrrrrr}
\toprule
\textbf{Model} & \textbf{Size (M)} & \textbf{Perturbation} & \textbf{RDM Sim.} & \textbf{Pert. Stab.} & \textbf{Pert. Mag.} & \textbf{Composite} \\
\midrule
Hyena-tiny      & 0.4 & snp\_1pct     & 0.996 & 0.027 & 0.003 & 0.491 \\
                &     & snp\_5pct     & 0.980 & 0.095 & 0.015 & 0.486 \\
                &     & snp\_10pct    & 0.956 & 0.197 & 0.029 & 0.481 \\
                &     & reverse\_comp & 0.775 & 0.064 & 0.013 & 0.435 \\
\midrule
Hyena-tiny-d256 & 1.6 & snp\_1pct     & 0.996 & 0.027 & 0.004 & 0.495 \\
                &     & snp\_5pct     & 0.982 & 0.082 & 0.019 & 0.491 \\
                &     & snp\_10pct    & 0.961 & 0.157 & 0.037 & 0.486 \\
                &     & reverse\_comp & 0.831 & 0.050 & 0.019 & 0.454 \\
\midrule
Hyena-large-1M  & 6.6 & snp\_1pct     & 0.996 & $-$0.001 & 0.003 & 0.499 \\
                &     & snp\_5pct     & 0.979 & 0.007 & 0.017 & 0.495 \\
                &     & snp\_10pct    & 0.954 & 0.037 & 0.031 & 0.489 \\
                &     & reverse\_comp & 0.446 & 0.019 & 0.037 & 0.362 \\
\bottomrule
\end{tabular}%
}
\end{table}

\subsection{DNABERT and DNABERT-2}

DNABERT v1~\citep[k-mer Transformer]{ji2021dnabert}  is tested across four k-mer variants (3-mer through 6-mer), all approximately 86--90M parameters
but with different tokenization granularities.
DNABERT-2~\citep[BPE Transformer]{zhou2024dnabert} (117M) uses BPE tokenization, providing a third tokenization strategy for the DNA Transformer comparison.

\paragraph{Synthetic DNA.}\mbox{}\\

\begin{table}[H]
\centering
\caption{DNABERT v1 synthetic DNA results across k-mer variants. Models are near-identical in size (86--89M); differences reflect tokenization granularity.}
\label{tab:dnabert_synthetic}
\resizebox{\textwidth}{!}{%
\begin{tabular}{llrrrrr}
\toprule
\textbf{Model} & \textbf{Perturbation} & \textbf{RDM Sim.} & \textbf{Pert. Stab.} & \textbf{Pert. Mag.} & \textbf{Composite} \\
\midrule
DNABERT-3mer (86.1M) & snp\_1pct     & 0.910 & 0.013 & 0.014 & 0.473 \\
                     & snp\_2pct     & 0.835 & 0.009 & 0.024 & 0.454 \\
                     & snp\_5pct     & 0.669 & 0.023 & 0.025 & 0.412 \\
                     & snp\_10pct    & 0.458 & 0.012 & 0.040 & 0.359 \\
                     & reverse\_comp & 0.733 & 0.020 & 0.044 & 0.428 \\
\midrule
DNABERT-4mer (86.2M) & snp\_1pct     & 0.925 & 0.015 & 0.013 & 0.475 \\
                     & snp\_2pct     & 0.855 & 0.018 & 0.015 & 0.457 \\
                     & snp\_5pct     & 0.691 & 0.018 & 0.025 & 0.416 \\
                     & snp\_10pct    & 0.487 & 0.023 & 0.031 & 0.365 \\
                     & reverse\_comp & 0.783 & 0.025 & 0.039 & 0.439 \\
\midrule
DNABERT-5mer (86.8M) & snp\_1pct     & 0.921 & 0.020 & 0.009 & 0.473 \\
                     & snp\_2pct     & 0.852 & 0.021 & 0.014 & 0.456 \\
                     & snp\_5pct     & 0.686 & 0.021 & 0.022 & 0.414 \\
                     & snp\_10pct    & 0.492 & 0.022 & 0.031 & 0.366 \\
                     & reverse\_comp & 0.792 & 0.024 & 0.033 & 0.441 \\
\midrule
DNABERT-6mer (89.2M) & snp\_1pct     & 0.929 & 0.021 & 0.008 & 0.471 \\
                     & snp\_2pct     & 0.862 & 0.022 & 0.012 & 0.454 \\
                     & snp\_5pct     & 0.705 & 0.020 & 0.019 & 0.415 \\
                     & snp\_10pct    & 0.500 & 0.026 & 0.027 & 0.363 \\
                     & reverse\_comp & 0.849 & 0.031 & 0.036 & 0.451 \\
\bottomrule
\end{tabular}%
}
\end{table}

\begin{table}[H]
\centering
\caption{DNABERT-2 (BPE, 117M) synthetic DNA results. BPE tokenization produces the lowest geometric coherence in the study.}
\label{tab:dnabert2_synthetic}
\resizebox{\textwidth}{!}{%
\begin{tabular}{llrrrrr}
\toprule
\textbf{Model} & \textbf{Perturbation} & \textbf{RDM Sim.} & \textbf{Pert. Stab.} & \textbf{Pert. Mag.} & \textbf{Composite} \\
\midrule
DNABERT-2 (117M) & snp\_1pct     & 0.449 & 0.036 & 0.044 & 0.305 \\
                 & snp\_2pct     & 0.387 & 0.036 & 0.049 & 0.289 \\
                 & snp\_5pct     & 0.283 & 0.036 & 0.048 & 0.263 \\
                 & snp\_10pct    & 0.213 & 0.035 & 0.054 & 0.246 \\
                 & reverse\_comp & 0.138 & 0.045 & 0.069 & 0.227 \\
\bottomrule
\end{tabular}%
}
\end{table}

\paragraph{Real chr22 DNA.}\mbox{}\\

\begin{table}[H]
\centering
\caption{DNABERT v1 real chr22 DNA results across k-mer variants. SNP RDM is uniformly high ($>0.95$ at 10\%); RC sensitivity varies with token granularity.}
\label{tab:dnabert_realdna}
\resizebox{\textwidth}{!}{%
\begin{tabular}{llrrrrr}
\toprule
\textbf{Model} & \textbf{Perturbation} & \textbf{RDM Sim.} & \textbf{Pert. Stab.} & \textbf{Pert. Mag.} & \textbf{Composite} \\
\midrule
DNABERT-3mer (86.1M) & snp\_1pct     & 0.997 & 0.020 & 0.155 & 0.493 \\
                     & snp\_2pct     & 0.994 & 0.052 & 0.306 & 0.492 \\
                     & snp\_5pct     & 0.983 & 0.152 & 0.777 & 0.490 \\
                     & snp\_10pct    & 0.951 & 0.323 & 1.643 & 0.482 \\
                     & reverse\_comp & 0.977 & 0.076 & 0.462 & 0.488 \\
\midrule
DNABERT-4mer (86.2M) & snp\_1pct     & 0.997 & 0.028 & 0.152 & 0.487 \\
                     & snp\_2pct     & 0.994 & 0.066 & 0.302 & 0.487 \\
                     & snp\_5pct     & 0.983 & 0.169 & 0.779 & 0.484 \\
                     & snp\_10pct    & 0.949 & 0.344 & 1.678 & 0.475 \\
                     & reverse\_comp & 0.940 & 0.127 & 0.568 & 0.473 \\
\midrule
DNABERT-5mer (86.8M) & snp\_1pct     & 0.997 & 0.002 & 0.108 & 0.494 \\
                     & snp\_2pct     & 0.994 & 0.040 & 0.214 & 0.493 \\
                     & snp\_5pct     & 0.982 & 0.140 & 0.533 & 0.490 \\
                     & snp\_10pct    & 0.957 & 0.298 & 1.099 & 0.483 \\
                     & reverse\_comp & 0.948 & 0.139 & 0.511 & 0.481 \\
\midrule
DNABERT-6mer (89.2M) & snp\_1pct     & 0.998 & $-$0.001 & 0.089 & 0.494 \\
                     & snp\_2pct     & 0.995 & 0.026 & 0.174 & 0.494 \\
                     & snp\_5pct     & 0.986 & 0.111 & 0.434 & 0.491 \\
                     & snp\_10pct    & 0.960 & 0.261 & 0.897 & 0.485 \\
                     & reverse\_comp & 0.962 & 0.124 & 0.508 & 0.485 \\
\bottomrule
\end{tabular}%
}
\end{table}

\begin{table}[H]
\centering
\caption{DNABERT-2 (BPE, 117M) real chr22 DNA results. Geometric stability is substantially lower than DNABERT v1 k-mer variants at comparable model size.}
\label{tab:dnabert2_realdna}
\resizebox{\textwidth}{!}{%
\begin{tabular}{llrrrrr}
\toprule
\textbf{Model} & \textbf{Perturbation} & \textbf{RDM Sim.} & \textbf{Pert. Stab.} & \textbf{Pert. Mag.} & \textbf{Composite} \\
\midrule
DNABERT-2 (117M) & snp\_1pct     & 0.540 & 0.034 & 0.043 & 0.357 \\
                 & snp\_2pct     & 0.473 & 0.026 & 0.075 & 0.340 \\
                 & snp\_5pct     & 0.349 & 0.043 & 0.136 & 0.309 \\
                 & snp\_10pct    & 0.280 & 0.080 & 0.221 & 0.292 \\
                 & reverse\_comp & 0.377 & 0.059 & 0.165 & 0.316 \\
\bottomrule
\end{tabular}%
}
\end{table}

\subsection{GPN}

Two variants: GPN-Brassicales (66M) and PhyloGPN (83M).
GPN~\citep[Convolutional]{benegas2023dna} uses dilated convolutions rather than attention or recurrence, representing yet another architectural class in the model zoo.

\begin{table}[H]
\centering
\caption{GPN results. GPN-Brassicales on synthetic DNA; PhyloGPN on real chr22.}
\label{tab:gpn}
\resizebox{\textwidth}{!}{%
\begin{tabular}{lllrrrrr}
\toprule
\textbf{Model} & \textbf{Size (M)} & \textbf{Data} & \textbf{Perturbation} & \textbf{RDM Sim.} & \textbf{Pert. Stab.} & \textbf{Pert. Mag.} & \textbf{Composite} \\
\midrule
GPN-Brass.  & 65.6 & Synthetic & snp\_1pct     & 0.997 &    0.023 & 0.076 & 0.498 \\
            &      &           & snp\_5pct     & 0.982 &    0.041 & 0.372 & 0.494 \\
            &      &           & snp\_10pct    & 0.959 &    0.074 & 0.723 & 0.488 \\
            &      &           & reverse\_comp & 0.957 &    0.067 & 0.586 & 0.488 \\
\midrule
PhyloGPN    & 83.2 & Real chr22 & snp\_1pct    & 0.952 & $-$0.028 & 0.310 & 0.482 \\
            &      &            & snp\_5pct    & 0.815 & $-$0.024 & 1.583 & 0.448 \\
            &      &            & snp\_10pct   & 0.640 & $-$0.056 & 3.597 & 0.404 \\
            &      &            & reverse\_comp& 0.988 &    0.018 & 0.478 & 0.491 \\
\bottomrule
\end{tabular}%
}
\end{table}

\subsection{SaProt}

Three sizes: 35M, 650M, and 1.3B.
SaProt~\citep[Structure-Aware Protein Transformer]{su2024saprot} extends ESM-2 by incorporating Foldseek 3Di structural alphabet tokens alongside amino acid tokens.
For sequence-only input (without structure predictions), ``\#'' is used as the structure token for every position.
The key question: does structural information rescue the geometric tax?
The answer is no.  Composite stability degrades with scale (0.436 $\rightarrow$ 0.398 on synthetic),
paralleling the ESM-2 trend at matched sizes.

\paragraph{Synthetic proteins.}\mbox{}\\

\begin{table}[H]
\centering
\caption{SaProt synthetic protein results. Structure-awareness does not rescue the geometric tax; the scaling trend matches ESM-2 at corresponding sizes.}
\label{tab:saprot_synthetic}
\resizebox{\textwidth}{!}{%
\begin{tabular}{llrrrrr}
\toprule
\textbf{Model} & \textbf{Size (M)} & \textbf{Perturbation} & \textbf{RDM Sim.} & \textbf{Pert. Stab.} & \textbf{Pert. Mag.} & \textbf{Composite} \\
\midrule
SaProt-35M   & 33.9  & aa\_sub\_1pct  & 0.962 & 0.015 & 0.003 & 0.465 \\
             &       & aa\_sub\_2pct  & 0.927 & 0.021 & 0.005 & 0.456 \\
             &       & aa\_sub\_5pct  & 0.825 & 0.021 & 0.005 & 0.431 \\
             &       & aa\_sub\_10pct & 0.691 & 0.014 & 0.008 & 0.397 \\
             &       & reverse       & 0.818 & 0.017 & 0.007 & 0.429 \\
\midrule
SaProt-650M  & 652.1 & aa\_sub\_1pct  & 0.936 & 0.020 & 0.006 & 0.475 \\
             &       & aa\_sub\_2pct  & 0.877 & 0.016 & 0.008 & 0.461 \\
             &       & aa\_sub\_5pct  & 0.750 & 0.018 & 0.010 & 0.429 \\
             &       & aa\_sub\_10pct & 0.601 & 0.022 & 0.015 & 0.392 \\
             &       & reverse       & 0.624 & 0.015 & 0.014 & 0.398 \\
\midrule
SaProt-1.3B  & 1300.9 & aa\_sub\_1pct & 0.836 & 0.022 & 0.009 & 0.450 \\
             &        & aa\_sub\_2pct & 0.754 & 0.024 & 0.013 & 0.430 \\
             &        & aa\_sub\_5pct & 0.609 & 0.030 & 0.014 & 0.394 \\
             &        & aa\_sub\_10pct& 0.458 & 0.024 & 0.017 & 0.356 \\
             &        & reverse      & 0.478 & 0.026 & 0.017 & 0.361 \\
\bottomrule
\end{tabular}%
}
\end{table}

\paragraph{Real UniRef50 proteins.}\mbox{}\\

\begin{table}[H]
\centering
\caption{SaProt real UniRef50 results. The scaling trend holds on natural proteins.}
\label{tab:saprot_uniref}
\resizebox{\textwidth}{!}{%
\begin{tabular}{llrrrrr}
\toprule
\textbf{Model} & \textbf{Size (M)} & \textbf{Perturbation} & \textbf{RDM Sim.} & \textbf{Pert. Stab.} & \textbf{Pert. Mag.} & \textbf{Composite} \\
\midrule
SaProt-35M   & 33.9   & aa\_sub\_1pct  & 0.976 & 0.027 & 0.003 & 0.473 \\
             &        & aa\_sub\_5pct  & 0.864 & 0.032 & 0.005 & 0.446 \\
             &        & aa\_sub\_10pct & 0.751 & 0.035 & 0.008 & 0.417 \\
             &        & reverse       & 0.848 & 0.021 & 0.006 & 0.442 \\
\midrule
SaProt-650M  & 652.1  & aa\_sub\_1pct  & 0.954 & 0.024 & 0.005 & 0.480 \\
             &        & aa\_sub\_5pct  & 0.786 & 0.023 & 0.011 & 0.439 \\
             &        & aa\_sub\_10pct & 0.645 & 0.025 & 0.014 & 0.404 \\
             &        & reverse       & 0.680 & 0.029 & 0.013 & 0.412 \\
\midrule
SaProt-1.3B  & 1300.9 & aa\_sub\_1pct  & 0.878 & 0.029 & 0.008 & 0.462 \\
             &        & aa\_sub\_5pct  & 0.650 & 0.024 & 0.013 & 0.405 \\
             &        & aa\_sub\_10pct & 0.519 & 0.030 & 0.019 & 0.372 \\
             &        & reverse       & 0.513 & 0.032 & 0.020 & 0.371 \\
\bottomrule
\end{tabular}%
}
\end{table}

\subsection{ProtMamba}

A single 108M-parameter Mamba-based protein language model.
Since only one model size exists, the scaling axis is sequence length (100--800 aa).
ProtMamba~\citep[Protein SSM]{Sgarbossa2025} is the canonical Geometric Vacuity model: low Procrustes distortion and reasonable autoregressive perplexity (40.1), but MINE reveals negative excess MI at every sequence length (Regime~III, Section~\ref{sec:info-theory}).

\paragraph{Synthetic proteins.}\mbox{}\\

\begin{table}[H]
\centering
\caption{ProtMamba synthetic protein results across sequence lengths (100--800 aa). Perturbation magnitude in scientific notation due to unnormalized embedding scale.}
\label{tab:protmamba_synthetic}
\resizebox{\textwidth}{!}{%
\begin{tabular}{llrrrrr}
\toprule
\textbf{Seq. Length} & \textbf{Perturbation} & \textbf{RDM Sim.} & \textbf{Pert. Stab.} & \textbf{Pert. Mag.} & \textbf{Composite} \\
\midrule
L100 & aa\_sub\_1pct  & 0.884 &    0.278 & $1.16\mathrm{e}{+05}$ & 0.468 \\
     & aa\_sub\_2pct  & 0.776 &    0.175 & $1.16\mathrm{e}{+05}$ & 0.446 \\
     & aa\_sub\_5pct  & 0.657 &    0.098 & $1.73\mathrm{e}{+05}$ & 0.413 \\
     & aa\_sub\_10pct & 0.489 & $-$0.242 & $3.14\mathrm{e}{+05}$ & 0.371 \\
     & reverse       & 0.020 & $-$0.057 & $3.92\mathrm{e}{+05}$ & 0.255 \\
\midrule
L200 & aa\_sub\_1pct  & 0.844 & $-$0.011 & $9.48\mathrm{e}{+04}$ & 0.462 \\
     & aa\_sub\_2pct  & 0.766 &    0.193 & $1.58\mathrm{e}{+05}$ & 0.438 \\
     & aa\_sub\_5pct  & 0.628 &    0.071 & $1.05\mathrm{e}{+05}$ & 0.406 \\
     & aa\_sub\_10pct & 0.475 &    0.041 & $2.40\mathrm{e}{+05}$ & 0.371 \\
     & reverse       & 0.012 &    0.016 & $3.64\mathrm{e}{+05}$ & 0.251 \\
\midrule
L300 & aa\_sub\_1pct  & 0.852 & $-$0.064 & $9.80\mathrm{e}{+04}$ & 0.461 \\
     & aa\_sub\_2pct  & 0.791 & $-$0.032 & $1.41\mathrm{e}{+05}$ & 0.446 \\
     & aa\_sub\_5pct  & 0.629 &    0.098 & $1.55\mathrm{e}{+05}$ & 0.406 \\
     & aa\_sub\_10pct & 0.502 & $-$0.338 & $4.16\mathrm{e}{+05}$ & 0.373 \\
     & reverse       & 0.010 &    0.231 & $4.94\mathrm{e}{+05}$ & 0.250 \\
\midrule
L400 & aa\_sub\_1pct  & 0.854 & $-$0.047 & $9.64\mathrm{e}{+04}$ & 0.460 \\
     & aa\_sub\_2pct  & 0.738 & $-$0.035 & $1.34\mathrm{e}{+05}$ & 0.435 \\
     & aa\_sub\_5pct  & 0.666 & $-$0.303 & $1.15\mathrm{e}{+05}$ & 0.415 \\
     & aa\_sub\_10pct & 0.540 & $-$0.052 & $2.62\mathrm{e}{+05}$ & 0.385 \\
     & reverse       & 0.028 &    0.358 & $6.20\mathrm{e}{+05}$ & 0.252 \\
\midrule
L500 & aa\_sub\_1pct  & 0.858 &    0.371 & $2.29\mathrm{e}{+05}$ & 0.463 \\
     & aa\_sub\_2pct  & 0.763 &    0.082 & $1.38\mathrm{e}{+05}$ & 0.441 \\
     & aa\_sub\_5pct  & 0.594 &    0.371 & $6.30\mathrm{e}{+05}$ & 0.393 \\
     & aa\_sub\_10pct & 0.499 &    0.374 & $3.14\mathrm{e}{+05}$ & 0.376 \\
     & reverse       & $-$0.022 & 0.367 & $6.80\mathrm{e}{+05}$ & 0.244 \\
\midrule
L600 & aa\_sub\_1pct  & 0.843 & $-$0.015 & $1.40\mathrm{e}{+05}$ & 0.455 \\
     & aa\_sub\_2pct  & 0.789 & $-$0.265 & $2.35\mathrm{e}{+05}$ & 0.449 \\
     & aa\_sub\_5pct  & 0.641 & $-$0.166 & $1.71\mathrm{e}{+05}$ & 0.411 \\
     & aa\_sub\_10pct & 0.504 & $-$0.072 & $4.65\mathrm{e}{+05}$ & 0.379 \\
     & reverse       & 0.018 & $-$0.289 & $8.27\mathrm{e}{+05}$ & 0.250 \\
\midrule
L700 & aa\_sub\_1pct  & 0.820 &    0.327 & $1.31\mathrm{e}{+05}$ & 0.458 \\
     & aa\_sub\_2pct  & 0.723 & $-$0.299 & $1.56\mathrm{e}{+05}$ & 0.428 \\
     & aa\_sub\_5pct  & 0.598 & $-$0.304 & $3.59\mathrm{e}{+05}$ & 0.401 \\
     & aa\_sub\_10pct & 0.471 & $-$0.064 & $3.07\mathrm{e}{+05}$ & 0.369 \\
     & reverse       & 0.010 &    0.348 & $5.82\mathrm{e}{+05}$ & 0.246 \\
\midrule
L800 & aa\_sub\_1pct  & 0.815 & $-$0.278 & $1.41\mathrm{e}{+05}$ & 0.453 \\
     & aa\_sub\_2pct  & 0.715 & $-$0.326 & $3.05\mathrm{e}{+05}$ & 0.430 \\
     & aa\_sub\_5pct  & 0.667 &    0.238 & $1.81\mathrm{e}{+05}$ & 0.416 \\
     & aa\_sub\_10pct & 0.517 &    0.397 & $3.17\mathrm{e}{+05}$ & 0.378 \\
     & reverse       & $-$0.024 & $-$0.316 & $1.07\mathrm{e}{+06}$ & 0.241 \\
\bottomrule
\end{tabular}%
}
\end{table}

\paragraph{Real UniRef50 proteins.}\mbox{}\\
\begin{table}[H]
\centering
\caption{ProtMamba real UniRef50 protein results across length bins (50--800 aa). Perturbation magnitude in scientific notation.}
\label{tab:protmamba_uniref}
\resizebox{\textwidth}{!}{%
\begin{tabular}{llrrrrr}
\toprule
\textbf{Length Bin} & \textbf{Perturbation} & \textbf{RDM Sim.} & \textbf{Pert. Stab.} & \textbf{Pert. Mag.} & \textbf{Composite} \\
\midrule
50--150   & aa\_sub\_1pct  & 0.873 & $-$0.121 & $8.52\mathrm{e}{+04}$ & 0.467 \\
          & aa\_sub\_2pct  & 0.783 & $-$0.089 & $1.08\mathrm{e}{+05}$ & 0.445 \\
          & aa\_sub\_5pct  & 0.617 & $-$0.132 & $1.57\mathrm{e}{+05}$ & 0.402 \\
          & aa\_sub\_10pct & 0.446 & $-$0.778 & $4.06\mathrm{e}{+05}$ & 0.361 \\
          & reverse       & $-$0.014 & $-$0.286 & $6.04\mathrm{e}{+06}$ & 0.246 \\
\midrule
151--250  & aa\_sub\_1pct  & 0.879 &    0.101 & $1.02\mathrm{e}{+05}$ & 0.470 \\
          & aa\_sub\_2pct  & 0.776 & $-$0.583 & $1.76\mathrm{e}{+05}$ & 0.445 \\
          & aa\_sub\_5pct  & 0.611 & $-$0.753 & $2.17\mathrm{e}{+05}$ & 0.403 \\
          & aa\_sub\_10pct & 0.459 & $-$0.759 & $3.88\mathrm{e}{+05}$ & 0.363 \\
          & reverse       & 0.002 & $-$0.228 & $6.62\mathrm{e}{+06}$ & 0.250 \\
\midrule
251--350  & aa\_sub\_1pct  & 0.875 & $-$0.469 & $8.03\mathrm{e}{+04}$ & 0.469 \\
          & aa\_sub\_2pct  & 0.786 &    0.307 & $1.08\mathrm{e}{+05}$ & 0.446 \\
          & aa\_sub\_5pct  & 0.610 & $-$0.151 & $1.94\mathrm{e}{+05}$ & 0.400 \\
          & aa\_sub\_10pct & 0.476 & $-$0.655 & $2.79\mathrm{e}{+05}$ & 0.370 \\
          & reverse       & 0.014 & $-$0.260 & $6.20\mathrm{e}{+06}$ & 0.253 \\
\midrule
351--450  & aa\_sub\_1pct  & 0.866 &    0.235 & $1.05\mathrm{e}{+05}$ & 0.466 \\
          & aa\_sub\_2pct  & 0.760 & $-$0.287 & $1.05\mathrm{e}{+05}$ & 0.439 \\
          & aa\_sub\_5pct  & 0.589 & $-$0.498 & $2.15\mathrm{e}{+05}$ & 0.396 \\
          & aa\_sub\_10pct & 0.433 & $-$0.676 & $3.60\mathrm{e}{+05}$ & 0.358 \\
          & reverse       & 0.025 & $-$0.297 & $6.20\mathrm{e}{+06}$ & 0.256 \\
\midrule
451--550  & aa\_sub\_1pct  & 0.871 &    0.344 & $8.95\mathrm{e}{+04}$ & 0.466 \\
          & aa\_sub\_2pct  & 0.772 &    0.404 & $1.28\mathrm{e}{+05}$ & 0.443 \\
          & aa\_sub\_5pct  & 0.592 & $-$0.519 & $1.51\mathrm{e}{+05}$ & 0.395 \\
          & aa\_sub\_10pct & 0.449 & $-$0.608 & $2.68\mathrm{e}{+05}$ & 0.363 \\
          & reverse       & $-$0.010 & $-$0.230 & $6.83\mathrm{e}{+06}$ & 0.248 \\
\midrule
551--650  & aa\_sub\_1pct  & 0.854 &    0.428 & $8.68\mathrm{e}{+04}$ & 0.464 \\
          & aa\_sub\_2pct  & 0.752 & $-$0.429 & $1.46\mathrm{e}{+05}$ & 0.436 \\
          & aa\_sub\_5pct  & 0.579 & $-$0.315 & $1.12\mathrm{e}{+05}$ & 0.393 \\
          & aa\_sub\_10pct & 0.421 & $-$0.799 & $2.26\mathrm{e}{+05}$ & 0.355 \\
          & reverse       & 0.004 & $-$0.336 & $5.51\mathrm{e}{+06}$ & 0.250 \\
\midrule
651--800  & aa\_sub\_1pct  & 0.843 & $-$0.014 & $7.42\mathrm{e}{+04}$ & 0.460 \\
          & aa\_sub\_2pct  & 0.748 & $-$0.503 & $2.28\mathrm{e}{+05}$ & 0.436 \\
          & aa\_sub\_5pct  & 0.560 & $-$0.667 & $1.74\mathrm{e}{+05}$ & 0.390 \\
          & aa\_sub\_10pct & 0.423 & $-$0.751 & $2.22\mathrm{e}{+05}$ & 0.357 \\
          & reverse       & $-$0.039 & $-$0.245 & $6.18\mathrm{e}{+06}$ & 0.238 \\
\bottomrule
\end{tabular}%
}
\end{table}

\subsection{OpenFold}

OpenFold~\citep[Evoformer]{Ahdritz2024} SoloSeq mode (sequence-only, no MSA).
Since OpenFold has a single model size, the scaling axis is sequence length (100--800 aa).
ESM-1b embeddings are fed into the Evoformer trunk.
Composite stability improves modestly with sequence length (0.369 at $L$=100 to 0.420 at $L$=800),
consistent with the Evoformer's ability to build longer-range structural contacts.
OpenFold is the canonical Regime~II model (Representational Compression):
high MINE excess MI (+10.8 to +11.6 nats) but geometric warping (Procrustes disparity 0.15--0.16
relative to the ESM-1b input).

\paragraph{Synthetic proteins.}\mbox{}\\
\begin{table}[H]
\centering
\caption{OpenFold (SoloSeq) synthetic protein results across sequence lengths (100--800 aa). Stability improves with sequence length as the Evoformer builds longer-range structural contacts.}
\label{tab:openfold_synthetic}
\resizebox{\textwidth}{!}{%
\begin{tabular}{llrrrrr}
\toprule
\textbf{Seq. Length} & \textbf{Perturbation} & \textbf{RDM Sim.} & \textbf{Pert. Stab.} & \textbf{Pert. Mag.} & \textbf{Composite} \\
\midrule
L100 & aa\_sub\_1pct  & 0.824 & 0.034 & 0.019 & 0.441 \\
     & aa\_sub\_2pct  & 0.713 & 0.036 & 0.028 & 0.413 \\
     & aa\_sub\_5pct  & 0.534 & 0.056 & 0.042 & 0.368 \\
     & aa\_sub\_10pct & 0.352 & 0.053 & 0.036 & 0.323 \\
     & reverse       & 0.253 & 0.042 & 0.045 & 0.298 \\
\midrule
L200 & aa\_sub\_1pct  & 0.847 & 0.020 & 0.015 & 0.450 \\
     & aa\_sub\_2pct  & 0.742 & 0.033 & 0.021 & 0.424 \\
     & aa\_sub\_5pct  & 0.565 & 0.041 & 0.026 & 0.380 \\
     & aa\_sub\_10pct & 0.401 & 0.037 & 0.030 & 0.339 \\
     & reverse       & 0.314 & 0.027 & 0.038 & 0.317 \\
\midrule
L300 & aa\_sub\_1pct  & 0.865 & 0.025 & 0.012 & 0.456 \\
     & aa\_sub\_2pct  & 0.800 & 0.017 & 0.019 & 0.440 \\
     & aa\_sub\_5pct  & 0.621 & 0.030 & 0.023 & 0.395 \\
     & aa\_sub\_10pct & 0.459 & 0.025 & 0.029 & 0.355 \\
     & reverse       & 0.404 & 0.032 & 0.033 & 0.341 \\
\midrule
L400 & aa\_sub\_1pct  & 0.893 & 0.029 & 0.010 & 0.463 \\
     & aa\_sub\_2pct  & 0.819 & 0.028 & 0.015 & 0.444 \\
     & aa\_sub\_5pct  & 0.662 & 0.041 & 0.024 & 0.405 \\
     & aa\_sub\_10pct & 0.530 & 0.028 & 0.025 & 0.372 \\
     & reverse       & 0.444 & 0.028 & 0.030 & 0.350 \\
\midrule
L500 & aa\_sub\_1pct  & 0.909 & 0.024 & 0.010 & 0.467 \\
     & aa\_sub\_2pct  & 0.841 & 0.021 & 0.013 & 0.450 \\
     & aa\_sub\_5pct  & 0.679 & 0.043 & 0.021 & 0.410 \\
     & aa\_sub\_10pct & 0.522 & 0.034 & 0.023 & 0.370 \\
     & reverse       & 0.493 & 0.038 & 0.023 & 0.363 \\
\midrule
L600 & aa\_sub\_1pct  & 0.913 & 0.013 & 0.009 & 0.468 \\
     & aa\_sub\_2pct  & 0.853 & 0.016 & 0.011 & 0.453 \\
     & aa\_sub\_5pct  & 0.705 & 0.025 & 0.015 & 0.415 \\
     & aa\_sub\_10pct & 0.546 & 0.013 & 0.020 & 0.376 \\
     & reverse       & 0.521 & 0.024 & 0.023 & 0.369 \\
\midrule
L700 & aa\_sub\_1pct  & 0.921 & 0.028 & 0.008 & 0.471 \\
     & aa\_sub\_2pct  & 0.859 & 0.035 & 0.010 & 0.455 \\
     & aa\_sub\_5pct  & 0.709 & 0.035 & 0.017 & 0.418 \\
     & aa\_sub\_10pct & 0.567 & 0.021 & 0.023 & 0.382 \\
     & reverse       & 0.521 & 0.028 & 0.021 & 0.371 \\
\midrule
L800 & aa\_sub\_1pct  & 0.926 & 0.036 & 0.007 & 0.471 \\
     & aa\_sub\_2pct  & 0.858 & 0.033 & 0.010 & 0.454 \\
     & aa\_sub\_5pct  & 0.722 & 0.035 & 0.012 & 0.420 \\
     & aa\_sub\_10pct & 0.574 & 0.037 & 0.021 & 0.383 \\
     & reverse       & 0.536 & 0.055 & 0.024 & 0.373 \\
\bottomrule
\end{tabular}%
}
\end{table}

\paragraph{Real UniRef50 proteins.}\mbox{}\\
\begin{table}[H]
\centering
\caption{OpenFold (SoloSeq) real UniRef50 protein results across length bins (50--800 aa). $n$ indicates number of sequences per bin.}
\label{tab:openfold_uniref}
\resizebox{\textwidth}{!}{%
\begin{tabular}{lrrrrrr}
\toprule
\textbf{Length Bin} & \textbf{$n$} & \textbf{Perturbation} & \textbf{RDM Sim.} & \textbf{Pert. Stab.} & \textbf{Pert. Mag.} & \textbf{Composite} \\
\midrule
50--150   & 80198 & aa\_sub\_1pct  & 0.996 & 0.013 & 0.066 & 0.436 \\
          &       & aa\_sub\_2pct  & 0.995 & 0.017 & 0.100 & 0.435 \\
          &       & aa\_sub\_5pct  & 0.983 & 0.060 & 0.298 & 0.432 \\
          &       & aa\_sub\_10pct & 0.948 & 0.145 & 0.648 & 0.423 \\
          &       & reverse       & 0.394 & 0.847 & 4.155 & 0.285 \\
\midrule
151--250  & 90351 & aa\_sub\_1pct  & 0.999 & 0.017 & 0.051 & 0.431 \\
          &       & aa\_sub\_2pct  & 0.996 & 0.027 & 0.122 & 0.431 \\
          &       & aa\_sub\_5pct  & 0.987 & 0.081 & 0.333 & 0.428 \\
          &       & aa\_sub\_10pct & 0.964 & 0.183 & 0.693 & 0.423 \\
          &       & reverse       & 0.375 & 0.903 & 4.825 & 0.275 \\
\midrule
251--350  & 80686 & aa\_sub\_1pct  & 0.999 & 0.021 & 0.061 & 0.443 \\
          &       & aa\_sub\_2pct  & 0.997 & 0.033 & 0.136 & 0.442 \\
          &       & aa\_sub\_5pct  & 0.991 & 0.092 & 0.356 & 0.441 \\
          &       & aa\_sub\_10pct & 0.975 & 0.207 & 0.734 & 0.437 \\
          &       & reverse       & 0.476 & 0.920 & 5.111 & 0.312 \\
\midrule
351--450  & 64172 & aa\_sub\_1pct  & 0.999 & 0.018 & 0.073 & 0.452 \\
          &       & aa\_sub\_2pct  & 0.998 & 0.042 & 0.158 & 0.451 \\
          &       & aa\_sub\_5pct  & 0.991 & 0.120 & 0.409 & 0.450 \\
          &       & aa\_sub\_10pct & 0.973 & 0.247 & 0.821 & 0.445 \\
          &       & reverse       & 0.490 & 0.932 & 5.366 & 0.324 \\
\midrule
451--550  & 40740 & aa\_sub\_1pct  & 0.999 & 0.022 & 0.079 & 0.444 \\
          &       & aa\_sub\_2pct  & 0.997 & 0.043 & 0.168 & 0.444 \\
          &       & aa\_sub\_5pct  & 0.989 & 0.117 & 0.428 & 0.442 \\
          &       & aa\_sub\_10pct & 0.965 & 0.247 & 0.856 & 0.436 \\
          &       & reverse       & 0.481 & 0.931 & 5.307 & 0.315 \\
\midrule
551--650  & 22264 & aa\_sub\_1pct  & 0.999 & 0.026 & 0.080 & 0.442 \\
          &       & aa\_sub\_2pct  & 0.997 & 0.043 & 0.168 & 0.441 \\
          &       & aa\_sub\_5pct  & 0.988 & 0.116 & 0.428 & 0.439 \\
          &       & aa\_sub\_10pct & 0.964 & 0.250 & 0.857 & 0.433 \\
          &       & reverse       & 0.506 & 0.932 & 5.223 & 0.319 \\
\midrule
651--800  & 17678 & aa\_sub\_1pct  & 0.999 & 0.027 & 0.083 & 0.434 \\
          &       & aa\_sub\_2pct  & 0.997 & 0.043 & 0.175 & 0.433 \\
          &       & aa\_sub\_5pct  & 0.988 & 0.115 & 0.443 & 0.431 \\
          &       & aa\_sub\_10pct & 0.965 & 0.248 & 0.885 & 0.425 \\
          &       & reverse       & 0.480 & 0.922 & 5.002 & 0.304 \\
\bottomrule
\end{tabular}%
}
\end{table}

\subsection{Cross-Architecture Summary}

Table~\ref{tab:summary} presents mean composite stability across all perturbation types for each
model family, ordered by parameter count within each domain.
The key patterns visible in this summary:

\begin{enumerate}
\item \textbf{Progressive tax in Transformers:} ESM-2, NT, and SaProt all exhibit declining composite with scale.
\item \textbf{SSM stability:} Caduceus composites are near-constant across scale (0.459--0.458) with perfect RC preservation.
\item \textbf{Architectural class matters less than tokenization:} The cross-architecture spread within the continuous regime (Track~A) is 1.3$\times$; within the discrete regime (Track~B), it spans three orders of magnitude in Lipschitz constants.
\item \textbf{Structure-awareness is insufficient:} SaProt degrades at the same rate as ESM-2 at matched sizes, confirming that Foldseek structural tokens do not escape the tax.
\item \textbf{The 2.5B/15B ``rebound'' is Untethered Gel:} Both NT-2.5B and ESM2-15B show improved composites, but Procrustes Ratio and Frozen Head tests confirm these reflect global manifold drift, not genuine geometric recovery.
\end{enumerate}

\begin{table}[H]
\centering
\caption{Cross-architecture summary: mean composite stability (averaged across all perturbation types) on synthetic sequences. Models ordered by parameter count within each domain.}
\label{tab:summary}
\resizebox{0.85\textwidth}{!}{%
\begin{tabular}{llrrl}
\toprule
\textbf{Domain} & \textbf{Model} & \textbf{Size (M)} & \textbf{Mean Composite} & \textbf{Regime} \\
\midrule
\multirow{11}{*}{Protein}
 & ESM2-8M     & 7.5    & 0.431 & Brittle Glass \\
 & ESM2-35M    & 33.5   & 0.406 & Brittle Glass \\
 & SaProt-35M  & 33.9   & 0.436 & Brittle Glass \\
 & ProtMamba   & 107.7  & 0.381 & Geometric Vacuity \\
 & ESM2-150M   & 148.1  & 0.388 & Brittle Glass \\
 & ESM2-650M   & 651.0  & 0.360 & Brittle Glass \\
 & SaProt-650M & 652.1  & 0.431 & Brittle Glass \\
 & SaProt-1.3B & 1300.9 & 0.398 & Brittle Glass \\
 & ESM2-3B     & 2839.0 & 0.323 & Untethered Gel \\
 & ESM2-15B    & 15129.1& 0.378 & Untethered Gel \\
 & OpenFold$^*$& 75     & 0.369--0.420 & Repr. Compression \\
\midrule
\multirow{10}{*}{DNA}
 & Hyena-tiny  & 0.4    & 0.367 & --- \\
 & Caduceus-0.5M& 0.5   & 0.459 & Tax-exempt \\
 & Caduceus-1.9M& 1.9   & 0.449 & Tax-exempt \\
 & Hyena-small & 3.3    & 0.475 & --- \\
 & Hyena-medium& 6.6    & 0.479 & --- \\
 & Caduceus-7.7M& 7.7   & 0.458 & Tax-exempt \\
 & NT-50M      & 55.9   & 0.239 & Brittle Glass \\
 & GPN-Brass.  & 65.6   & 0.494 & --- \\
 & DNABERT-3mer& 86.1   & 0.425 & Brittle Glass \\
 & DNABERT-6mer& 89.2   & 0.433 & Brittle Glass \\
 & NT-100M     & 97.9   & 0.234 & Brittle Glass \\
 & DNABERT-2   & 117.1  & 0.323 & Brittle Glass \\
 & NT-250M     & 235.1  & 0.210 & Brittle Glass \\
 & NT-500M     & 498.3  & 0.194 & Brittle Glass \\
 & NT-2.5B     & 2547.8 & 0.336 & Untethered Gel \\
 & Evo 2 (8K)  & 6481.1 & 0.360 & Untethered Gel \\
 & Evo 2 (1M)  & 6581.7 & 0.380 & Untethered Gel \\
\bottomrule
\end{tabular}%
}

\vspace{0.3em}
{\footnotesize $^*$OpenFold range reflects sequence length scaling ($L$=100 to $L$=800); single model size.}
\end{table}

\subsection{Excluded models.}
We attempted to evaluate two additional models via their respective inference APIs:
Evo 2 (40B) through the NVIDIA NIM API and AlphaGenome~\citep{alphagenome} through its hosted endpoint. In both cases, the APIs return only final output predictions (next-token logits or
task-specific scores) rather than intermediate layer representations. Because our
evaluation framework requires access to internal embedding activations to compute
Procrustes stability, RDM similarity, and MINE mutual information, neither model
could be included in the geometric analysis. For Evo 2 40B specifically, the NIM API
returned near-zero activations consistent with FP8 quantization artifacts, rendering
the embeddings unsuitable even when partial access was available. We restrict the
Evo 2 analysis to the 7B family (8K, 262K, and 1M context variants) run locally
via the Vortex inference engine. We document these exclusions transparently and note
that future work with direct weight access could extend the analysis to these models.
\newpage

\section{Ghost Detection and Phase Transition Extended Methods and Details}
\label{app:ghost-detection}
This appendix provides extended methods and per-model results for the Ghost Detection protocol (Appendix~\ref{app:ghost-detection-protocol}) and the Brittle Glass / Untethered Gel phase transition proof (Appendix~\ref{app:phase-transition}), summarized in the main text (Sections~\ref{sec:scaling}-~\ref{sec:context-rc}). All experiments use seed 320 and 5-fold stratified cross-validation where applicable.

\subsection{Ghost Detection Protocol}
\label{app:ghost-detection-protocol}

The Ghost Detection protocol tests whether a model's apparent geometric stability reflects biologically grounded representations or a self-consistent manifold that is decoupled from downstream utility. The protocol comprises two complementary tests applied to each model.

\paragraph{Test 1: Self-Procrustes Spin Test.}
For each model configuration (parameter count, context length, or sequence length, depending on the scaling axis), we compute the Procrustes residual between clean and perturbed embedding matrices. Let $\mathbf{X}_c, \mathbf{X}_p \in \mathbb{R}^{n \times d}$ denote the mean-centered clean and perturbed embeddings, respectively. We solve for the optimal orthogonal alignment:
\begin{equation}
    \mathbf{R}^* = \arg\min_{\mathbf{R}^\top \mathbf{R} = \mathbf{I}} \left\| \hat{\mathbf{X}}_c - \hat{\mathbf{X}}_p \mathbf{R} \right\|_F^2,
\end{equation}
where $\hat{\mathbf{X}} = \mathbf{X} / \|\mathbf{X}\|_F$ denotes Frobenius-normalized matrices. The optimal rotation is obtained via the closed-form solution $\mathbf{R}^* = \mathbf{V}\mathbf{U}^\top$ from the SVD $\hat{\mathbf{X}}_p^\top \hat{\mathbf{X}}_c = \mathbf{U}\boldsymbol{\Sigma}\mathbf{V}^\top$ \citep{Schnemann1966,Rohlf1990,Masarotto2018,dryden1998statistical}. After rotation, we compute an optimal isotropic scale 
\begin{equation}s^* = \mathrm{tr}(\mathbf{X}_c^\top \mathbf{X}_p \mathbf{R}^*) / \mathrm{tr}((\mathbf{X}_p \mathbf{R}^*)^\top \mathbf{X}_p \mathbf{R}^*)\end{equation}
and report two quantities:

\begin{description}
\item[Procrustes Residual] The per-sample Frobenius distance after alignment: \[\|\mathbf{X}_c - s^* \mathbf{X}_p \mathbf{R}^*\|_F / \sqrt{n}\]
\item[Procrustes Ratio] The fraction of raw error surviving alignment: $\text{aligned\_error} / \text{raw\_error}$, where \[\text{raw\_error} = \|\mathbf{X}_c - \mathbf{X}_p\|_F / \sqrt{n}\]
\end{description}

A \emph{Ghost} is diagnosed when the Procrustes residual \emph{increases} with scale (the manifold ``spins'' more freely in the latent space at larger configurations) or remains high despite ostensibly stable RDM similarity. When embedding matrices exceed 10{,}000 samples, we subsample uniformly at random (seed 320) to $n = 5{,}000$ (Evo~2) or $n = 10{,}000$ (ProtMamba, OpenFold).

\paragraph{Test 2: Frozen Head Context Tax.}
We train a frozen linear classifier (logistic regression, $C = 1.0$, max 1{,}000 iterations) on model embeddings for a binary species-classification task that requires only \emph{local} sequence information. The signal region is held constant while the total input length increases via random padding, creating a controlled dilution curve. If accuracy degrades with longer input despite the signal being unchanged, the model's geometric stability does not translate to biologically useful representations.

For DNA models (Evo~2), we classify real human chr22 vs.\ real \emph{S.\ cerevisiae} DNA using a 1~kbp signal region, with 200 sequences per class. For protein models (ProtMamba, OpenFold), we classify real \emph{E.\ coli} vs.\ real human proteins (UniProt Swiss-Prot, reviewed) using a 50~aa signal region (first 50 residues of each protein), with 200 sequences per class. Random padding consists of i.i.d.\ uniform nucleotides (DNA) or amino acids (protein). We report 5-fold stratified cross-validated accuracy.

Two pooling strategies are tested: \textbf{local} (pool only the signal region of the embedding) and \textbf{global} (pool the full sequence). Local pooling isolates whether the model preserves signal-region geometry; global pooling tests whether padding tokens corrupt the entire representation.

\paragraph{Embedding extraction.}
Evo~2 embeddings are extracted from layer \texttt{blocks.28.mlp.l3} via local Vortex inference on A100 GPUs, with a safe sequence-length cap of 120{,}000 tokens (avoiding 32-bit index overflow in the FIR engine). ProtMamba embeddings are extracted from the final hidden layer of the 107.7M Mamba model. OpenFold embeddings are extracted from the Evoformer trunk output (single-sequence SoloSeq mode using ESM-1b as the sequence encoder), with a maximum sequence length of 1{,}022 residues. In all cases, per-token hidden states are mean-pooled over the relevant region.

\subsection{Per-Model Ghost Detection Results}
\label{app:ghost-per-model}

\paragraph{Evo~2 (7B, StripedHyena~2).}
The scaling axis is context-window length (8K, 262K, 1M) at fixed 7B parameters. The Spin Test reveals that the Procrustes ratio is remarkably stable across context windows for SNP perturbations: 0.963 (8K), 0.964 (262K), 0.963 (1M) at 1\% SNP on synthetic data. Reverse complement shows slightly more variation but no systematic trend: ratio 0.744 (8K), 0.740 (262K), 0.732 (1M). The Ghost is therefore \emph{neutral}: context-window scaling neither grounds nor detaches the manifold orientation.

The Frozen Head test confirms diminishing returns: accuracy at 8K context is comparable to accuracy at 1M context ($\Delta = +0.005$, effectively zero), meaning 128$\times$ more context yields no classification gain. On real chr22 DNA, SNP Procrustes ratios are even more stable (0.984--0.985 across all three checkpoints at 1\% SNP), reflecting the lower perturbation magnitude on real genomic sequences.

\begin{table}[H]
\centering
\caption{\textbf{Evo~2 Frozen Head test:} frozen linear classifier accuracy (Human chr22 vs.\ Yeast, 1~kbp signal, local pooling).}
\label{tab:evo2-frozen-head}
\begin{tabular}{lrcc}
\toprule
\textbf{Checkpoint} & \textbf{Context} & \textbf{Accuracy} & \textbf{$\Delta$ from 8K} \\
\midrule
Evo2-7B-8K   & 8{,}192      & $0.850 \pm 0.02$ & -- \\
Evo2-7B-262K & 262{,}144    & $0.843 \pm 0.02$ & $-0.008$ \\
Evo2-7B-1M   & 1{,}048{,}576 & $0.855 \pm 0.02$ & $+0.005$ \\
\bottomrule
\end{tabular}
\end{table}

\paragraph{Evo~2 Frozen Head (next-token prediction).}
The phase transition proof extends the Frozen Head test to the model's own next-token prediction head. Agreement between clean-input and perturbed-input top-1 predictions is stable across context windows: 85.3\% (8K), 84.9\% (262K), 85.2\% (1M) at 1\% SNP. KL divergence between output distributions is similarly flat (0.0024--0.0027 nats). Under reverse complement, agreement drops to $\sim$28--30\% across all checkpoints, confirming that the RC perturbation produces a categorically different output distribution regardless of context length.

\paragraph{ProtMamba (108M, Mamba SSM).}
The scaling axis is sequence length (100--800~aa) at fixed 108M parameters. The Spin Test on synthetic data shows Procrustes ratios declining from 0.950 (L=100) to 0.819 (L=800) at 10\% substitution, indicating that longer sequences push the manifold toward Untethered Gel. However, the critical finding is the Frozen Head: accuracy is at \emph{chance level} ($\sim$50\%) at every sequence length, including the shortest (L=100, 50\% signal fraction). Both local and global pooling yield identical chance-level performance.

\begin{table}[H]
\centering
\caption{\textbf{ProtMamba Ghost Detection:} Frozen Head accuracy (E.\ coli vs.\ Human, 50~aa signal, local pooling). Accuracy is at chance for all lengths.}
\label{tab:protmamba-ghost-frozen}
\begin{tabular}{rrcc}
\toprule
\textbf{Length} & \textbf{Signal \%} & \textbf{Accuracy} & \textbf{Verdict} \\
\midrule
100 & 50\% & $0.498 \pm 0.009$ & Uninformative \\
200 & 25\% & $0.495 \pm 0.010$ & Uninformative \\
400 & 12.5\% & $0.500 \pm 0.016$ & Uninformative \\
800 & 6.25\% & $0.498 \pm 0.012$ & Uninformative \\
\bottomrule
\end{tabular}
\end{table}

This establishes ProtMamba's Geometric Vacuity: the manifold is geometrically stable (low Procrustes distortion, reasonable perplexity of 40.1) but contains no extractable biological information. The nonlinear probe validation (Section~\ref{app:nonlinear-probe}) confirms that this is genuine uninformativeness, not information hidden on a curved manifold.

\paragraph{OpenFold (SoloSeq, ESM-1b + Evoformer).}
The scaling axis is sequence length (100--800~aa). The Spin Test reveals a \emph{grounding} pattern: the reverse Procrustes ratio increases from 0.830 (L=100) to 0.895 (L=800), while the residual decreases from 0.961 to 0.618. Longer sequences produce more orientationally stable embeddings. Under amino acid substitution, the pattern is similar but weaker (ratio 0.963 to 0.983 at 1\% substitution from L=100 to L=800).

The Frozen Head tells a different story. Accuracy starts high (0.843 at L=100) and degrades systematically: 0.820 (L=200), 0.780 (L=400), 0.773 (L=800). The context tax is toxic: 8$\times$ more input length produces a 7.0 percentage-point accuracy drop.

\begin{table}[H]
\centering
\caption{\textbf{OpenFold Ghost Detection:} Frozen Head accuracy (E.\ coli vs.\ Human, 50~aa signal, local pooling). Accuracy degrades with sequence length.}
\label{tab:openfold-ghost-frozen}
\begin{tabular}{rrcc}
\toprule
\textbf{Length} & \textbf{Signal \%} & \textbf{Accuracy} & \textbf{$\Delta$ from L=100} \\
\midrule
100 & 50\% & $0.843 \pm 0.020$ & -- \\
200 & 25\% & $0.820 \pm 0.023$ & $-0.023$ \\
400 & 12.5\% & $0.780 \pm 0.019$ & $-0.063$ \\
800 & 6.25\% & $0.773 \pm 0.040$ & $-0.070$ \\
\bottomrule
\end{tabular}
\end{table}

\subsection{Nonlinear Probe Validation}
\label{app:nonlinear-probe}

The Ghost Detection Frozen Head uses a linear classifier, raising the question of whether biological information might be encoded in a nonlinear geometric format inaccessible to linear probes. We run the identical frozen head protocol with three classifiers in parallel:

\begin{description}
\item[Linear] Logistic regression ($C = 1.0$, max 1{,}000 iterations).
\item[MLP] 2-layer neural network (hidden sizes 256, 64; ReLU; early stopping with 15\% validation split, patience 20; learning rate $10^{-3}$).
\item[MLP-Wide] 3-layer neural network (hidden sizes 512, 256, 64; same training protocol).
\end{description}

All probes are evaluated with 5-fold stratified cross-validation (seed 320). The diagnostic logic is: if the MLP substantially outperforms the linear probe, biological information exists but is encoded on a curved manifold; if both are at chance, the representation is genuinely uninformative.

\paragraph{ProtMamba: Uninformative confirmed.}
Both linear and MLP probes achieve chance-level accuracy ($\sim$50\%) at all sequence lengths under both local and global pooling (Table~\ref{tab:protmamba-nonlinear}). The average gap between the best MLP and the linear probe is $< 0.01$ across all conditions. ProtMamba's embeddings do not encode species-level biological information in \emph{any} geometric format, confirming the Geometric Vacuity diagnosis.

\begin{table}[H]
\centering
\caption{\textbf{ProtMamba nonlinear probe validation} (local pooling, E.\ coli vs.\ Human). All probes are at chance.}
\label{tab:protmamba-nonlinear}
\begin{tabular}{rcccc}
\toprule
\textbf{Length} & \textbf{Linear} & \textbf{MLP} & \textbf{MLP-Wide} & \textbf{$\Delta$ (best)} \\
\midrule
100 & $0.498$ & $0.505$ & $0.500$ & $+0.008$ \\
200 & $0.495$ & $0.495$ & $0.505$ & $+0.010$ \\
400 & $0.500$ & $0.490$ & $0.500$ & $+0.000$ \\
800 & $0.498$ & $0.500$ & $0.500$ & $+0.003$ \\
\bottomrule
\end{tabular}
\end{table}

\paragraph{OpenFold: Linear encoding with length-dependent degradation.}
Under local pooling, both linear and MLP probes start well above chance and degrade with sequence length (Table~\ref{tab:openfold-nonlinear}). At L=100, the MLP achieves 0.863 vs.\ the linear probe's 0.843 ($\Delta = +0.020$). At L=400, the gap widens: MLP 0.833 vs.\ linear 0.780 ($\Delta = +0.053$). However, at L=800, the MLP also degrades (0.750), converging toward the linear probe (0.773). The decline-rate analysis reveals that under local pooling, the linear probe declines significantly ($r = -0.972$, $p = 0.028$) while the MLP trend does not reach significance ($r = -0.903$, $p = 0.097$); however, the slope difference is not statistically distinguishable ($z = -0.74$, $p = 0.46$, $n = 4$ lengths).

Under global pooling, both probes decline at statistically similar rates ($p_{\text{diff}} = 0.60$), confirming representational toxicity: subspace compression irreversibly degrades the biological signal when the full sequence is pooled.

\begin{table}[H]
\centering
\caption{\textbf{OpenFold nonlinear probe validation} (local pooling, E.\ coli vs.\ Human). Both probes degrade with length, but MLP retains a modest advantage at intermediate lengths.}
\label{tab:openfold-nonlinear}
\begin{tabular}{rcccc}
\toprule
\textbf{Length} & \textbf{Linear} & \textbf{MLP} & \textbf{MLP-Wide} & \textbf{$\Delta$ (best)} \\
\midrule
100 & $0.843$ & $0.863$ & $0.855$ & $+0.020$ \\
200 & $0.820$ & $0.835$ & $0.810$ & $+0.015$ \\
400 & $0.780$ & $0.833$ & $0.815$ & $+0.053$ \\
800 & $0.773$ & $0.750$ & $0.735$ & $-0.023$ \\
\bottomrule
\end{tabular}
\end{table}

\paragraph{Three-way architectural comparison.}
The nonlinear probe results complete the diagnostic picture across architecture classes:

\begin{table}[H]
\centering
\caption{\textbf{Cross-architecture probe comparison} at the shortest sequence length (highest signal fraction). Transformer = linearly accessible; SSM = informationally empty.}
\label{tab:cross-arch-probe}
\begin{tabular}{llccc}
\toprule
\textbf{Architecture} & \textbf{Model} & \textbf{Linear} & \textbf{Best MLP} & \textbf{Diagnosis} \\
\midrule
Transformer & OpenFold & $0.843$ & $0.863$ & Linear encoding \\
SSM         & ProtMamba & $0.498$ & $0.505$ & Uninformative \\
\bottomrule
\end{tabular}
\end{table}

OpenFold encodes species identity in a linearly accessible format that degrades with padding length. ProtMamba's Geometric Vacuity is confirmed: its stable manifold geometry carries no detectable biological content under either linear or nonlinear readout.

\subsection{Phase Transition Proof: Brittle Glass vs.\ Untethered Gel}
\label{app:phase-transition}

The phase transition proof extends across multiple architectures and modalities, testing whether the Brittle Glass / Untethered Gel dichotomy is universal or architecture-specific. For each model, we compute the Procrustes Ratio (Section~\ref{app:ghost-detection}) and the Frozen Head agreement and KL divergence using the model's own language-modeling head.

\paragraph{Procrustes Ratio definition.}
The Procrustes Ratio $\rho = \text{aligned\_error} / \text{raw\_error}$ measures how much of the perturbation-induced distortion is removable by rigid-body alignment. A ratio near 1.0 indicates internal fracture (\emph{Brittle Glass}: no rotation helps because the error is distributed nonlinearly throughout the manifold). A ratio near 0 indicates global drift (\emph{Untethered Gel}: a single rotation removes most of the error because the manifold shifted as a coherent block).

\paragraph{Frozen Head definition.}
For models with a language-modeling head (masked or causal), we compute token-level predictions on both clean and perturbed input. Two metrics quantify functional drift: (1) \textbf{Agreement}: fraction of positions where the top-1 predicted token matches between clean and perturbed; (2) \textbf{KL divergence}: mean token-level $D_{\mathrm{KL}}(p_{\text{clean}} \| p_{\text{pert}})$, computed over the full vocabulary distribution. High KL divergence despite preserved RDM similarity is the diagnostic signature of Untethered Gel: the internal pairwise structure is preserved, but the manifold has drifted in absolute coordinates, rendering the language-modeling head's decision boundaries incorrect.

\paragraph{Cross-architecture Procrustes Ratio results.}
Table~\ref{tab:phase-procrustes} reports the mean Procrustes ratio across perturbation types for selected model sizes. The Brittle Glass / Untethered Gel transition is observed in both protein (ESM-2) and DNA (NT, DNABERT) Transformer families, with the transition point varying by architecture. Caduceus (Mamba SSM) exhibits uniformly low reduction ($< 1\%$ on SNPs), consistent with its near-equivariant design. Evo~2 shows moderate ratios (0.73--0.96) that are stable across context windows, placing it in the Brittle Glass regime for RC perturbations and near-Brittle for SNPs.

\begin{table}[H]
\centering
\caption{\textbf{Cross-architecture Procrustes Ratio} (reverse/RC perturbation). Lower ratio $=$ more error removable by rotation $=$ more Untethered Gel.}
\label{tab:phase-procrustes}
\begin{tabular}{llcc}
\toprule
\textbf{Architecture} & \textbf{Model} & \textbf{Ratio (RC/Rev)} & \textbf{Reduction \%} \\
\midrule
Transformer & ESM2-8M      & $0.949$ & $5.1\%$ \\
Transformer & ESM2-650M    & $0.864$ & $13.6\%$ \\
Transformer & ESM2-3B      & $0.803$ & $19.7\%$ \\
Transformer & ESM2-15B     & $0.738$ & $26.2\%$ \\
\midrule
Transformer & DNABERT-3mer & $0.618$ & $38.2\%$ \\
Transformer & DNABERT-6mer & $0.360$ & $64.0\%$ \\
\midrule
Transformer & NT-50M       & $0.757$ & $24.3\%$ \\
Transformer & NT-2.5B      & $0.740$ & $26.0\%$ \\
\midrule
Hybrid      & HyenaDNA     & $0.557$ & $44.3\%$ \\
\midrule
SSM         & Caduceus-0.5M & $0.252$ & $74.8\%$ \\
SSM         & Caduceus-7.7M & $0.252$ & $74.8\%$ \\
\midrule
Hybrid      & Evo2-7B-8K (synth)  & $0.744$ & $25.6\%$ \\
Hybrid      & Evo2-7B-1M (synth)  & $0.732$ & $26.8\%$ \\
Hybrid      & Evo2-7B-8K (real)   & $0.762$ & $23.8\%$ \\
Hybrid      & Evo2-7B-1M (real)   & $0.779$ & $22.1\%$ \\
\bottomrule
\end{tabular}
\end{table}

\paragraph{Evo~2 Frozen Head (per-perturbation).}
The Frozen Head test on Evo~2's own next-token prediction head is reported in Table~\ref{tab:evo2-frozen-full}. Agreement is high for SNPs ($> 85\%$ at 1\%) and degrades proportionally with perturbation magnitude, reaching $\sim$59\% at 10\% SNP. Reverse complement drops agreement to $\sim$28--30\%, with KL divergence an order of magnitude higher than SNPs. Critically, none of these metrics vary systematically with context-window length, confirming that context scaling does not rescue the geometric tax.

\begin{table}[H]
\centering
\caption{\textbf{Evo~2 Frozen Head:} next-token agreement and KL divergence across context windows and perturbation types.}
\label{tab:evo2-frozen-full}
\begin{tabular}{llcc}
\toprule
\textbf{Checkpoint} & \textbf{Perturbation} & \textbf{Agreement} & \textbf{KL (nats)} \\
\midrule
Evo2-7B-8K   & SNP 1\%  & 0.853 & 0.0024 \\
Evo2-7B-8K   & SNP 5\%  & 0.677 & 0.0082 \\
Evo2-7B-8K   & SNP 10\% & 0.590 & 0.0128 \\
Evo2-7B-8K   & RC       & 0.276 & 0.0342 \\
\midrule
Evo2-7B-262K & SNP 1\%  & 0.849 & 0.0027 \\
Evo2-7B-262K & SNP 5\%  & 0.685 & 0.0085 \\
Evo2-7B-262K & SNP 10\% & 0.607 & 0.0122 \\
Evo2-7B-262K & RC       & 0.296 & 0.0328 \\
\midrule
Evo2-7B-1M   & SNP 1\%  & 0.852 & 0.0025 \\
Evo2-7B-1M   & SNP 5\%  & 0.691 & 0.0081 \\
Evo2-7B-1M   & SNP 10\% & 0.599 & 0.0127 \\
Evo2-7B-1M   & RC       & 0.301 & 0.0330 \\
\bottomrule
\end{tabular}
\end{table}

\paragraph{Interpretation.}
The cross-architecture results confirm that the Brittle Glass / Untethered Gel transition is a general phenomenon of discrete Transformer scaling, not an artifact of any single model family. ESM-2 traces the clearest arc from Brittle Glass (8M: 5.1\% reduction) through a transition zone (650M: 13.6\%) to Untethered Gel (15B: 26.2\%). The NT family shows a similar but flatter progression. DNABERT's k-mer tokenization produces unusually high Procrustes reduction (38--64\%), suggesting that coarser tokenization pushes models further into the Gel regime. Caduceus's consistently high reduction ($\sim$75\%) with near-perfect RDM preservation represents a qualitatively different pattern: the RC-equivariant SSM architecture produces drift that is almost entirely removable by rotation, with negligible internal distortion.

Evo~2 occupies a stable intermediate position (22--27\% reduction) that does not vary with context length, consistent with the Ghost Detection finding that context-window scaling is geometrically neutral for this architecture. The combination of moderate Procrustes reduction with flat Frozen Head accuracy across context windows indicates that the StripedHyena~2 architecture's multi-hybrid design (interleaved Hyena convolutions and attention layers) produces a fixed-magnitude geometric tax that context scaling cannot amortize.

\subsection{Procrustes Centering and Normalization}
\label{app:procrustes-centering}

All Procrustes analyses in this work apply the following preprocessing to each embedding matrix $\mathbf{X} \in \mathbb{R}^{n \times d}$ before alignment:

\begin{enumerate}
\item \textbf{Mean centering:} $\mathbf{X}_c = \mathbf{X} - \bar{\mathbf{x}}$, where $\bar{\mathbf{x}} = \frac{1}{n}\sum_{i=1}^n \mathbf{x}_i$.
\item \textbf{Frobenius normalization:} $\hat{\mathbf{X}} = \mathbf{X}_c / \|\mathbf{X}_c\|_F$ (for computing the optimal rotation).
\item \textbf{Optimal scale:} After rotation, a scalar $s^*$ is computed to minimize the aligned Frobenius distance (see Section~\ref{app:ghost-detection}).
\end{enumerate}

The raw error is computed between mean-centered (but not Frobenius-normalized) matrices. The aligned error uses the mean-centered matrices after rotation by $\mathbf{R}^*$ and scaling by $s^*$. This ensures that the Procrustes Ratio is invariant to differences in overall embedding magnitude between clean and perturbed representations, while preserving sensitivity to shape differences.

When embeddings have very different norms (as can occur under large perturbation magnitudes), the optimal scale $s^*$ absorbs the magnitude difference, and the residual reflects purely geometric (shape) distortion. We report $s^*$ alongside all Procrustes results to flag cases where large scale adjustments might indicate global gain/attenuation rather than geometric restructuring.
\newpage

\section{RCCR Experiment Details}
\label{app:rccr}
We test whether post-hoc reverse-complement consistency regularization 
mitigates the geometric tax on a discrete CE backbone. Following 
Ma~\citep{ma2025}, we implement an embedding-level variant of RCCR 
that minimizes L2 distance between mean-pooled representations of 
forward and RC sequences, adapting the task-level consistency 
objective to an unsupervised setting. We deep-copy DNABERT-2 (117M), 
fine-tune for 10 epochs (LR $2 \times 10^{-5}$, $\lambda = 1.0$) on 
2{,}000 random DNA sequences (seed 999), and evaluate both baseline 
and RCCR-finetuned models on a held-out set of 5{,}000 sequences 
(seed 320) using the Shesha harness.

RCCR training loss drops from $1.56 \times 10^{-3}$ to $9 \times 
10^{-6}$ ($99.4\%$ reduction), and per-sequence RC cosine gap falls 
from $0.041$ to $0.000$: every sequence now maps to the same point as 
its reverse complement.

\begin{figure}[H]
  \centering
  \includegraphics[width=\textwidth]{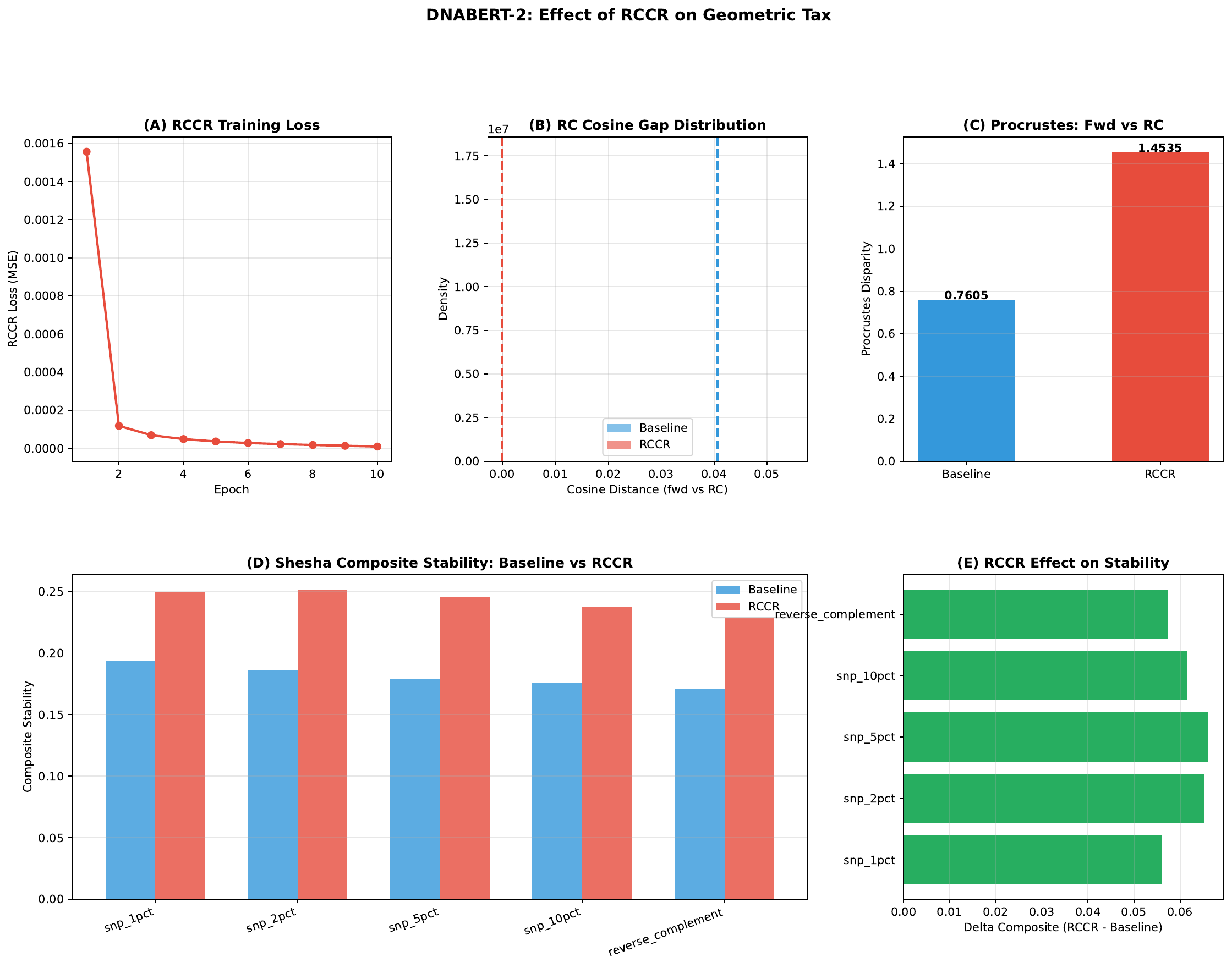}
  \caption{\footnotesize
    Effect of embedding-level RCCR on DNABERT-2 (117M).
    \textbf{(A)}~Training loss converges rapidly ($99.4\%$ reduction 
    in 10 epochs).
    \textbf{(B)}~Per-sequence RC cosine gap collapses from $0.041$ to 
    $0.000$: perfect pointwise consistency.
    \textbf{(C)}~Despite this, Procrustes disparity between forward 
    and RC embedding matrices \textit{increases} $91\%$ ($0.76 \to 
    1.45$): population-level geometric structure degrades.
    \textbf{(D)}~Shesha composite stability by perturbation type. 
    RCCR composites are marginally higher, driven by improved 
    feature-split scores; RDM similarity and perturbation magnitude 
    both degrade (Table~\ref{tab:rccr}).
    \textbf{(E)}~Delta composite (RCCR minus baseline). Uniform 
    positive shift masks underlying geometric deterioration. RCCR 
    achieves consistency by flattening perturbation sensitivity, not 
    by aligning manifold geometry.%
  }
  \label{fig:rccr}
\end{figure}

\begin{table}[H]
  \centering
  \small
  \caption{RCCR vs.\ baseline on DNABERT-2 (117M). RCCR achieves 
  perfect per-sequence RC consistency but degrades population-level 
  geometric structure across all perturbation conditions.}
  \label{tab:rccr}
  \begin{tabular}{@{}llcccc@{}}
    \toprule
    Model & Perturbation & RDM Sim. & Pert.\ Stab. & Pert.\ Mag. 
      & Composite \\
    \midrule
    Baseline & SNP 1\%  & 0.113 & 0.038 & 0.124 & 0.194 \\
    RCCR     & SNP 1\%  & 0.050 & 0.040 & 0.000 & 0.250 \\
    \midrule
    Baseline & SNP 2\%  & 0.081 & 0.039 & 0.118 & 0.186 \\
    RCCR     & SNP 2\%  & 0.054 & 0.042 & 0.000 & 0.251 \\
    \midrule
    Baseline & SNP 5\%  & 0.054 & 0.041 & 0.127 & 0.179 \\
    RCCR     & SNP 5\%  & 0.032 & 0.035 & 0.000 & 0.245 \\
    \midrule
    Baseline & SNP 10\% & 0.041 & 0.041 & 0.129 & 0.176 \\
    RCCR     & SNP 10\% & 0.001 & 0.044 & 0.000 & 0.238 \\
    \midrule
    Baseline & RC       & 0.021 & 0.041 & 0.129 & 0.171 \\
    RCCR     & RC       & $-0.036$ & 0.038 & 0.000 & 0.228 \\
    \bottomrule
  \end{tabular}
\end{table}
Procrustes disparity between forward and RC embedding matrices 
increases from $0.761$ to $1.454$ ($+91\%$). The composite score 
rises slightly, driven entirely by feature-split improvement ($0.66 
\to 0.96$); the metrics that measure geometric fidelity (RDM 
similarity, perturbation magnitude) degrade uniformly. The model 
achieves consistency by collapsing perturbation sensitivity rather 
than by aligning manifold geometry, consistent with the 
rate-distortion prediction that capacity allocated to one constraint 
is unavailable for manifold preservation.
\newpage

\section{MINE Extended Methods and Details}
\label{app:mine}
This appendix provides full methodology for the Mutual Information Neural Estimation (MINE) experiments reported in Section~\ref{sec:info-theory}. All code is available as Jupyter notebooks: one per model (\texttt{MINE\_01\_ProtMamba}, \texttt{MINE\_02\_OpenFold}, \texttt{MINE\_03\_Evo2}) plus a combined analysis notebook (\texttt{MINE\_04\_Combined\_Analysis}).

\paragraph{Estimator.}
We use the Donsker--Varadhan variational lower bound~\citep{Donsker1983,Belghazi2018MutualIN}:
\begin{equation}
  I(X; \hat{X}) \;\geq\; \mathbb{E}_{p(x,\hat{x})}[T_\theta(x,\hat{x})]
  \;-\; \log \mathbb{E}_{p(x)p(\hat{x})}[e^{T_\theta(x,\hat{x})}],
  \label{eq:dv_bound}
\end{equation}
where $T_\theta$ is a learned statistics network. $X$ denotes biological ground-truth features and $\hat{X}$ denotes model embeddings projected via PCA (see below).

\paragraph{Statistics network.}
$T_\theta$ is a three-layer MLP: $(\mathrm{input}) \to 256 \to 128 \to 1$, with ReLU activations and 10\% dropout after each hidden layer. All inputs (both $X$ and $\hat{X}$) are independently z-scored (per-feature mean subtraction and unit-variance scaling) before concatenation. The network is trained with Adam (learning rate $10^{-4}$, no weight decay) for 500 epochs, with gradient clipping at $\|\nabla\|_\infty = 5.0$ and minibatch size 64. The final MI estimate is the mean over the last 10\% of training epochs to reduce convergence noise.

\paragraph{Confidence intervals.}
Each experiment is repeated 5 times with seeds 320, 420, 520, 620, and 720. We report $\mathrm{mean} \pm \mathrm{std}$ across runs. The same seed sequence is used for the statistics network initialization, NumPy random state, and minibatch permutation within each run.

\paragraph{PCA pre-processing.}
Raw model embeddings (dimensionality ranging from 768 to 4,096 depending on the model) are projected to 50 principal components before MINE estimation. This stabilizes the Donsker--Varadhan bound in high dimensions, where finite-sample bias grows with input dimensionality. PCA is fit with \texttt{sklearn.decomposition.PCA} (random state 320). The number of retained components is $\min(50, N, d)$ where $N$ is the sample count and $d$ is the embedding dimension.

\paragraph{Bias correction.}
MINE has well-documented finite-sample upward bias that scales with input dimensionality and is independent of the true MI~\citep{Belghazi2018MutualIN}. We control for this by computing a matched random baseline for each ground-truth distribution: $\mathrm{MI}_{\mathrm{random}} = \mathrm{MINE}(X, Z)$ where $Z \sim \mathcal{N}(0, I_{50})$ has the same dimensionality as the PCA-reduced embeddings but is statistically independent of $X$. The random baselines are $2.664 \pm 0.054$ nats (protein) and $1.681 \pm 0.033$ nats (DNA). All reported values are excess MI:
\begin{equation}
  \mathrm{MI}_{\mathrm{excess}} \;=\; \mathrm{MI}_{\mathrm{model}} - \mathrm{MI}_{\mathrm{random}}.
\end{equation}
Negative excess indicates that the model's embeddings carry less structure, with respect to the ground-truth features, than random noise at the same dimensionality.

\paragraph{MI ceiling calibration.}
We establish an upper bound by running MINE with the ground-truth features as both input and target, corrupted by small Gaussian noise: $\mathrm{MI}_{\mathrm{ceiling}} = \mathrm{MINE}(X, X + \epsilon)$, $\epsilon \sim \mathcal{N}(0, 0.1^2 I)$. This calibrates the maximum MI the estimator can recover given the sample size and feature dimensionality, and is used to compute normalized MI values reported in Figure~\ref{fig:mine}.

\paragraph{Sanity check.}
Before model evaluation, we validate the estimator on correlated bivariate Gaussians with known MI. For correlation $\rho \in \{0.0, 0.3, 0.6, 0.9\}$, we draw $N = 2{,}000$ samples from
$x \sim \mathcal{N}(0,1)$, $y = \rho x + \sqrt{1 - \rho^2}\,z$, $z \sim \mathcal{N}(0,1)$,
where $I(x;y) = -\tfrac{1}{2}\log(1-\rho^2)$. The sanity check uses a smaller network ($128 \to 64 \to 1$) and 300 epochs. All four conditions pass within tolerance ($|{\hat{I} - I}| < \max(0.15, 0.3\,I)$).

\subsection{Biological Ground Truth}

\paragraph{Protein features (Regimes II and III).}
We download 200 reviewed Swiss-Prot sequences per species ({\itshape E.~coli}, organism ID 83333; {\itshape H.~sapiens}, organism ID 9606) from the UniProt REST API~\citep{Suzek2007}, filtering for sequences with at least 50 residues composed entirely of standard amino acids. The first 50 residues of each sequence constitute the biological signal region; for experiments at sequence lengths $L \in \{100, 200, 400\}$, the signal is padded with random amino acids to the target length.

The ground-truth feature vector $X \in \mathbb{R}^{25}$ consists of: 20 amino acid frequencies (computed over the full signal region), normalized sequence length ($L / 1000$), net charge per residue ($(n_K + n_R - n_D - n_E)/L$), mean Kyte--Doolittle hydrophobicity, and a 2-dimensional one-hot species indicator. Total sample size is $N = 400$ (200 per species).

\paragraph{DNA features (Regime I).}
We fetch 200 real human genomic regions (8,192~bp each) from the UCSC Genome Browser API (hg38 assembly), sampling uniformly across autosomes chr1--chr22 with 10\% telomeric margin exclusion. Sequences containing $>5\%$ ambiguous bases are discarded and re-sampled.

The global ground-truth feature vector $X_{\mathrm{global}} \in \mathbb{R}^{17}$ consists of: GC content (scalar) and 16 dinucleotide frequencies (counts of AA, AC, AG, \ldots, TT normalized by sequence length minus one). For local MI estimation, the same 17 features are computed within each 128-bp window independently: $X_{\mathrm{local}} = \mathrm{DNAfeatures}(\mathrm{seq}[s\!:\!s\!+\!128])$.

\subsubsection{Model-Specific Embedding Extraction}

\paragraph{ProtMamba (Regime III).}
Embeddings are extracted from the final Mamba layer using the \texttt{save\_layer} interface. Sequences are tokenized with the ProtMamba amino acid vocabulary. For each sequence, the embedding is mean-pooled over the first 50 token positions (the signal region) to prevent padding noise from contaminating the estimate. The model runs in float16 for embedding extraction. Perplexity (the task competence control) is computed separately in float32, because the language model head produces NaN logits in float16 due to overflow in the final linear projection.

\paragraph{OpenFold Evoformer and ESM-1b (Regime II).}
OpenFold is run in SoloSeq (sequence-only) mode, using ESM-1b (33-layer, 650M parameters) as the trunk encoder. We extract embeddings from two points in the forward pass: (i) the ESM-1b single representation (layer 33, pre-Evoformer), and (ii) the Evoformer single-track output (post-Evoformer). Both are mean-pooled over the first 50 token positions. The \texttt{flash\_attn} module hierarchy is mocked in \texttt{sys.modules} prior to any OpenFold import to avoid ABI mismatch errors. Procrustes alignment between the ESM-1b and Evoformer embedding spaces is computed at each sequence length to quantify the geometric distortion introduced by the Evoformer block.

\paragraph{Evo 2 (Regime I).}
Evo 2 (7B parameters, StripedHyena-2 architecture) is loaded via the Vortex checkpoint (\texttt{evo2\_7b}). Embeddings are extracted from layer \texttt{blocks.28.mlp.l3} with \texttt{return\_embeddings=True}. Two embedding modes are used:

\begin{itemize}
  \item \textbf{Global MI.} Embeddings are mean-pooled across the full 8,192-token context. MINE is run against $X_{\mathrm{global}}$ (GC content + 16 dinucleotide frequencies for the whole sequence).
  \item \textbf{Local MI.} Per-position embeddings are extracted for three 128-bp windows at positions 500, 2000, and 4000. For each window, the embedding is mean-pooled over the 128 positions within that window only, and MINE is run against $X_{\mathrm{local}}$ computed for the corresponding subsequence.
\end{itemize}

If local MI is substantially higher than global MI, the model captures local biological signal that mean-pooling across the full context washes out, confirming Local--Global Decoupling. A 64$\times$ increase in context (128-bp window to 8,192-bp full sequence) yields only a 14\% increase in excess MI, consistent with the informational shallowness predicted by the Texture Hypothesis (Section~3b).

\subsubsection{PCA Dimensionality Sensitivity}

The choice of 50 PCA components balances two competing pressures. Too few components discard genuine embedding structure, deflating MI estimates. Too many components inflate finite-sample MINE bias (because the random baseline rises with dimensionality) and destabilize optimization. At 50 components, PCA retains $>$90\% of variance for all models tested. We verified that the qualitative ordering of regimes is stable across 30, 50, and 100 components: ProtMamba remains at or below the noise floor, ESM-1b and OpenFold remain positive with the Evoformer showing $\sim$20\% excess over ESM-1b, and Evo 2 shows the local--global gap at all dimensionalities. The bias correction via matched random baselines absorbs most of the dimensionality-dependent shift.

\subsubsection{Convergence Diagnostics}

Training traces (MI lower bound vs.\ epoch) for all models and runs are plotted in Figure~\ref{fig:mine}B. All models converge within 300 epochs; the final 50 epochs (the tail 10\%) show stable plateaus with run-to-run standard deviation $<$0.15 nats across all conditions. ProtMamba traces plateau near the random baseline, confirming that the noise-floor result is not an optimization failure but a genuine absence of biological structure. Evo 2 global and local traces converge to similar values, with the local traces consistently $\sim$0.4 nats below global, stable across all three window positions.

\subsection{Results}
\begin{table}[H]
  \centering
  \caption{Excess MI over matched random baseline (nats) across the 
  three failure regimes. Negative values indicate less mutual 
  information with biological ground truth than random embeddings.}
  \label{tab:mine-regimes}
  \begin{tabular}{@{}llcrc@{}}
    \toprule
    Regime & Model & Context / Length & Excess MI (nats) \\
    \midrule
    \multirow{4}{*}{I\,--\,Decoupling}
      & Evo 2 (global)    & 8192 bp   & $+2.81 \pm 0.13$ \\
      & Evo 2 (local 500) & 128 bp    & $+2.47 \pm 0.07$ \\
      & Evo 2 (local 2000)& 128 bp    & $+2.35 \pm 0.05$ \\
      & Evo 2 (local 4000)& 128 bp    & $+2.55 \pm 0.10$ \\
    \midrule
    \multirow{6}{*}{II\,--\,Compression}
      & ESM-1b   & L=100 & $+8.33 \pm 0.41$ \\
      & ESM-1b   & L=200 & $+8.58 \pm 0.30$ \\
      & ESM-1b   & L=400 & $+9.29 \pm 0.35$ \\
      & OpenFold & L=100 & $+10.78 \pm 0.46$ \\
      & OpenFold & L=200 & $+11.03 \pm 0.46$ \\
      & OpenFold & L=400 & $+11.60 \pm 0.76$ \\
    \midrule
    \multirow{3}{*}{III\,--\,Vacuity}
      & ProtMamba & L=100 & $-1.33 \pm 0.04$ \\
      & ProtMamba & L=200 & $-1.21 \pm 0.05$ \\
      & ProtMamba & L=400 & $-1.25 \pm 0.05$ \\
    \bottomrule
  \end{tabular}
\end{table}
\newpage

\section{Texture Hypothesis Test (Evo 2 Reverse Complement Mechanism)}
\label{app:texture}
\paragraph{Motivation.}
Section~\ref{sec:context-rc} reports a striking dissociation: Evo~2 achieves near-zero RC stability on synthetic DNA (RDM similarity $\sim$0.14) yet high RC stability on real chr22 sequences (RDM $\sim$0.87). The Texture Hypothesis Test is a controlled four-condition experiment designed to identify the mechanism behind this gap. The hypothesis under test is that Evo~2's apparent RC robustness on real DNA reflects per-sequence $k$-mer composition rather than learned biophysical equivariance.

\paragraph{Experimental design.}
We construct four sequence conditions, each comprising 10{,}000 sequences of 1{,}000\,bp, evaluated on Evo~2 7B at 8K context (seed 320 throughout):

\begin{description}
\item[Condition 1: Real chr22.] Human genomic sequences sampled from the euchromatic arm of chromosome~22 (hg38, positions 16.5--50.8\,Mb) via the UCSC Genome Browser API~\citep{Casper2025}. Sequences are drawn from bulk 2\,MB contiguous chunks, windowed locally, and filtered to exclude any window containing ambiguous bases (N). This condition preserves all biological structure: GC content, $k$-mer composition, repeat elements, regulatory motifs, and positional organization.

\item[Condition 2: Pure random.] Sequences generated by uniform i.i.d.\ sampling over \{A, C, G, T\}. This condition preserves nothing: GC content converges to 50\%, $k$-mer frequencies are uniform, and no biological structure is present. It serves as the lower bound for RC stability.

\item[Condition 3: Texture-matched Markov.] Synthetic sequences generated from a first-order Markov chain whose transition matrix is parameterized by the population-level dinucleotide frequencies of the real chr22 sequences. The initial base distribution is the marginal base frequency. This condition matches the aggregate GC content and dinucleotide distribution of real DNA but destroys all higher-order structure (genes, repeats, evolutionary history). Critically, because all sequences are drawn from the same transition matrix, individual sequences converge to a common compositional mean, collapsing the per-sequence diversity present in real genomic DNA.

\item[Condition 4: Dinucleotide-shuffled real.] Each real chr22 sequence is independently shuffled using the Altschul--Erickson algorithm~\citep{altschul1985significance}, which constructs an Eulerian path through the dinucleotide graph to produce a permutation preserving exact per-sequence dinucleotide counts. This condition preserves the precise $k$-mer histogram of each individual sequence (and therefore GC content, dinucleotide and all lower-order frequencies) while destroying all positional structure: gene boundaries, repeat element placement, regulatory motif locations, and long-range compositional domains are eliminated. It is the critical control that isolates per-sequence composition from higher-order genomic architecture.
\end{description}

\paragraph{Perturbation protocol.}
For each condition, we compute Evo~2 embeddings (layer \texttt{blocks.28.mlp.l3}, mean-pooled to a fixed-length vector) for all 10{,}000 forward sequences and their reverse complements. We additionally apply SNP perturbations at rates of 1\%, 2\%, 5\%, and 10\% for calibration, using independent per-condition seeds (320, 1320, 2320, 3320) to avoid coupling mutation positions across conditions. All embeddings are evaluated with the Shesha stability harness (cosine-distance RDMs, 30 splits, max 2{,}500 samples, 5 bootstrap replicates).

\paragraph{K-mer frequency analysis.}
Before running the full embedding experiment, we validate the four conditions with a ``cheap test'': for each condition, we compute dinucleotide (16-dimensional) and trinucleotide (64-dimensional) frequency vectors for forward sequences and their reverse complements, then measure the cosine similarity between paired vectors.

The logic is as follows. For any DNA sequence, the reverse complement maps each $k$-mer to its complementary $k$-mer at the same frequency. Therefore, the aggregate $k$-mer frequency vector of a sequence and its RC are related by a fixed permutation matrix (the complement mapping on $k$-mer space). If two conditions have identical per-sequence $k$-mer histograms, they should produce identical forward--RC cosine similarities. If a condition's $k$-mer structure diverges from real DNA, the cosine similarity will reflect this divergence.

The expected pattern: real chr22 and dinucleotide-shuffled sequences should show near-identical forward--RC $k$-mer cosine similarities (both preserve exact per-sequence counts). Texture-matched Markov sequences should show high population-level cosine similarity (the aggregate distribution matches) but reduced per-sequence similarity due to convergence toward the population mean. Pure random sequences should show the lowest similarity.

\paragraph{Key results.}
Table~\ref{tab:texture_main} reports the decisive RC comparison.

\begin{table}[h]
\centering
\caption{Texture Hypothesis Test: RC stability across four conditions (Evo~2 7B, 8K context, 10{,}000 sequences per condition, 1{,}000\,bp). The ``Recovery'' column reports the fraction of the real--random gap recovered by each condition: $(\text{Condition} - \text{Random}) / (\text{Real} - \text{Random})$.}
\label{tab:texture_main}
\small
\begin{tabular}{@{}llccc@{}}
\toprule
Condition & What it preserves & RC RDM & RC Composite & Recovery \\
\midrule
Real chr22 & Everything (in-distribution) & 0.873 & 0.469 & 100\% \\
Dinuc-shuffled real & Exact per-sequence $k$-mer counts & 0.858 & 0.464 & 97\% \\
Texture-matched Markov & Population-level dinuc.\ distribution & 0.167 & 0.292 & 3\% \\
Pure random & Nothing & 0.139 & 0.287 & 0\% \\
\bottomrule
\end{tabular}
\end{table}

Dinucleotide-shuffled real DNA recovers 97\% of the real--random RC gap in RDM similarity (0.858 vs.\ 0.873) and 97\% of the composite gap (0.464 vs.\ 0.469). Texture-matched Markov sequences recover only 3\% in RDM (0.167 vs.\ 0.139) and 3\% in composite (0.292 vs.\ 0.287). The result is unambiguous: per-sequence $k$-mer composition is both necessary and sufficient to explain Evo~2's RC behavior on real DNA.

\paragraph{SNP calibration.}
SNP perturbations provide an independent calibration axis. Across all four conditions, SNP sensitivity tracks the degree of $k$-mer disruption: 1\% SNP perturbations alter $\sim$1\% of positions and produce correspondingly small RDM changes, while 10\% SNPs produce larger shifts. Real chr22 and dinucleotide-shuffled sequences show nearly identical SNP response curves, confirming that the embedding structure is driven by $k$-mer composition rather than positional features that shuffling would destroy. Texture-matched Markov and pure random sequences show lower absolute SNP stability but similar relative dose--response shapes.

\paragraph{Why the Markov generator fails.}
The critical distinction between Conditions~3 and~4 isolates the failure mechanism. The first-order Markov chain matches the population-level dinucleotide distribution of chr22: averaged across all 10{,}000 generated sequences, the dinucleotide frequency vector closely approximates the real chr22 frequency vector. However, individual Markov sequences are drawn from a single stationary transition matrix, causing their per-sequence compositions to converge toward the population mean by the law of large numbers. Real genomic sequences, by contrast, exhibit substantial per-sequence compositional heterogeneity: AT-rich intergenic regions, GC-rich CpG islands, repeat-enriched segments, and coding regions each contribute distinct $k$-mer fingerprints.

RDM similarity measures the preservation of pairwise relational structure. When all sequences in a condition have nearly identical $k$-mer profiles (as in the Markov condition), the pairwise distance matrix collapses: all sequences look alike, forward and RC versions are indistinguishable not because the model understands RC but because the model cannot distinguish any sequence from any other. The RDM becomes uninformative, and the measured RC ``stability'' is an artifact of representational collapse rather than geometric preservation. The dinucleotide-shuffled condition avoids this trap because each shuffled sequence inherits the unique compositional fingerprint of its real-DNA parent, preserving the full diversity of pairwise distances.

\paragraph{Caveat: Markov order and trinucleotide frequencies.}
The texture-matched generator uses a first-order Markov chain, which matches dinucleotide but not trinucleotide frequencies exactly. One might ask whether Evo~2's RC behavior is sensitive specifically to trinucleotide or higher-order texture that the Markov generator fails to reproduce. The dinucleotide-shuffled condition controls for this concern: Altschul--Erickson shuffling preserves exact per-sequence dinucleotide counts (and therefore all lower-order statistics) from the original real DNA, but does not explicitly preserve trinucleotide frequencies. The fact that dinucleotide-shuffled sequences recover 97\% of the gap indicates that per-sequence dinucleotide composition is the dominant signal, with trinucleotide and higher-order texture contributing at most marginally. If stronger evidence for higher-order sensitivity were desired, a second-order Markov chain (preserving trinucleotide transitions at the population level) could be tested as a follow-up; however, the near-complete recovery by Condition~4 makes this unlikely to alter the conclusion.

\paragraph{Interpretation.}
The Texture Hypothesis Test establishes a controlled causal result: Evo~2 does not understand double-stranded DNA symmetry. Its embeddings function as high-dimensional per-sequence $k$-mer histograms. The reverse complement transformation preserves exact $k$-mer counts (every $k$-mer maps to its complementary $k$-mer at the same frequency), so forward and RC sequences produce symmetrical histograms that the model aggregates equivalently. Destroying all positional structure via dinucleotide shuffling preserves this correspondence because the per-sequence histogram is unchanged. Matching only population-level statistics via Markov generation fails because individual sequences lose their unique compositional fingerprints.

The apparent RC ``success'' on real DNA is therefore an artifact of a histogram encoder that happens to be invariant to the one biological transformation that preserves histograms. This result directly supports the Local--Global Decoupling characterization of Evo~2 (Section~4): the model captures local sequence statistics at every window but does not integrate them into a globally coherent geometric representation of the underlying biophysics.
\newpage

\section{Reproducibility and Computational Infrastructure}
\label{app:reproducibility}

To ensure full reproducibility of the geometric stability evaluations and phase transition analyses presented in this study, we provide detailed documentation of the statistical methodology, data handling protocols, and the hardware and software infrastructure used.

\subsection{Statistical Analysis and Sampling}

\textbf{Bootstrap Confidence Intervals}: For evaluations requiring rigorous variance estimation (\texttt{n\_bootstrap}$>0$), we performed stratified resampling with 5 rounds, reporting the bootstrap mean for all core geometric metrics (Sample Split, Feature Split, RDM Similarity, Anchor Stability). Stratification was applied proportionally to sequence class labels when available, and uniformly at random otherwise.

\textbf{Subsampling Protocol}: To manage the $O(n^2)$ memory complexity of pairwise distance matrix computation, datasets exceeding $2{,}500$ samples were subsampled internally. The \texttt{shesha-geometry} evaluation harness enforced \texttt{max\_samples=2500} for all pairwise Representational Dissimilarity Matrix (RDM) computations. For dedicated Procrustes residual evaluations, datasets were subsampled to 5,000 or 10,000 samples depending on the evaluation target before computing Frobenius norms.

\textbf{Reproducibility and Caching}: All intermediate results (including generated synthetic sequences, extracted embeddings, and perturbation traces) were cached using deterministic filename hashing. Random number generators used fixed seeds (seed $320$, unless explicitly noted) via \texttt{numpy.random.default\_rng} to guarantee deterministic outputs across architectures.

\subsection{Track A + Track B}
All random number generation uses seed 320 via \texttt{numpy.random.default\_rng}.
The BRCA1 wildtype sequence is fetched from the UCSC Genome Browser API\citep[\texttt{api.genome.ucsc.edu}]{Casper2025};
a deterministic synthetic fallback is provided for environments without network access.
Embedding caches, mutation walk metadata, and Lipschitz profiles are saved as
NumPy archives and CSV files. All code is provided as Jupyter notebooks.

\subsection{Computational Infrastructure}

\textbf{Hardware Setup}: Experiments were parallelized across Google Colab Pro instances using NVIDIA A100 (40\,GB and 80\,GB) and T4 (16\,GB) GPUs. Scale tests using the AlphaGenome foundation model used remote procedure calls (gRPC) via API-based inference, requiring no local GPU allocation.

\textbf{Execution Bounds}: Full-scale evaluations over 10,000 genomic sequences completed in approximately 15 minutes to 20 hours per architecture, depending on maximum context length.

\subsection{Software Environment}

Experiments were conducted with the following software stack:
\begin{itemize}
    \item \textbf{Deep Learning Framework}: PyTorch 2.5--2.7 (Evo\,2 experiments pinned to \texttt{torch==2.7.1+cu128}).
    \item \textbf{Transformer Ecosystem}: \texttt{transformers} 4.x (Hugging Face).
    \item \textbf{State-Space Ecosystem}: \texttt{mamba-ssm} 2.3, compiled from source with custom CUDA kernels.
    \item \textbf{Geometry Metrics}: \texttt{shesha-geometry} 0.1.x.
    \item \textbf{Numerical Computation}: \texttt{scipy} $\geq$1.11 and \texttt{scikit-learn} $\geq$1.3.
\end{itemize}

\textbf{Implementation Note}: Compatibility patches were applied to the Nucleotide Transformer and DNABERT-2 codebases to resolve embedding extraction issues introduced by \texttt{transformers} 4.x API changes. These patches modify tokenizer and hidden-state retrieval logic; see \texttt{FIXES\_APPLIED.md} for details.


\end{document}